%% file: main.tex
\documentclass{article}

\usepackage{microtype}
\usepackage{graphicx}
\usepackage{subfigure}
\usepackage{booktabs} % for professional tables

\usepackage[accepted]{icml2023}

\usepackage{amsmath}
\usepackage{amssymb}
\usepackage{mathtools}
\usepackage{amsthm}

\theoremstyle{plain}

\theoremstyle{definition}

\theoremstyle{remark}

\usepackage{amsfonts}       % blackboard math symbols
\usepackage{xcolor}         % colors
\usepackage{url}
\usepackage{bm}
\usepackage{footnote}
\usepackage{multirow, multicol}
\usepackage{wrapfig}
\usepackage{comment}
\usepackage{caption}
\usepackage{makecell}
\usepackage{threeparttable}

\usepackage{colortbl}
\usepackage{arydshln}
\usepackage{eqparbox}

\usepackage{graphbox}
\DeclareMathOperator*{\minimize}{minimize}

\usepackage{hyperref}

\definecolor{Red}{rgb}{0.6,0,0}
\definecolor{Blue}{rgb}{0,0,1}
\definecolor{Green}{rgb}{0,0.4,0.7}
\definecolor{airforceblue}{rgb}{0.36, 0.54, 0.66}
\definecolor{ao(english)}{rgb}{0.0, 0.5, 0.0}
\definecolor{azure(colorwheel)}{rgb}{0.0, 0.5, 1.0}
\definecolor{crimson}{rgb}{0.86, 0.08, 0.24}
\definecolor{darkcerulean}{rgb}{0.03, 0.27, 0.49}
\definecolor{cobalt}{rgb}{0.0, 0.28, 0.67}
\definecolor{rosegold}{rgb}{0.72, 0.43, 0.47}
\definecolor{orange-red}{rgb}{1.0, 0.27, 0.0}
\definecolor{mountainmeadow}{rgb}{0.19, 0.73, 0.56}
\definecolor{malachite}{rgb}{0.04, 0.85, 0.32}
\definecolor{darkblue}{rgb}{0.0, 0.0, 0.55}
\definecolor{MidnightBlue}{rgb}{0.0, 0.0, 0.55}

\definecolor{customblue}{rgb}{0.2, 0.35, 0.8}
\definecolor{customcolor}{gray}{0.}
\definecolor{gg}{gray}{0.9}
\definecolor{tg}{gray}{0.6}
\hypersetup{colorlinks=true}
\hypersetup{linktoc=all}
\hypersetup{citecolor=customblue}
\hypersetup{linkcolor=crimson}
\hypersetup{urlcolor=MidnightBlue}
\usepackage[all]{hypcap}
\usepackage[nameinlink,noabbrev]{cleveref}

\newcommand{\bsy}{\boldsymbol}
\newcommand{\highlight}[1]{{\color{crimson}{#1}}}
\makeatletter
\def\adl@drawiv#1#2#3{%
        \hskip.5\tabcolsep
        \xleaders#3{#2.5\@tempdimb #1{1}#2.5\@tempdimb}%
                #2\z@ plus1fil minus1fil\relax
        \hskip.5\tabcolsep}
\newcommand{\cdashlinelr}[1]{%
  \noalign{\vskip\aboverulesep
           \global\let\@dashdrawstore\adl@draw
           \global\let\adl@draw\adl@drawiv}
  \cdashline{#1}
  \noalign{\global\let\adl@draw\@dashdrawstore
           \vskip\belowrulesep}}
\makeatother

\renewcommand{\algorithmiccomment}[1]{\hfill\eqparbox{COMMENT}{\footnotesize $\rhd$ #1}}
\newcommand{\eat}[1]{}

\usepackage[textsize=tiny]{todonotes}

\icmltitlerunning{Continual Learners are Incremental Model Generalizers}

\begin{document}

\twocolumn[
\icmltitle{Continual Learners are Incremental Model Generalizers}
% \icmlsetsymbol{intern}{*}

\begin{icmlauthorlist}
\icmlauthor{Jaehong Yoon}{kaist,msr}%\thanks{The work was done while the author was an intern at Microsoft Research.}
\icmlauthor{Sung Ju Hwang}{kaist,da}
\icmlauthor{Yue Cao}{baai}
\end{icmlauthorlist}

\icmlaffiliation{kaist}{Korea Advanced Institute of Science and Technology}
\icmlaffiliation{msr}{Microsoft Research}
\icmlaffiliation{baai}{Beijing Academy of Artificial Intelligence}
\icmlaffiliation{da}{DeepAuto}
% \icmlaffiliation{intern}{The work was done while the author was an intern at Microsoft Research.}

\icmlcorrespondingauthor{Jaehong Yoon}{jaehong.yoon@kaist.ac.kr}
% \icmlcorrespondingauthor{Jaehong Yoon}{sjhwang8282@kaist.ac.kr}

\icmlkeywords{Machine Learning, ICML}

\vskip 0.3in
]
\printAffiliationsAndNotice{JY was an intern at Microsoft Research.}  % leave blank if no need to mention equal contribution
% \printAffiliationsAndNotice{}

\input{sections/1_abstract}
\input{sections/2_introduction}
\input{sections/3_related_work}
\input{sections/4_method_basic}
\input{sections/4_method_details}
\input{sections/5_experiments}
\input{sections/6_conclusion}

% \section*{Reproducibility Statement}
% \label{sec:reproduce}
% We will release our code and the implementation details to be publicly available for reproducibility. In addition, we provide more details in Appendix for reproducibility, including architecture and baselines setup with hyperparameters.

% {\small
\bibliographystyle{icml2023}
\bibliography{references}
% }

\input{sections/7_appendix}
%%%%%%%%%%%%%%%%%%%%%%%%%%%%%%%%%%%%%%%%%%%%%%%%%%%%%%%%%%%%
\end{document}

%% file: sections/1_abstract.tex
\begin{abstract}
Motivated by the efficiency and rapid convergence of pre-trained models for solving downstream tasks, this paper extensively studies the impact of Continual Learning (CL) models as pre-trainers. 
In both supervised and unsupervised CL, we find that the transfer quality of the representation often increases gradually without noticeable degradation in fine-tuning performance. This is because CL models can learn improved task-general features when easily forgetting task-specific knowledge. Based on this observation, we suggest a new unsupervised CL framework with masked modeling, which aims to capture fluent task-generic representation during training.
Furthermore, we propose a new fine-tuning scheme, \emph{GLobal Attention Discretization (GLAD)}, that preserves rich task-generic representation during solving downstream tasks. The model fine-tuned with GLAD achieves competitive performance and can also be used as a good pre-trained model itself.
We believe this paper breaks the barriers between pre-training and fine-tuning steps and leads to a sustainable learning framework in which the continual learner incrementally improves model generalization, yielding better transfer to unseen tasks.
\end{abstract}

%% file: sections/2_introduction.tex
\section{Introduction}
Unsupervised Representation Learning (URL)~\cite{radford2015unsupervised, gidaris2018unsupervised, grill20byol, xie2021propagate} is a pertinent branch of machine learning in which a model exploits data without human-generated signals to extract the generic representations. 
Although the standard URL scenario assumes that we have a complete unlabeled dataset before training, this setting is often unrealistic in the real world; as the world persistently changes, the model should cope with non-stationary data throughout its lifespan. It carries the lifelong learnability of the representation model. As motivated by the Continual Learning (CL) field~\cite{ThrunS1995, silver2002task, KumarA2012icml, LiZ2016eccv}, Unsupervised Continual Learning (UCL)~\cite{rao19curl, madaan2022rethinking, fini2022self} has recently been explored to address the limitations of the conventional representation learning setup and provides comprehensive analyses regarding the quality of learned representations along with their forgetting.

However, the recently proposed UCL frameworks have clear limitations in their interpretation of model transfer to downstream tasks. Suppose $r_{i,j}$ be the performance of a pre-defined supervised metric for task $j$ using a sequentially pre-trained backbone model from the first to $i^{th}$ task. They train on $T$ sequential tasks $\{\mathcal{T}_t\}^T_{t=1}$ without labels, and measure the effectiveness of their representation model leveraging two supervised metrics: 1) the averaged performance gap measured \emph{immediately after the task is learned} and \emph{after all tasks are learned}  $\sum^{T-1}_{t=1}r_{t}=r_{T,t}-r_{t,t}$ (backward transfer)~\cite{lopez2017gradient} and 2) the averaged performance of all tasks $\sum^{T}_{t=1}r_{t}=r_{T,t}$.
Though maximizing the transferability of the learned representations on the target problem is essential for general-purpose models, prior UCL works have confined their validations and analyses to \textbf{linear evaluations} as updating the linear classifier with keeping fixed representation model backbones. This can be suitable for measuring direct differences in model drift (i.e., catastrophic forgetting) of continual learners. 
Yet, it cannot disclose the change in knowledge transferability during task sequential training, which is crucial in utilizing the pre-trained model in practice. 

\input{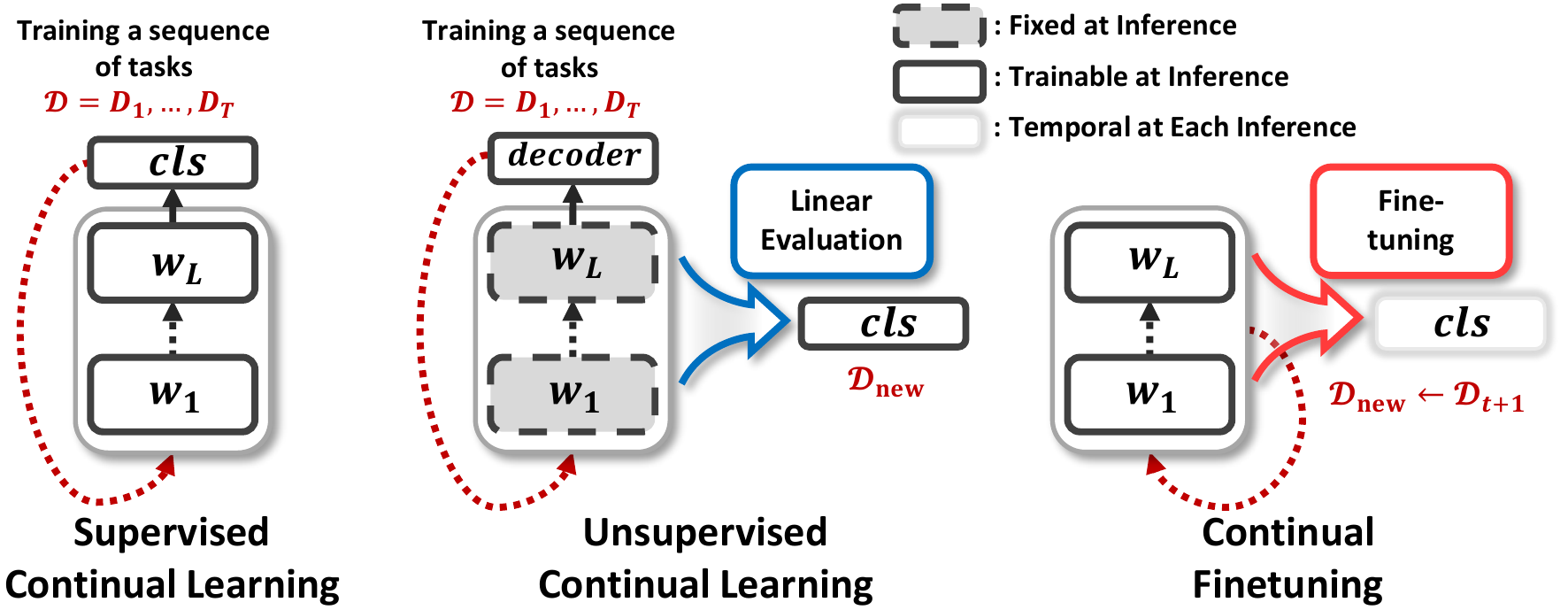}
Beyond the limited understanding of knowledge transfer from prior works, we provide comprehensive transferability analyses with varying evaluation setups via supervised and unsupervised CL methods to explore their potential as a pre-trainer.
In~\Cref{fig:concept}, we perform a simple experiment to investigate the potential of incremental model generalization via a continual learning setup.
% First, we perform a simple experiment to confirm the potential of continual learning for pre-training. 
A model sequentially learns the nine tasks from ImageNet1K-split (containing ten tasks in total) using supervised CL methods: \emph{Base} (a CL model without any additional method), \textit{SI}~\cite{zenke2017continual}, and \textit{DER}~\cite{buzzega2020dark}.
Then, we measure the change of representation quality through linear evaluation for the first task, T0, and fine-tuning for the out-of-distribution task, T9. As shown in the increased performance in \Cref{fig:concept} \highlight{Right}, we find that CL methods gain steady increases in their transferability to unseen tasks without the concern of performance degeneration from representational forgetting if the model trains on more tasks in a sequential manner. It is distinguishable from T0 linear evaluation results of the same methods, which suffer from performance degradation (\Cref{fig:concept} \highlight{Left}). 
% We found that this is because the model obtains more diverse low-level features throughout continual learning (\Cref{fig:attn-dist-and-entropy,fig:swin-attn-div}).

% Surprisingly, the fine-tuning performance on OOD tasks consistently increases if the continual learning model trains on more tasks in a sequential manner, based on the results of various CL methods (vanilla, regularization-based, and rehearsal-based CL). Moreover, we found that this is because the model obtains more diverse low-level features throughout continual learning (Figures 2 and 5).

We explain these phenomena because the transferability hinges upon rich task-generic features in pre-trained models~\cite{xie2022revealing,wei2022contrastive}, while the model mostly loses task-specific features during CL, particularly severe when they train in a supervised/contrastive manner (please see \Cref{fig:attn-dist-and-entropy,fig:swin-attn-div}). Inspired by our above observations, we propose a new UCL framework based on Masked Image Modeling (MIM)~\cite{xie2021simmim,he2022masked} that improves task-generic representation across all layers during training. Our framework retains fluent task-generic features over all layers by learning to predict masked regions of input images during unsupervised CL, conditioning other available areas. and outperforms existing supervised/unsupervised CL baselines in fine-tuning. Then we demonstrated that the suggested reconstruction-based CL framework achieves substantially higher fine-tuning performance on OOD tasks (\Cref{fig:exp-anal}) than existing CL frameworks that aim to learn class-discriminative features, particularly at deeper layers.

% Our framework retains fluent task-generic features over all layers by learning to predict masked regions of input images, conditioning other available 
% areas. Then we demonstrated that the suggested reconstruction-based CL framework achieves substantially higher fine-tuning performance on OOD tasks (Figure 6) than existing CL frameworks that aim to learn class-discriminative features, particularly at deeper layers (Figure 2).

Additionally, as we observe that continual learning models improve model generalization for downstream tasks, we raise the question of how a fine-tuned model can become a good pre-trained model itself since normal fine-tuning shifts generic features to be task-dependent. To this end, we additionally explore the potential of continual pre-training, reusing the fine-tuned model as a pre-trained model for other downstream tasks. The objective is to encourage the fine-tuning model to retain rich task-generic features during supervised fine-tuning.
Leveraging our motivations for building the MIM-based UCL framework, we suggest a new method, named \textbf{GL}obal \textbf{A}ttention \textbf{D}iscretization (\textbf{GLAD}).
Our proposed method introduces a lightweight, trainable adapter in the multi-head attention module of the ViT backbone. For future reuse of fine-tuned networks, GLAD aims to keep improving global (task-generic) attention during fine-tuning through a constraint to encourage diversity of attention distance in the adaptor-free backbone while the model solves the downstream task by focusing on local (task-adaptive) features utilizing adaptor-guided attentions.
We believe our observations and proposed approach lead to removing the barriers between the standard pretraining-finetuning scheme and continual learning towards 
%incremental model generalization via never-ending continual training. 
improving model generalization via never-ending continual training while alleviating the threat of loss of generality of large pre-trained models during downstream task fine-tuning.

The main contributions of the paper are threefold:
\begin{itemize}
\item We unveil the behavior of representational transferability and forgetting of task-generic and task-specific features under multiple supervised/unsupervised continual learning frameworks at scale, with Vision Transformer backbones.

\item We suggest a new learning/evaluation paradigm of the popular pretraining-finetuning scheme amalgamating to continual learning that aims to continuously increase the generalization of the pre-training backbone during the endless sequential fine-tuning phases.

\item We further suggest a simple yet efficient remedy to increment task-generic feature expressiveness throughout continual pre-training, dubbed \emph{GLAD}, which enables the model rapidly adapts to the target problem while preserving high transfer affinity to future tasks.
\end{itemize}

% \TBD{
% We first unveiled the potential of incremental model generalization via a continual learning setup (Figure 1). 
% Surprisingly, the fine-tuning performance on OOD tasks consistently increases if the continual learning model trains on more tasks in a sequential manner, based on the results of various CL methods (vanilla, regularization-based, and rehearsal-based CL). Moreover, we found that this is because the model obtains more diverse low-level features throughout continual learning (Figures 2 and 5).

% Inspired by our above observations, we proposed a new UCL method to improve model generalization ability for future tasks. Our framework retains fluent task-generic features over all layers by learning to predict masked regions of input images, conditioning other available areas. Then we demonstrated that the suggested reconstruction-based CL framework achieves substantially higher fine-tuning performance on OOD tasks (Figure 6) than existing CL frameworks that aim to learn class-discriminative features, particularly at deeper layers (Figure 2).
% }

% We believe our observations and proposed approach lead to removing the barriers between the standard pretraining-finetuning scheme and continual learning towards improving model generalization via never-ending continual training while alleviating the threat of loss of generality of large pre-trained models during downstream task fine-tuning.}

%% file: materials/figures/concept_figure.tex
\begin{figure}[t]
    \small
    \centering
    \vspace{-0.1in}
    \hspace{-0.1in}
    \begin{tabular}{cc}          
    \hspace{-0.2in}\includegraphics[height=3.7cm]{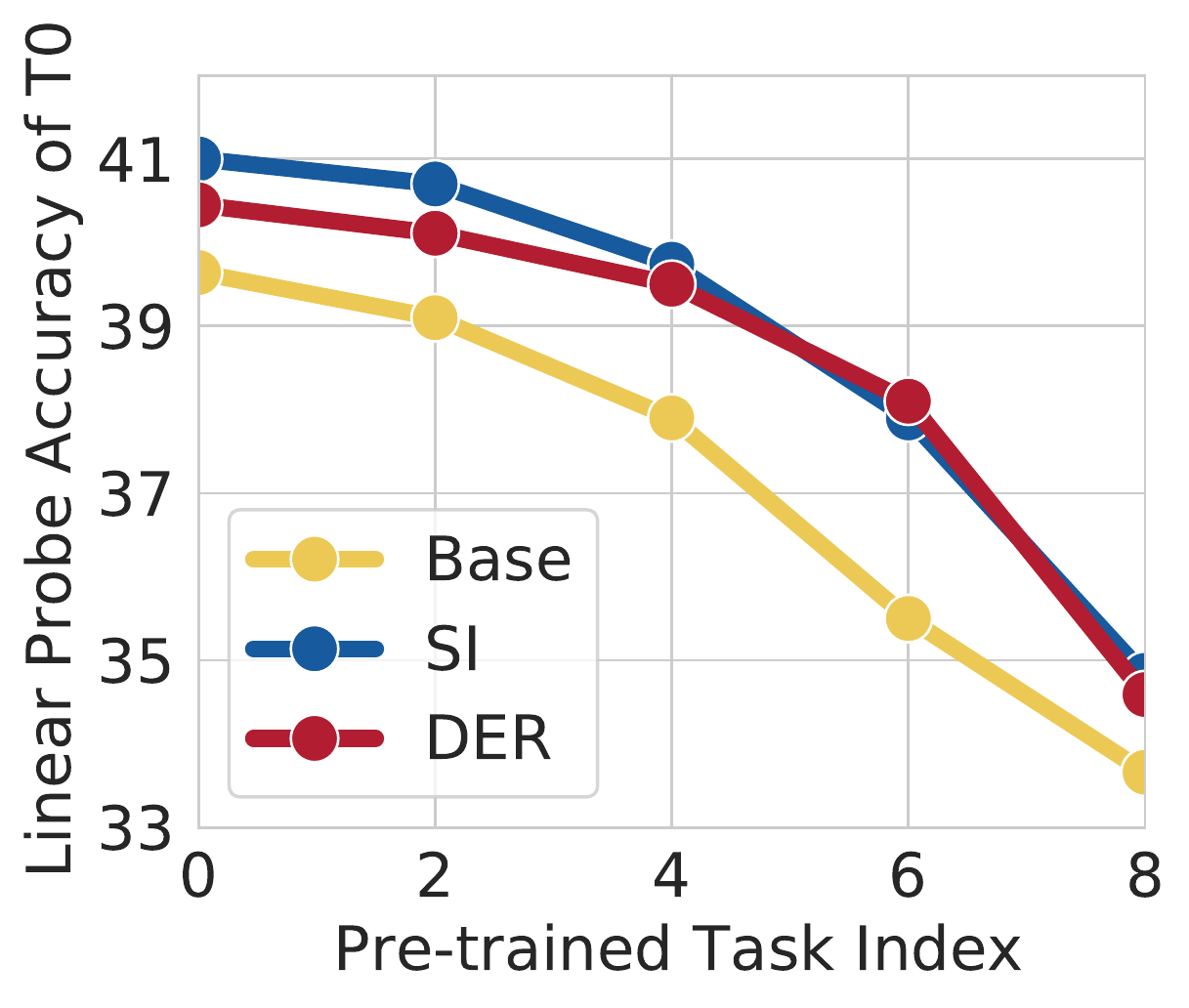}&
    \hspace{-0.2in}\includegraphics[height=3.7cm]{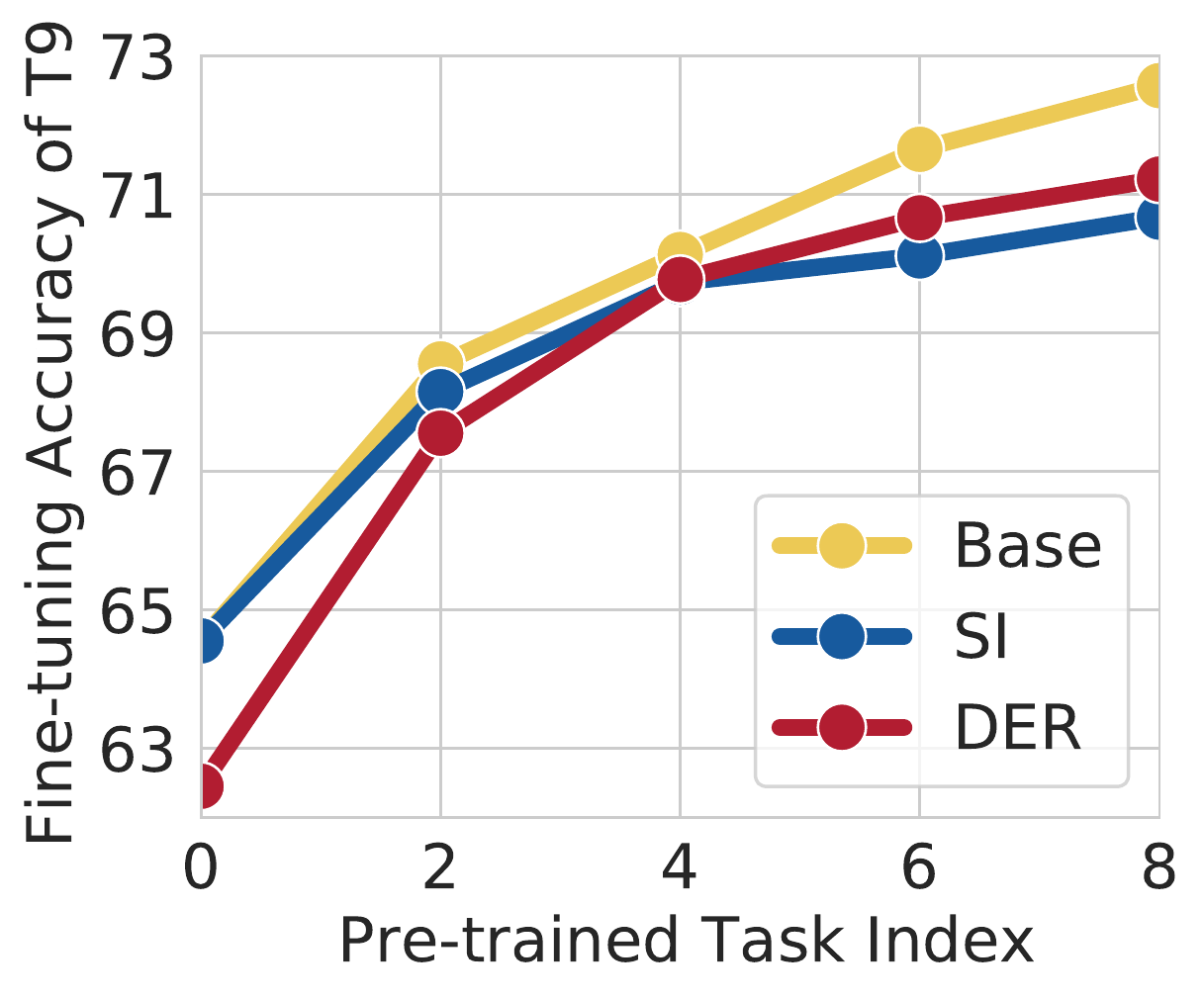}\\
    \end{tabular}
    \vspace{-0.1in}
    \caption{\small
    \textbf{Linear evaluation of the first task (T0) and fine-tuning of the unseen task (T9) during supervised CL} over a sequence of nine tasks from ImageNet1K-split (T0-T8). Sequential training on more tasks decreases the linear evaluation performance, but it increases fine-tuning performance on the unseen task.}
    \label{fig:concept}
    \vspace{-0.1in}
\end{figure}

% \begin{figure*}
%     \small
%     \centering
%     \vspace{-0.1in}
%     \begin{tabular}{cc}          
%     \includegraphics[height=4cm]{figures/concept_figure2.pdf}&
%     \includegraphics[height=4cm]{figures/temp_fortask9.pdf} \\
%     (a) Illustration of continual training schemes & (b) Generalization of SCL/UCL\\
%     \end{tabular}
%     \vspace{-0.05in}
%     \caption{\footnotesize \textbf{(a) Continual Pre-training is concerned with maximizing model generalization for full-finetuning on future tasks.} It performs pre-training on sequential tasks regardless of label supervision, while UCL framework prevents the model from updating backbone weights at evaluation to measure representational drift when it proceeds to the next task.
%     \textbf{(b) Continual pre-training increments model generalization.} Sequential pre-training on more tasks further increases fine-tuning performance on the unseen downstream task (T9).}
%     \label{fig:concept}
%     % \vspace{-0.1in}
% \end{figure*}

%% file: sections/3_related_work.tex
\section{Related Work}

\paragraph{Continual learning} SI~\cite{zenke2017continual} introduces an additional surrogate loss that reduces the weight shift during continual learning by maintaining the training trajectory according to the weight importance of previous tasks. 
DEN~\cite{yoon2018lifelong} adaptively controls the network capacity by adding/pruning parameters when new tasks arrive. DER~\cite{buzzega2020dark} stores a few training instances of previous tasks as well as their predicted logits and minimize the similarity to produce similar logit predictions on past tasks. BiC~\cite{wu2019large} adds a new layer at the top of the backbone to correct classification bias on new tasks. Similarly, WA~\cite{zhao2020maintaining} corrects the prediction bias by rescaling the FC layer with averaged weights normalization on past tasks. DyToX~\cite{douillard2021dytox} adopts ViT and performs ensembled prediction with task-specific classifiers leveraging additional task-specific tokens. However, dominant research resorts to the sophisticated human annotation of inputs during training a sequence of tasks.

CURL~\cite{rao19curl} learns unsupervised representation on task sequences with a generative model adopting task-specific inference. However, the method is validated for only MNIST-scale datasets due to their limited scalability by design. \citet{madaan2022rethinking} suggest a new unsupervised continual learning framework in a contrastive manner using Siamese structures. They demonstrate the scalability of the proposed framework through comprehensive analyses of learned representations. CaSSLe~\cite{fini2022self} utilizes a similar contrastive self-supervised framework for unsupervised continual learning, yet provides further extensive validations including diverse self-supervised learning backbones over ImageNet-100. 

Very recently, \citet{chen2023forgetting} provides intriguing discussions on forgetting and forward transfer of \textit{FOMAML}~\cite{finn2017model} during supervised CL, primarily via quantitative evaluation with a few-shot linear probe.
On the other hand, we extensively consider not only supervised CL but also siamese network-based contrastive and reconstruction-based unsupervised CL frameworks. 
In our work, we found that the continually learned representation behaves differently, depending on whether transferring via fine-tuning the entire model or linear evaluation with a scalable test set. And we deliver various discussion points on representation transferability through both quantitative and qualitative analyses, allowing the re-update of learned backbone weights for downstream tasks. Also, we further propose the GLAD module based on our findings, which preserves rich task-generic representation during solving downstream tasks. More discussions regarding meta-learning are provided in~\Cref{sup:sec:meta-learning}.

\paragraph{Self-supervised learning} 
Simsiam~\cite{chen2020simple} maximizes the similarity of input prediction upon two different augmentations using the Siamese network, learning input-invariant self-supervision. BarlowTwins~\cite{zbontar2021barlow} aims to remove cross-correlation across different feature vector embeddings from Siamese networks. DINO~\cite{caron2021emerging} distills teacher model predictions to the student by minimizing cross-entropy loss between their predictions, where the teacher model is updated through an exponential moving average from the student model. Unlike contrastive learning-based directions, Masked Image Modeling (MIM) has recently been developed inspired by masked language models for natural language understanding. SimMIM~\cite{xie2021simmim} and MAE~\cite{he2022masked} adopt an encoder-decoder structure that zeroes out random spatial patches in each patchfied image and learns representations by predicting pixel values in masked patches. MSN~\cite{assran2022masked} combines Siamese networks with masked modeling that maximize the prediction similarity between patchfied masked inputs and the augmented target views.

%% file: sections/4_method_basic.tex
% \section{Continual fine-tuning: Integrated formulation of pre-training-fine-tuning and Continual Learning}
% \section{Recap: Pretraining-Finetuning and Continual Learning}
\section{Preliminaries}
% This section introduces each standard scenario for pre-training-fine-tuning, and supervised and unsupervised continual learning. and explicates the difference between them. 

\subsection{Pre-training and Fine-tuning}\label{subsec:pf}
Given a neural network $f(\cdot;\bm{w})$ parameterized by weights $\bm{w}$, recent works have addressed the broad machine learning problems described to $\mathcal{D}_{target}$ by optimizing learnable weights with respect to complex objective functions. Beyond statistical initialization of network weights~\cite{glorot2010understanding,he2015delving}, pre-training, where leveraging learned weights from scaled benchmark datasets (e.g., ImageNet~\cite{deng2009imagenet}) as the initialization of $\bm{w}$, has been widely adopted to promote a rapid and stable convergence curve during training. 
Self-supervised learning~\cite{chen2020simple, he2020momentum, caron2021emerging, bardes2021vicreg, xie2022revealing} has recently become prevalent for pre-training, demonstrating superior generalization performance compared to supervised counterparts by capturing task-agnostic input features. %However, the representation model loses the high transferability on future generic problems by fitting to the target task after fine-tuning phase.
While multiple different frameworks are considered for self-supervised learning, we exemplify the encoder-decoder framework in this paragraph. Let $h$ and $g$ be an encoder and a decoder parameterized by $\bsy\theta$ and $\bsy\phi$, respectively, the objective function is to minimize self-supervised loss given input data $\bm{d}$ without supervision:
\begin{equation}\label{eq:prt-ft}
\begin{split}
% \minimize_{\bsy\theta, \bsy\phi} \ell\left(g_{\bsy\phi} \circ h_{\bsy\theta}\left(\bm{d}\right)\right),
\bsy\theta^*,~\bsy\phi^*=\text{arg}\min_{\bsy\theta, \bsy\phi} \ell\left(g_{\bsy\phi} \circ h_{\bsy\theta}\left(\bm{d}\right)\right),
\end{split}
\end{equation}
where $\circ$ indicates function composition. The loss function is often designed in several formulations based on similarity, identity correlation, and contrastive loss. After the pre-training phase, the encoder transfers learned features to backbone neural networks for fine-tuning, $\bm{w}\leftarrow\bsy\theta^*$.

\input{materials/figures/4_attn_visualization.tex}

\subsection{Continual Learning Paradigms}\label{subsec:scl}
Supervised Continual Learning (SCL)~\cite{mallya2018packnet, riemer2018learning, aljundi2019online, chaudhry2019continual, chaudhry2020continual, chrysakis2020online, titsias2019functional,  shen2020generative, douillard2021dytox, Yoon2020Scalable, yoon2022online} is about a sustainable adaptation to unlimited task sequences while maintaining proficiency on previous tasks.
Let us consider an intuitive example with the image-based problem: suppose $\mathcal{T}=\{\mathcal{T}_1, ..., \mathcal{T}_T\}$ be a sequence of $T$ tasks, where the dataset $\mathcal{D}_t$ for the $t$-th task consists of $n_t$ training instances $\mathcal{X}_t\in\mathbb{R}^{n_t\times C\times H\times W}$ and corresponding labels $\mathcal{Y}_t\in\mathbb{R}^{n_t}$. That is, $C, H, \text{and~} W$ denotes a channel, height, and width of images, respectively. A continual learner $f_{\bm{w}}$, parameterized by a set of weights $\bm{w}$, aims to predict classes by minimizing the optimization problem: {${minimize}_{\bm{w}}~ \sum^T_{t=1}L_{CE}\left( f\left( \mathcal{X}_{t}; \bm{w} \right), \mathcal{Y}_{t} \right)$}, where $L_{CE}$ is a cross-entropy loss. Yet, we assume that $f_{\bm{w}}$ can access each task in a specific timestep that loses the authorization to revisit previous tasks' data instances when the next task arrives. That is, the model solves the following non-stationary problem at task $t$ throughout task sequential training: 
\begin{equation}\label{eq:scl}
\begin{split}
\bm{w}^*&=arg\min_{\bm{w}}~L_{CE}\left(f\left(\mathcal{X}_t;\bm{w}\right), \mathcal{Y}_t\right)\\
&\approx arg\min_{\bm{w}}~\sum^t_{i=1}L_{CE}\left(f\left(\mathcal{X}_i;\bm{w}\right), \mathcal{Y}_i\right).
\end{split}
\end{equation}
Obtained models directly evaluate the performance of each task, categorizing several incremental learning setups according to the accessibility to task oracle during inference.

% , the $t$-th task consists of training instances $\mathcal{X}_t$ without any human-annotated labels and 
Unsupervised Continual Learning (UCL) is formulated in representation learning frameworks on a sequence of unlabeled tasks $\{\mathcal{X}_t\}_{t=1}^T$, often referred to as continual self-supervised learning.
A learner $f_{\bm{w}}$ aims to find the best solution that learns the informative representation of multiple tasks sequentially. At each timestep $t$, the model resorts to the accessible dataset $\mathcal{X}_t$ without any human-annotated labels to solve the problem:% as follows: % during continual representation learning. 
\begin{equation}\label{eq:ucl}
\begin{split}
\bm{w}^*,~\bm{w}_{ext}&=arg\min_{\bm{w}}~L\left(f\left(\mathcal{X}_t;\bm{w}, \bm{w}_{ext}\right)\right),
\end{split}
\end{equation}
where $L$ is an arbitrary loss function for representation learning (e.g., self-supervised losses~\cite{chen2020mocov2,grill20byol,zbontar2021barlow}) {and $\bm{w}_{ext}$ is an (optional) extra weights for additional structures not included in backbone weights $\bm{w}$, such as a decoder, a projection layer, and a predictor.} Since a direct comparison of the quality of representation models is intractable, recent representation learning literature validates obtained representation models by probing generic transferability on multiple downstream tasks. In a similar vein, prior UCL works~\cite{madaan2022rethinking, fini2022self} adopt supervised prediction tools like the KNN classifier and linear evaluation while keeping the learned backbone fixed. However, we argue that such evaluation paradigms cannot appropriately measure the transferability of representation on unseen tasks.

%% file: materials/figures/4_attn_visualization.tex
\begin{figure*}[t!]
    \small
    \centering
    \resizebox{0.9\textwidth}{!}{%    
    \hspace{-0.1in}
    \begin{tabular}{l@{\hspace{6pt}}cccc}
    \rotatebox[origin=c]{90}{\scriptsize Supervised}\hspace{-0.05in}    
    &
    \includegraphics[align=c, height=1.9cm]{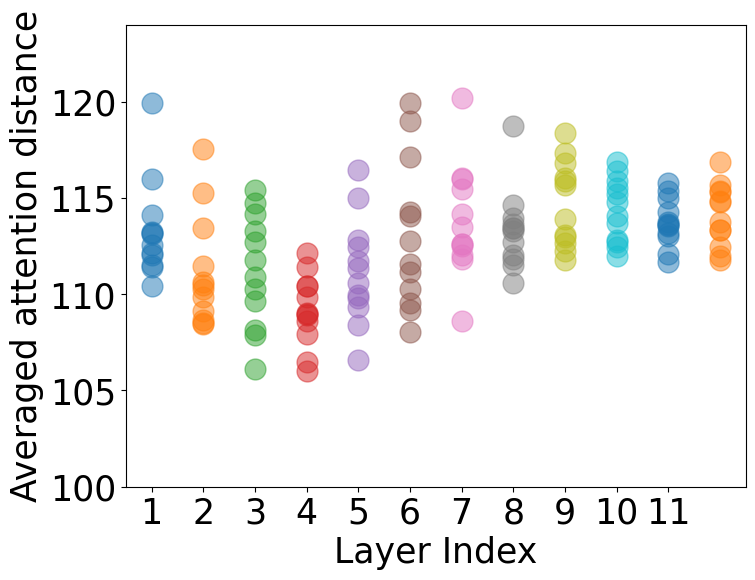}&\hspace{-0.15in}
    \includegraphics[align=c, height=1.9cm]{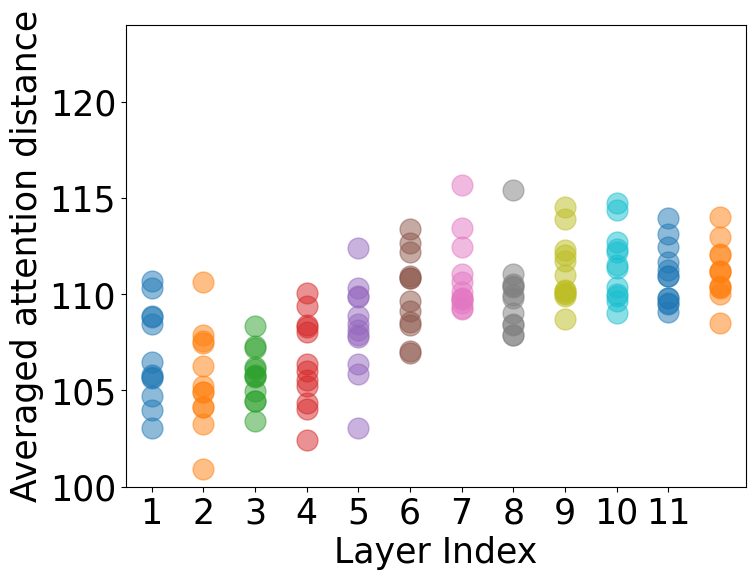}&
    \includegraphics[align=c, height=1.9cm]{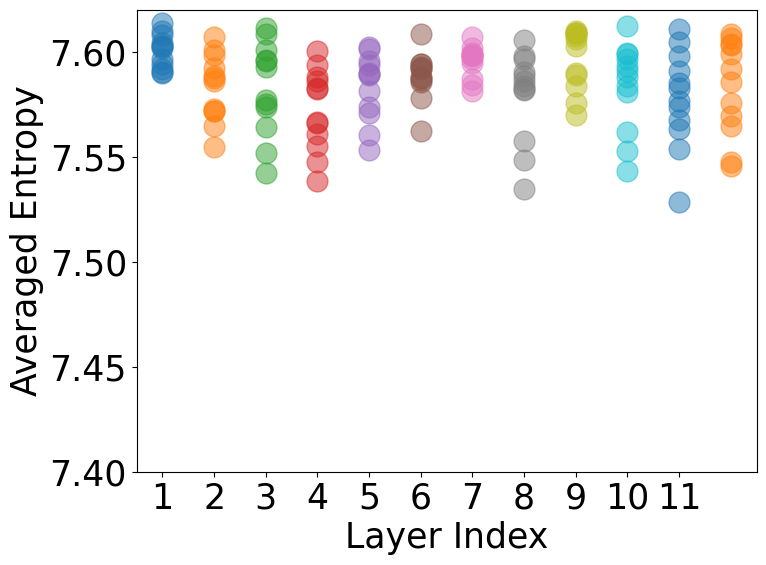}&\hspace{-0.15in}
    \includegraphics[align=c, height=1.9cm]{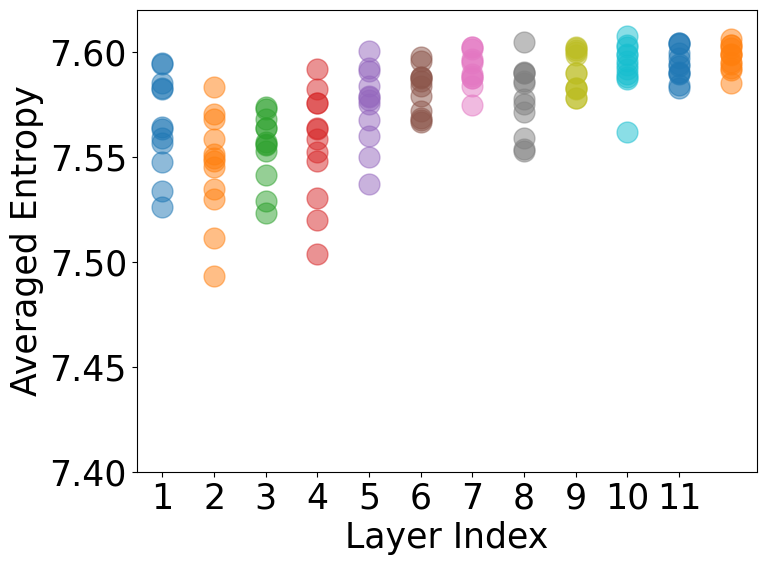}\\
    \rotatebox[origin=c]{90}{\scriptsize  SimSiam}\hspace{-0.05in}    
    &
    \includegraphics[align=c, height=1.9cm]{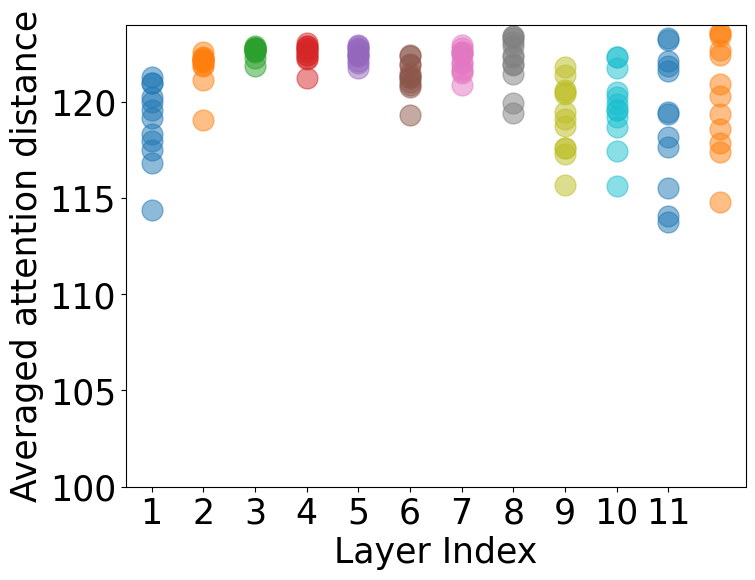}&\hspace{-0.15in}
    \includegraphics[align=c, height=1.9cm]{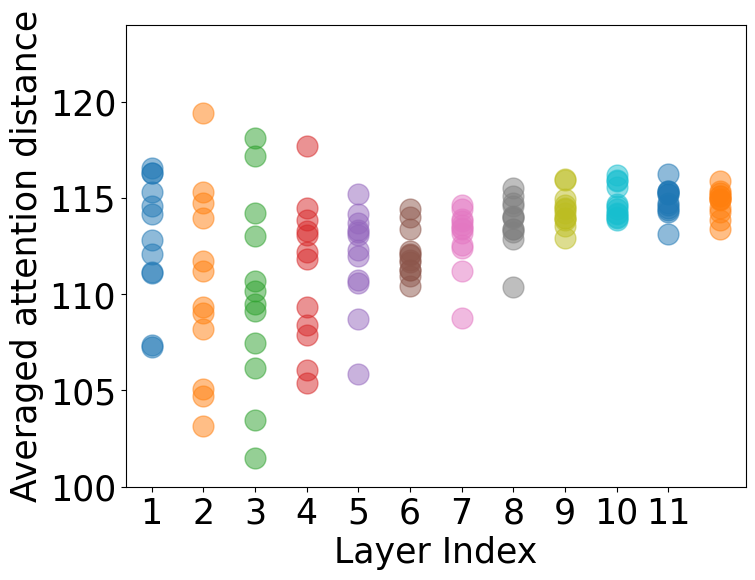}&
    \includegraphics[align=c, height=1.9cm]{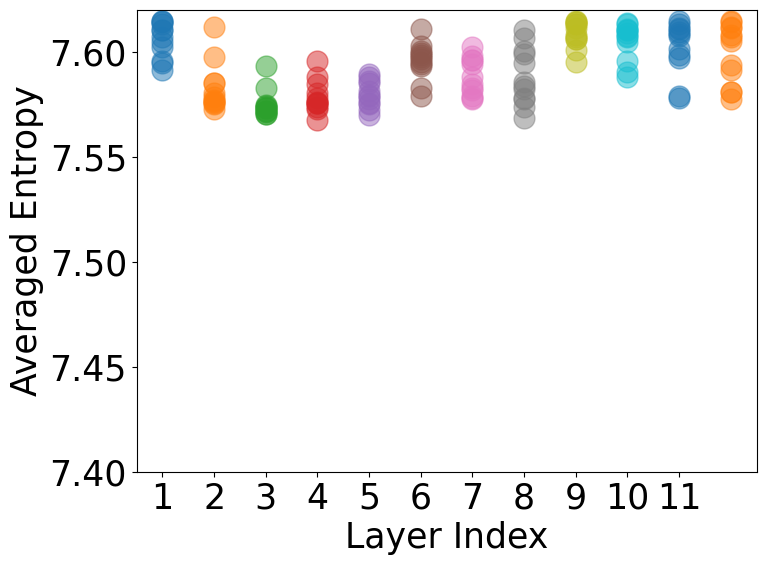}&\hspace{-0.15in}
    \includegraphics[align=c, height=1.9cm]{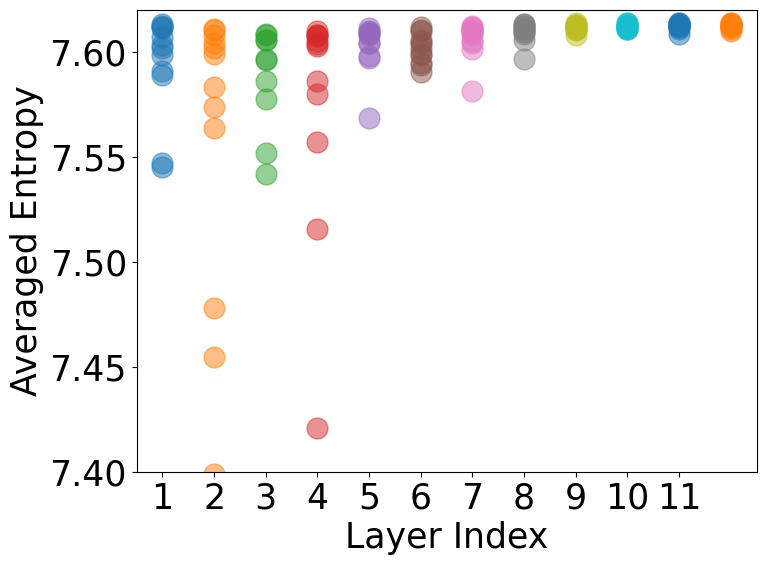}\\
    \rotatebox[origin=c]{90}{\scriptsize SimMIM}\hspace{-0.05in}    
    & 
    \includegraphics[align=c, height=1.9cm]{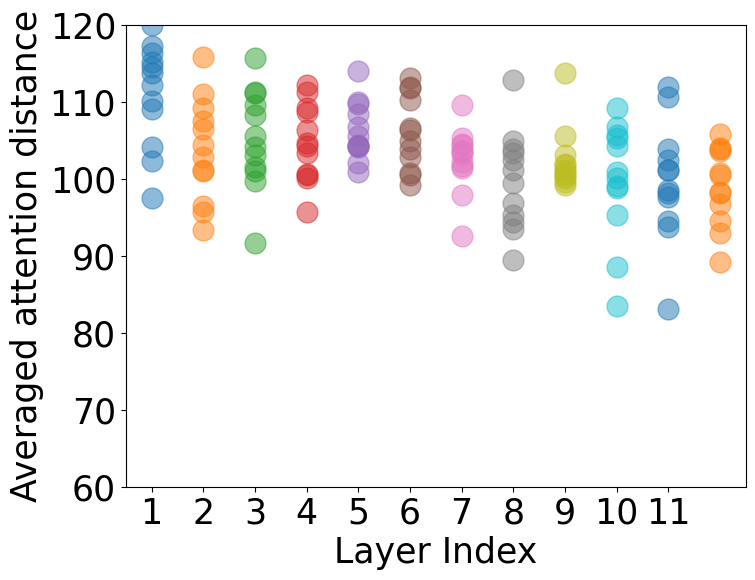}&\hspace{-0.15in}
    \includegraphics[align=c, height=1.9cm]{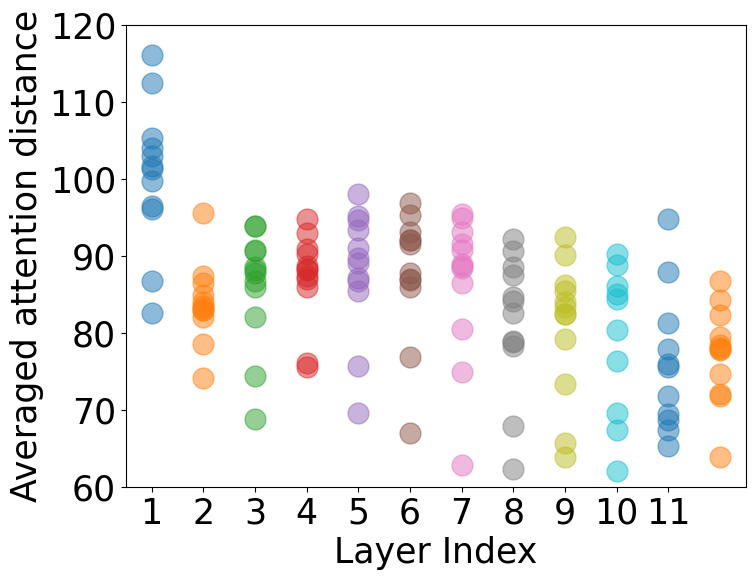}&
    \includegraphics[align=c, height=1.9cm]{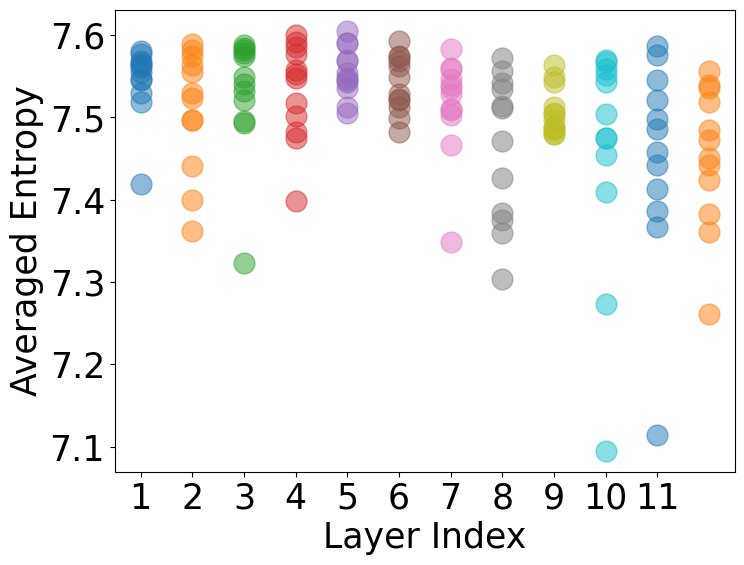}&\hspace{-0.15in}
    \includegraphics[align=c, height=1.9cm]{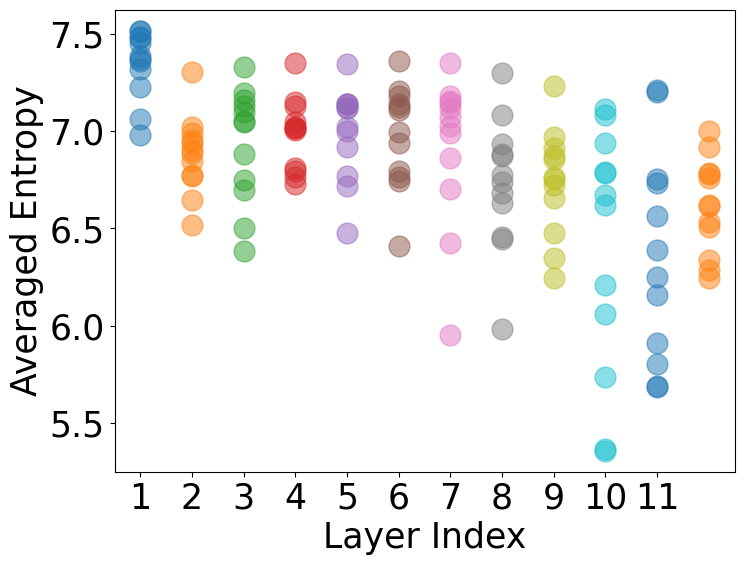}  \vspace{-0.05in}
    \\
    &
    \scriptsize $\bm{w}_{\bf{T0}}$ & \hspace{-0.15in}
    \scriptsize $\bm{w}_{\bf{T0 \rightarrow T8}}$ &\hspace{-0.1in}
    \scriptsize $\bm{w}_{\bf{T0}}$ &\hspace{-0.15in}
    \scriptsize $\bm{w}_{\bf{T0 \rightarrow T8}}$ 
    \end{tabular}}
    \hspace{0.15in} 
    \begin{tabular}{cccc}
    \multicolumn{2}{c}{\small (a) Attention Distance} \hspace{0.25\linewidth} &
    \multicolumn{2}{c}{\small (b) Attention Entropy}
    \end{tabular}
    % \vspace{-0.05in}
    \caption{\small \textbf{(a): Visualization of the attention distance} of out-of-distribution task (T9) with respect to three continual learning frameworks right after the completion of the first ($\bm{w}_{\bf{T0}}$) and last task ($\bm{w}_{\bf{T0\rightarrow T8}}$). \textbf{(b): Visualization of the entropy of each attention head's distribution} of unseen task T9. We use a ViT~\cite{dosovitskiy2020image} backbone for visualization.}
    \label{fig:attn-dist-and-entropy}
    % \vspace{-0.1in}
\end{figure*}

%% file: sections/4_method_details.tex
\section{Continual Learning for Incremental \\Model Generalization}

\subsection{The Role of Global and Local Attention during Continual Learning}\label{subsec:fg}
Prior continual learning literature has demonstrated that a model with a standard CL setting suffers from forgetting due to loss of local features and attention to past tasks. But, we argue that they have barely discussed the generalization of continually learned representation, which can be a great source for deploying an improvable foundation model by fine-tuning an unlimited number of tasks. We throw a question mark at this point:
\begin{center}
    \vspace{-0.05in}
    \emph{"So, is the model generalization getting worse as it goes through training sequential tasks?"}
    \vspace{-0.05in}
\end{center}

\input{materials/figures/ours_figure}

Surprisingly, we found that this is not the case as the generalization (a.k.a., transfer quality) is consistently getting improved during CL as shown in \Cref{fig:concept} Right. 
To explicate behaviors of multi-head attention using transformer backbones, similar to the experiment in~\Cref{fig:concept}, we train supervised and unsupervised CL frameworks on the ImageNet-1K split dataset without any specific CL methods. After that, we investigate changes in regional inductive bias by measuring the average distance of attention heads~(defined by \Cref{eq:adist}) in each layer, visualized as different points in \Cref{fig:attn-dist-and-entropy} \highlight{(a) top} and \highlight{middle row}. Supervised and contrastive self-supervised (\emph{SimSiam}) continual learning models focus more on strong locality inductive bias at lower layers (decreased attention distance) and global attention at higher layers (increased attention distance) during continual learning as these frameworks are innately designed to cluster/classify input features in deeper layers. However, focusing on global attention to capture task-specific features is undesirable for transferring the representations to out-of-distribution tasks. Next, we visualize the entropy conditioned solely on the distribution of each attention head in \Cref{fig:attn-dist-and-entropy} \highlight{(b)} by computing $-\sum_i\bm{a}_i\log(\bm{a}_i)$ for each attention head $\bm{a}$. Supervised and contrastive unsupervised frameworks broadly focus on most attention heads during continual learning at deeper layers. This indicates that they already substantially adapt to pre-trained tasks while losing a degree of freedom to transfer downstream tasks.

To build a new UCL framework for a better generalizable representation model across all layers, we survey Masked Image modeling (MIM)~\cite{pathak2016context, he2022masked} that self-trains input representation by minimizing regression loss to predict RGB pixel values in randomly zeroed patches in patchified input images. MIM enjoys locality inductive bias with diverse attention across layers, allowing better transfer ability to unseen tasks. 
%Our proposed UCL framework with \emph{SimMIM} obtains superior model generalization ability over other supervised and unsupervised counterparts, demonstrating our motivation about the relationship between transferability and global attention.

\subsection{Continual Self-supervised Learning with \\Masked Modeling}\label{subsec:cssl-mim}
We formulate a representational learner $f_{\bm{w}}$, composed of a neural encoder $h_{\bsy\theta}$ and a decoder $g_{\bsy\phi}$. We build backbones using Vision Transformer variants~\cite{dosovitskiy2020image,liu2021swin} due to their powerful generality and remarkable performance on high-resolution visual tasks. They are flexible to transfer the obtained representations to downstream tasks requiring various input image sizes in demands when existing UCL frameworks~\cite{madaan2022rethinking,rao19curl} allow the fixed image size for representation learning and fine-tuning since the architectures are basically composed of multi-layer perceptrons and convolutional neural networks.
At training $t$-th task with a training instance $\bm{x}_{t}\in\mathbb{R}^{C\times H\times W}\in\mathcal{X}_t$, a model segments $\bm{x}_{t}$ into smaller image patches where the width and height are $s <H, W$, and randomly zeros a fraction of image patches out with a fixed ratio $\tau$. An encoder tokenizes masked patches to the embedding space and fed into multiple self-attention blocks to capture latent representation features. A decoder reconstructs encoded features to approximate the input image. The objective is to minimize the following loss function for continual representation learning ($\|\cdot\|_\mu$ denotes any norm, often $\mu\in\{1,2\}$ is used). Let $K=\left\lfloor\frac{H}{s}\right\rfloor\cdot\left\lfloor\frac{W}{s}\right\rfloor$ be the number of tokens at each image, we formulate the loss function as follows:
\begin{equation}\label{eq:mim}
\begin{split}
\ell\left(\bm{x}_t; \bm{w}\right)&=\|f(\bm{m}*_s\bm{x}_t;\bm{w})-\bm{m}*_s\bm{x}_t\|_\mu\\
&=\|g\left(h\left(\bm{m}*_s\bm{x}_t;\bsy\theta\right);\bsy\phi\right)-\bm{m}*_s\bm{x}_t\|_\mu,\\
~~\text{where}&~~\bm{m}=\{0,1\}^{K}\sim B\left(K,\rho\right).
\end{split}
\end{equation}
With patch size $s$, $*_s$ denotes a patch-wise multiplication operation between a training instance $\bm{x}$ and a generated mask vector $\bm{m}$ drawn by the binary distribution $B$ with sparsity ratio $\rho$.
%$\circ$ is function composition, and 
$B(i,\rho)$ is a $i$ independent binary sampling with a ratio $\rho$ to pick $1$. The model updates a set of weights that predicts masked regions of input images, conditioning other available areas. We simply adopt $\ell_1$ regularization to minimize the distance between predicted patches and the targets, followed by earlier reconstruction-based works~\cite{xie2021simmim, he2022masked}, and after completing a sequential training, the obtained encoder $h_{\bsy\theta}$ can be utilized for many different downstream tasks. And we find that our new framework outperforms supervised and contrastive benchmarks in model transferability during continual pre-training (Please see~\Cref{fig:exp-anal}).
Interestingly, unsupervised continual learning with the masked autoencoder framework (\emph{SimMIM} in~\Cref{fig:attn-dist-and-entropy} \highlight{bottom row}) behaves very differently from the other two frameworks. Almost all layers have a diverse focus on locality, and this tendency becomes stronger as they continue to pre-train more tasks. 
%{(Please see~\Cref{fig:exp-anal} \highlight{Left})}. 
% we observe that the attention of SimMIM focuses on much broader over many tokens while the other two frameworks concentrate on a few tokens, resulting in narrow attention.

\subsection{Continual Pre-training via Global Attention Discretization}\label{subsec:cp}
Our aim is to utilize the backbone of the fine-tuned model as a pre-training checkpoint for another problem in a supervised manner. %Note that a fine-tuned model on $\mathcal{T}_{target}$ is also used for fine-tuning future tasks. 
Given the target task dataset $\mathcal{D}_{target}=\{\mathcal{X}, \mathcal{Y}\}$ and a classifier $\delta(\cdot; {\bm{u}})$ parameterized by $\bm{u}$, we formulate the objective of continual pre-training as follows:
\begin{equation}\label{eq:cp}
\begin{split}
% \bm{w}^*=\bm{w}^{(t+1)}=arg\min_{\bm{w}^{(t)}}\ell\left(f\left(\mathcal{X};\bm{w}^{(t)}\right), \mathcal{Y}\right).
\minimize_{\bm{w}^{(t)},\bm{u}^{(t)}}\ell\left(\delta\left(f\left(\mathcal{X};\bm{w}^{(t)}\right);\bm{u}^{(t)}\right), \mathcal{Y}\right),%\\\text{~~~~s.t.~~~~} \bm{w}^{(t)}=\widehat{\bm{w}}^{(t-1)}.
\end{split}
\end{equation}
We suppose each fine-tuning step independently introduces its own classifier. The formulation is aligned with the continual learning problem described in~\Cref{subsec:scl}, but this setting is about never-ending model generalization to achieve a consistently improved adaptation to the out-of-distribution task in the future. That is, obtained representation model for fine-tuning task $t$, $\widehat{\bm{w}}^{(t)}$, will be reused for future task training ($\bm{w}^{(t+1)}=\widehat{\bm{w}}^{(t)}$).

However, fine-tuning often reduces the general transferability to adapt to different tasks, demonstrating a suboptimal model generalization of supervised CL compared to our MIM-based framework (Please see \Cref{fig:exp-anal}~\highlight{Left}).
Motivated by our findings in \Cref{subsec:fg} and \Cref{subsec:cssl-mim}, we propose a new method for continual pre-training, named \emph{GLobal Attention Discretization (GLAD)}. Our proposed method preserves diverse degrees of averaged distance at each attention head to preserve transferable backbone weights for future problems while capturing task-adaptive features guided by GLAD modules. As illustrated in~\Cref{fig:ours}, a multi-head self-attention operation with adaptor weights $\bm{v}$. We transform the task-generic MSA features to input-dependent by propagating adaptor $\bm{v}$. Then, the model enables solving the current task problem while constraining the backbone weights to preserve abundant locality inductive bias. Let $\bm{a}^{l,i}$ be an averaged entropy of the attention passed over adaptor operation (\emph{dark dashed arrow}) from $i$-th head at layer $l$, the objective function of our GLAD is:
\begingroup\makeatletter\def\f@size{8}\check@mathfonts
\def\maketag@@@#1{\hbox{\m@th\large\normalfont#1}}
\begin{equation}\label{eq:ours}
\begin{split}
&\minimize_{\bm{w},\bm{v}}\sum^{N}_{n=1}\ell\left(f\left(\bm{x}^{(n)};\bm{w}, \bm{v}\right), \bm{y}^{(n)}\right),\\
&+\frac{1}{L}\sum^{L}_{l=1}\left(\left(\sqrt{E\left[\left(\bm{a}^{l,i}-\overline{\bm{a}}^{l}\right)^2\right]}+\epsilon\right)^{-1}
+\lambda\left\|\overline{\bm{a}}^l\right\|^2_F\right),
\end{split}
\end{equation}
\endgroup
where $E$ indicates the expectation, $\overline{\bm{a}}^{l}=\frac{1}{H^{l}}\sum^{H^{l}}_{i}\bm{a}^{l,i}$ at layer $l$ with the number of its attention head $H^l$, $\lambda$ is a scaling factor. $\epsilon$ is a small constant value. We jointly minimize the task loss with an additional regularizer that constrains the entropy variance of attention heads to sufficiently diverge at each layer as an average of their inverse standard deviation. We add to minimize a Frobenius norm for {$\overline{\bm{a}}$ to promote abundant locality inductive bias for backbone attention weights.} Note that our proposed method is robust to utilize any kind of multi-head self-attention modules, we demonstrate the efficacy in vanilla Vision Transformer~\cite{dosovitskiy2020image} and Swin Transformer~\cite{liu2021swin}.
The learned backbone weights $\bm{w}$ excepting classifier and GLAD-adaptors can be reused for fine-tuning future tasks. We describe the overall continual pre-training procedure with GLAD in~\Cref{alg:algorithm}.

\input{materials/algorithm}

%% file: materials/figures/ours_figure.tex
\begin{figure*}
    % \vspace{-0.2in}
    \footnotesize
    \centering
    \includegraphics[height=6.2cm]{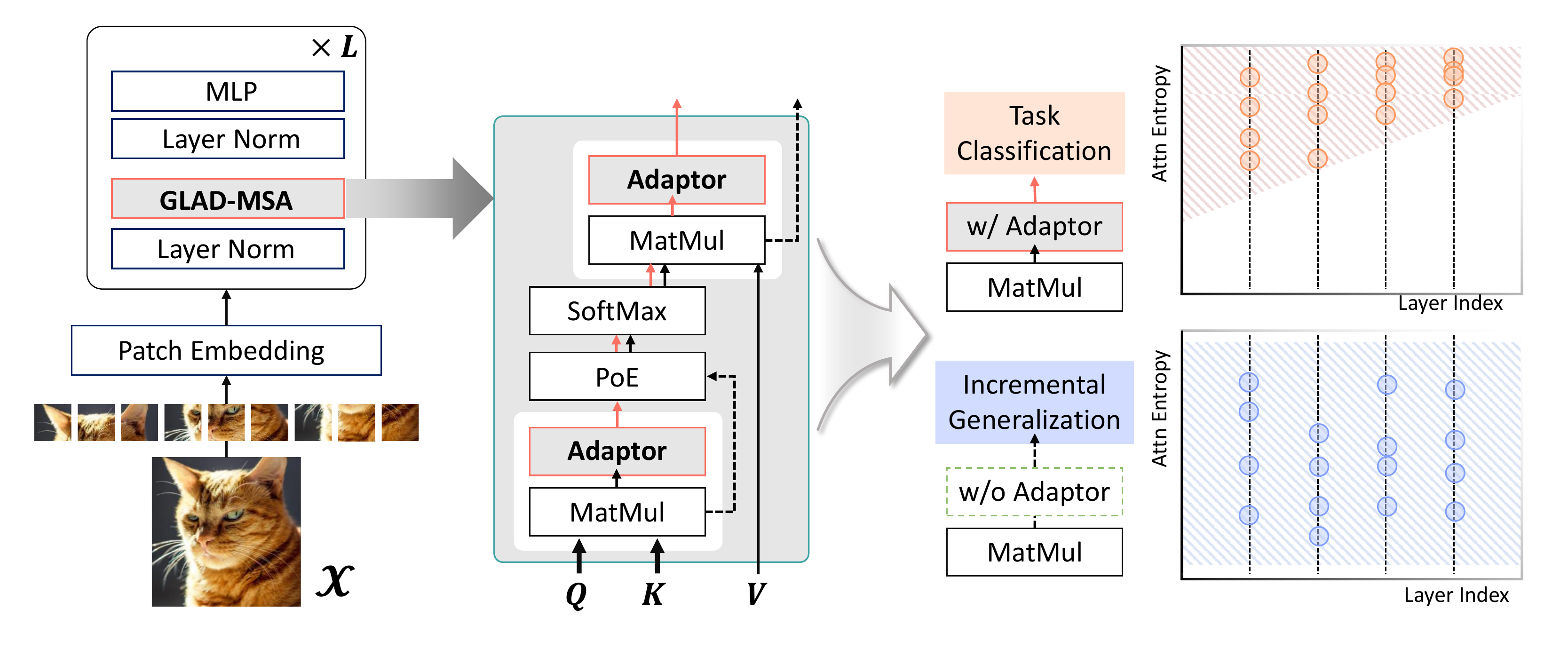}
    \vspace{-0.1in}
    \caption{\small \textbf{Illustration of the proposed GLobal Attention Discretization (GLAD) for Continual Pre-training.} Our GLAD introduces a new Multi-head Self-attention operation named GLAD-MSA with a parametric adaptor. The pre-trained model with adaptors fine-tunes given problems with a constraint that encourages divergence of attention entropy in each layer, leading to incremental positive transfer of backbone parameters for future tasks.}
    \label{fig:ours}
    \vspace{-0.1in}
\end{figure*}

%% file: materials/algorithm.tex
% \begin{minipage}[t!]{\linewidth}
\begin{algorithm}[H]
\caption{Continal Pre-training with GLAD}
\label{alg:algorithm}
\small
\begin{algorithmic}[1]
\INPUT A sequence of tasks $\{\mathcal{D}_1,\mathcal{D}_2,\cdots\}$, backbone network $f$, learning rate $\eta\in\mathbb{R}^{+}$, hyperparameter $\lambda$,\\ small constant $\epsilon$, initialization $\bm{w}_{\text{init}}, \bm{v}_{\text{init}}$
\FORALL{task $\mathcal{T}_t=\mathcal{T}_1, \mathcal{T}_2,\ldots$}
    \STATE Build a model $f_{\bm{w},\bm{v}}(\cdot)$ with GLAD-MSA \COMMENT{\Cref{fig:ours}}
    \STATE Initialize $\bm{w}\leftarrow\bm{w}^{*}$ excluding classifier \textbf{if} $f_{\bm{w}^{*}}$\textbf{exists,~otherwise} $\bm{w}\leftarrow\bm{w}_{\text{init}}$
    \STATE Initialize $\bm{v}\leftarrow diag(\bm{v}_{\text{init}})\coloneqq (1,\ldots, 1) \in \mathbb{R}^{d_{\text{out}}}$
    \FOR{batch $\bm{x}_n, \bm{y}_n \sim \mathcal{D}_t$}
    \STATE $\mathcal{L} = \ell\left( f\left( \bm{x}_n;\bm{w},\bm{v} \right), \bm{y}_n \right)+$\\
    $\frac{1}{L}\sum^{L}_{l=1}\left(\sqrt{E\left[\left(\bm{a}^{l,i}-\overline{\bm{a}}^{l}\right)^2\right]}+\epsilon\right)^{-1}$
    \COMMENT{\Cref{eq:ours}}
        % \STATE $\bm{w}\leftarrow\bm{w}-\eta\pdv{\mathcal{L}}{\bm{w}}$    
        \STATE $\bm{w} \leftarrow \nabla_{\bm{w}}\mathcal{L}$, $\bm{v} \leftarrow \nabla_{\bm{v}}\mathcal{L}$
        % \STATE $\bm{v} \leftarrow \nabla_{\bm{v}}\mathcal{L}$
    \ENDFOR
    \STATE $\bm{w}^*\leftarrow\bm{w}$ %for future use
\ENDFOR
\end{algorithmic}
\end{algorithm}
\vspace{-0.2in}

%% file: sections/5_experiments.tex
\section{Experiments}
We conduct the experiments on various continual learning frameworks with and without supervision using ImageNet-1K Split dataset against multiple strong baselines. 
for unsupervised continual learning that demonstrates the effectiveness of our proposed method on fine-tuning performance on downstream tasks.%, such as classification, object detection, and semantic segmentation.
\subsection{Architectures and Baselines}

\input{materials/tables/main_table}

\input{materials/figures/5_main_figure.tex}

% \paragraph{Architectures and baselines} 
We use ViT~\cite{dosovitskiy2020image} and Swin Transformer~\cite{liu2021swin} as backbone architectures. We follow Siamese network structure by \citet{madaan2022rethinking} and implement a MIM-based continual self-supervised learning framework under SimMIM~\cite{xie2021simmim} and MAE~\cite{he2022masked} for UCL. CURL~\cite{rao19curl} is one of the pioneer works on unsupervised continual learning literature, but it is not scalable for high-resolution visual images by design. We utilize several CL methods: SI~\cite{zenke2017continual}, DER~\cite{buzzega2020dark}, and LUMP~\cite{madaan2022rethinking}. We further describe details, including hyperparameter setups in~\Cref{sup:sec:details}.

\paragraph{Datasets}
We use ImageNet-1K~\cite{deng2009imagenet} and CIFAR-100 dataset~\cite{krizhevsky2009learning} by splitting it into ten tasks where each task contains $100$ and $10$ classes, respectively. In~\Cref{tab:main-table}, we construct a sequential dataset with nine earlier tasks and assign the last one as a downstream (validation) task with fine-tuning.
Additionally, we split CIFAR-100 into five tasks for the continual pre-training experiment.% by evaluating the OOD fine-tuning performance through CIFAR-10.

\input{materials/figures/5_cf_attn_anal}
\input{materials/figures/5_experiments_anal.tex}
\subsection{Experimental Results}
\paragraph{Evaluation performance during continual learning} We validate the transfer quality of representation of continual learning models through the first task (T0)'s evaluation in~\Cref{tab:main-table}. We measure the change in the linear evaluation and fine-tuning performance. The evaluation of T0 from full pre-trained models over the Imagenet-1K and -22K datasets obtains high validation accuracy on fine-tuning and linear evaluation as they train on entire datasets.
Fine-tuning the base continual learning models, which perform a simple CL strategy without additional methods during training task sequences, achieves performance increases in T0 as they pre-trained longer task sequences, obtaining positive values in backward transfer. In ImageNet 1K Split, the results are similar to all supervised and unsupervised continual learning frameworks, including Contrastive Self-supervised Learning~\cite{madaan2022rethinking}- and Masked Image Modeling-based UCL (Please see \Cref{subsec:cssl-mim}). We also performed multiple continual learning methods, such as SI, DER, and LUMP. We found that they follow consistent tendencies according to the CL frameworks when LUMP with masked image modeling gains the highest accuracy on T0 with the strong backward transfer. On the other hand, the model degrades the linear evaluation performance in supervised continual learning, which had to do with catastrophic forgetting reported in conventional CL scenarios. SI and DER achieve increased final performance since the model mitigates the weight drift preserving the task-specific features on learned tasks. 
In CIFAR-100 Split, 
% we observed that our proposed masked modeling-based UCL framework consistently surpasses supervised CL in terms of the fine-tuning performance on OOD task (T9) over all continual learning methods. 
% \TBD{Interestingly, our reconstruction-based UCL framework also achieves competitive linear evaluation performance with alternative CL frameworks, unlike ImageNet. We expect this is because low-resolution visual tasks have a lower discrepancy between task-general and task-adaptive features due to their limited representativeness compared to high-resolution visual tasks. %We will further include qualitative analyses similar to Figure 2 in our final revision, which will provide a clearer picture of the reduced difference in scales for task-generic and task-adaptive features in low-resolution tasks.
we similarly observed that fine-tuning the in-distribution task (i.e., the first task) of Base/SI/LUMP with reconstruction-based UCL framework achieves higher and positive BwT than the case of CL methods in a supervised manner, demonstrating that supervised CL suffers more severe forgetting of task-discriminative information at deeper layers, unlike reconstruction-based UCL focusing on task-generic features over all layers. Note that the linear evaluation on CIFAR-100 seems more robust to forgetting than ImageNet experiments. We expect that these relative benefits in forgetting came from the shallower data distribution space and simpler visual features of CIFAR-100 compared to ImageNet.

\paragraph{Analyses for an Out-Of-Distribution (OOD) task}
% We denote the last task (T9) as an out-of-distribution problem excluding the continual-pretraining phase. In~\Cref{fig:mainfig}, we visualize the top-1 validation accuracy on T9
We denote the last task (T9) as an out-of-distribution problem, excluding it from the continual task sequence. In~\Cref{fig:mainfig}, we visualize the top-1 validation accuracy on an OOD task over three continual learning frameworks w/ and w/o SI. Similar to in-distribution evaluation, the MIM-based UCL method achieves higher fine-tuning performance both on the base model and SI. The linear probe performance of Supervised CL surpasses unsupervised counterparts, and we expect that representation from supervised learning contains directly helpful features to classify the high-resolution and complex task problems even without the re-update of backbone weights. In contrast, MIM remarkably underperforms on linear evaluations due to its characteristic property; Masked Modeling focuses on capturing global attention rather than the local one, providing a better generalization ability to unseen tasks. But its linear evaluation without fine-tuning the weights is inadequate to solve the problem as they contain little task-discriminative information.

To understand how CL frameworks exhibit incremental model generalization and fine-tuning performance, we analyze the behavior of layer attention during continual learning using the Swin-T backbone. In \Cref{fig:swin-attn-div}, we visualize the layer-by-layer changes in aggregated attention distance for T9 while the model trains ImageNet-1K-Split sequentially until the penultimate task (T0$\rightarrow$T8). Interestingly, aggregated attention distance significantly decreases the scale and increases the diversity across attention heads. This demonstrates that the continual learner captures richer task-general (or low-level) features behaving with more local attention (i.e., lower attention distance), which retains localized information with strong local inductive bias, such as edges, patterns, and textures. In Supervised and Contrastive Continual Learning frameworks, lower layers tend to drastically change toward capturing task-generic attention, and it is also coincident with well-known observations that lower layers in neural networks are more concerned with low-level features. Also, Masked Image Modeling (SimMIM) results demonstrate the salient effectiveness of capturing task-generic attention compared with the other two frameworks.
% \TBD{This is because masked modeling-based UCL continuously trains on more improved generic representations, evident in~\Cref{fig:swin-attn-div} that continual learners gradually capture richer task-general (or low-level) features behaving with more local attention (i.e., lower attention distance), which retains localized information with strong local inductive bias, such as edges, patterns, and textures.}

\paragraph{Freezing the partial layer weights during continual learning} 
We further analyze the effect of the layers for incremental generalization during continual learning in~\Cref{fig:exp-anal} \highlight{right}. After supervised/unsupervised training of the first task (\emph{After T0}), we freeze the two lowest or two deepest layer weights during the successive continual learning up to the final task (\emph{After T8}). For the MIM-based UCL framework, we use MAE with a ViT-B backbone. In supervised learning, both partial gradient update policies reduce the degree of incremental generalization during continual learning. It significantly reduces the representation model's fine-tuning performance compared to the fully-trained model. However, interestingly, prohibiting the update of layer weights at both ends less affects MIM-based unsupervised continual learning. We expect that this property comes from its flexibility in learning diverse attention across all layers. For further analyses, please see~\Cref{sup:subsec:analyses-freeze}.

\input{materials/figures/5_glad_cifar100_figure}
\input{materials/tables/5_ours_table_one_column.tex}

\paragraph{Improving model generalization via Global Attention Discretization}
We now validate our proposed method, \emph{Global Attention Discretization (GLAD)}, which encourages incremental model generalization during supervised continual pre-training. As discussed earlier, supervised training tends to focus on task-specific attention at deeper layers. That is, the model is hard to stray far from the weight space of the limited locality inductive bias, which is evident in the slower movement of averaged attention distance from supervised continual pre-training compared to the SimMIM-based (Please see \Cref{fig:swin-attn-div}) and results in suboptimal adaptation to arriving tasks. In \Cref{tab:glad-results-one}, we report the fine-tuning performance during continual pre-training. Note that to see the effect of MIM-based continual pre-training, we first perform continual pre-training over the earlier five tasks from ImageNet-1K Split under supervised and MIM-based unsupervised continual learning. Next, the pre-trained models fine-tune the remaining tasks in a sequential manner. We adopt SimMIM and Masked AutoEncoder (MAE) to understand general behaviors in MIM frameworks during unsupervised continual learning. Our proposed GLAD achieves significant gains in the performance of each task during sequential full-finetuning over different pre-trained initialization from supervised and unsupervised learning. {In~\Cref{fig:glad-results-cifar}, we visualize that our proposed GLAD consistently performs well in view of the generalization using the in-distribution task (CIFAR-100) during continual pre-training phase.}
% We believe that this result would clearly address the Reviewer's concern "Experiments are limited. One dataset is considered, and generalization is estimated on the last task.".}

%% file: materials/tables/main_table.tex
\begin{table*}[t]
\small
\centering
\setlength{\tabcolsep}{3pt} % 
\resizebox{0.93\textwidth}{!}{
\begin{tabular}{l@{\hspace{6pt}}crrrrrr}
\toprule
\multicolumn{2}{c}{\textsc{ImageNet 1K Split (T=10)}}&\multicolumn{2}{c}{\textsc{Supervised}}&
\multicolumn{2}{c}{Contrastive~\cite{chen2021exploring}}%\hspace{1in}
&
\multicolumn{2}{c}{Masked Model~\cite{xie2021simmim}}\\
\midrule
&
& \textsc{Final Acc} & \textsc{Neg. BWT}
& \textsc{~~~~~~~Final Acc} & \textsc{Neg. BWT}
& \textsc{~~~~~~~Final Acc} & \textsc{Neg. BWT}\\
\midrule
\multirow{6}{*}{\rotatebox[origin=c]{90}{\small Fine-tuning (FT)}}
% for t0
& \textsc{1K Pretrained} & 87.48 / 98.08 & $-$ & $-$ & $-$  & $-$ & $-$ \\
& \textsc{22K Pretrained} & 87.76 / 98.48 & $-$ & $-$ & $-$  & $-$ & $-$ \\
\cmidrule{2-8}
& \textsc{Base Model} & 
{71.90} / {90.64} & {~~{6.88} / ~~3.70} & 
{64.38} / {86.18} & {~~{7.18} / ~~4.56} & % ep 100, 0.2-0.5
{73.18} / {91.76} & {13.26} / ~~{7.46} \\ 
& \textsc{Si}~{\small \cite{zenke2017continual}} & 
{{70.00} / 90.38} & {~~{7.40} / ~~4.52} & 
{{61.46} / 84.82} & {~~{4.15} / ~~2.42} &  % 60eps 0.5-0.5
{71.54} / 90.76 & {11.92} / ~~6.58 \\ % 5-2
& \textsc{Der}~{\small \cite{buzzega2020dark}} & 
{70.57} / 90.12 & {~~8.94} / ~~4.86 & 
{62.37} / 85.46 & {~~9.28} / ~~6.08 & 
{70.10} / 90.10 & {19.55} / {12.24} \\ % 5-5

& \textsc{LUMP}~{\small \cite{madaan2022rethinking}} & 
N/A & N/A &
{64.01} / 86.42 & {~~6.24} / ~~4.32 & 
\textbf{75.11} / \textbf{92.38} &
\textbf{21.28} / \textbf{12.42} \\
\midrule
\multirow{6}{*}{\rotatebox[origin=c]{90}{\small Linear Probe (LP)}}
& \textsc{1K Pretrained} & 87.48 / 97.98 & $-$ & $-$ & $-$  & $-$ & $-$ \\
& \textsc{22K Pretrained} & 86.53 / 98.06 & $-$ & $-$ & $-$  & $-$ & $-$ \\
\cmidrule{2-8}

& \textsc{Base Model} & 
% T0
{33.66} / 62.20 & {~-5.98} / ~-4.30 & 
{17.10} / 40.60 & \textbf{~~7.64} / \textbf{13.94} & % 0.2-0.5
{17.46} / 40.60 & {~~4.76} / ~~6.36 \\ % 5-10

& \textsc{Si}~{\small \cite{zenke2017continual}} & 
\bf{34.82} / \bf{63.18} & {~-6.18} / ~-5.56 &  % pretrain lrm 1 - 0.5
15.24 / 36.76 &  ~-1.42 / ~-2.38 & % 0.2 - 20
14.92 / 37.80 & ~~4.82 / ~~8.26 \\ % 5 - 5
& \textsc{Der}~{\small \cite{buzzega2020dark}} & 
{34.59} / 62.29 & {~-5.86} / ~-6.16 & 
{14.84} / 36.13 & {~~3.68} / ~~5.22 & 
{~~6.22} / 21.52 & {~-0.84} / ~-1.04 \\ % 5 - 5
& \textsc{LUMP}~{\small \cite{madaan2022rethinking}} & 
N/A & N/A &
{18.50} / 42.05 & {~~7.54} / 11.38 & 
{19.26} / 43.21 & {~~7.38} / 11.22 \\
\bottomrule
% \end{tabular}
% }
\vspace{-.05in}\\
% \resizebox{0.93\textwidth}{!}{
% \begin{tabular}{l@{\hspace{6pt}}crrrrrr}
\toprule
\multicolumn{2}{c}{\textsc{CIFAR-100 Split (T=10)}}&\multicolumn{2}{c}{\textsc{Supervised}}&\multicolumn{2}{c}{Contrastive~\cite{chen2021exploring}}%\hspace{1in}
&\multicolumn{2}{c}{Masked Model~\cite{xie2021simmim}}\\
\midrule
&
& \textsc{Final Acc} & \textsc{Neg. BWT}
& \textsc{~~~~~~~Final Acc} & \textsc{Neg. BWT}
& \textsc{~~~~~~~Final Acc} & \textsc{Neg. BWT}\\
\midrule
\multirow{3}{*}{\rotatebox[origin=c]{90}{\small FT}}

& \textsc{Base Model} & 
{79.3} & {3.8} & 
{49.3} & {12.6} & 
\textbf{88.9} & {18.3} \\ 

& \textsc{Si}~{\small \cite{zenke2017continual}} & 
{78.0} & {5.8} & 
{57.3} & {14.5} & 
{86.4} & \textbf{20.0} \\ 
& \textsc{LUMP}~{\small \cite{madaan2022rethinking}} & 
N/A & N/A &
{83.3} & {16.5} &
\textbf{88.6} & {18.7} \\
\midrule
\multirow{3}{*}{\rotatebox[origin=c]{90}{\small LP}}
& \textsc{Base Model} & 
{70.3} & {0.0} & 
{45.7} & {5.9} &
{73.0} & \textbf{11.1} \\

& \textsc{Si}~{\small \cite{zenke2017continual}} & 
{69.0} & {0.9} & 
{49.3} & {3.0} &
{68.4} & {9.1} \\
& \textsc{LUMP}~{\small \cite{madaan2022rethinking}} & 
N/A & N/A &
{73.6} & {9.8} & 
\textbf{77.1} & \textbf{11.5} \\
\bottomrule
\end{tabular}
}
\vspace{-0.05in}
\caption{\small {\bf Fine-tuning and linear evaluation performance with their negative backward transfer} of the first task on ImageNet 1K and CIFAR-100 Split after supervised/unsupervised continual learning. We report the Top-1/Top-5 performance for all individual experiments on ImageNet and the Top-1 performance on CIFAR-100. Higher is better for both metrics, and the best results are highlighted in {\bf bold}. \label{tab:main-table}}
\vspace{-0.1in}
\end{table*}

%% file: materials/figures/5_main_figure.tex
\begin{figure*}%[h!]
    \centering
    \resizebox{0.9\textwidth}{!}{%
    \begin{tabular}{cccc}
    \includegraphics[height=3.cm]{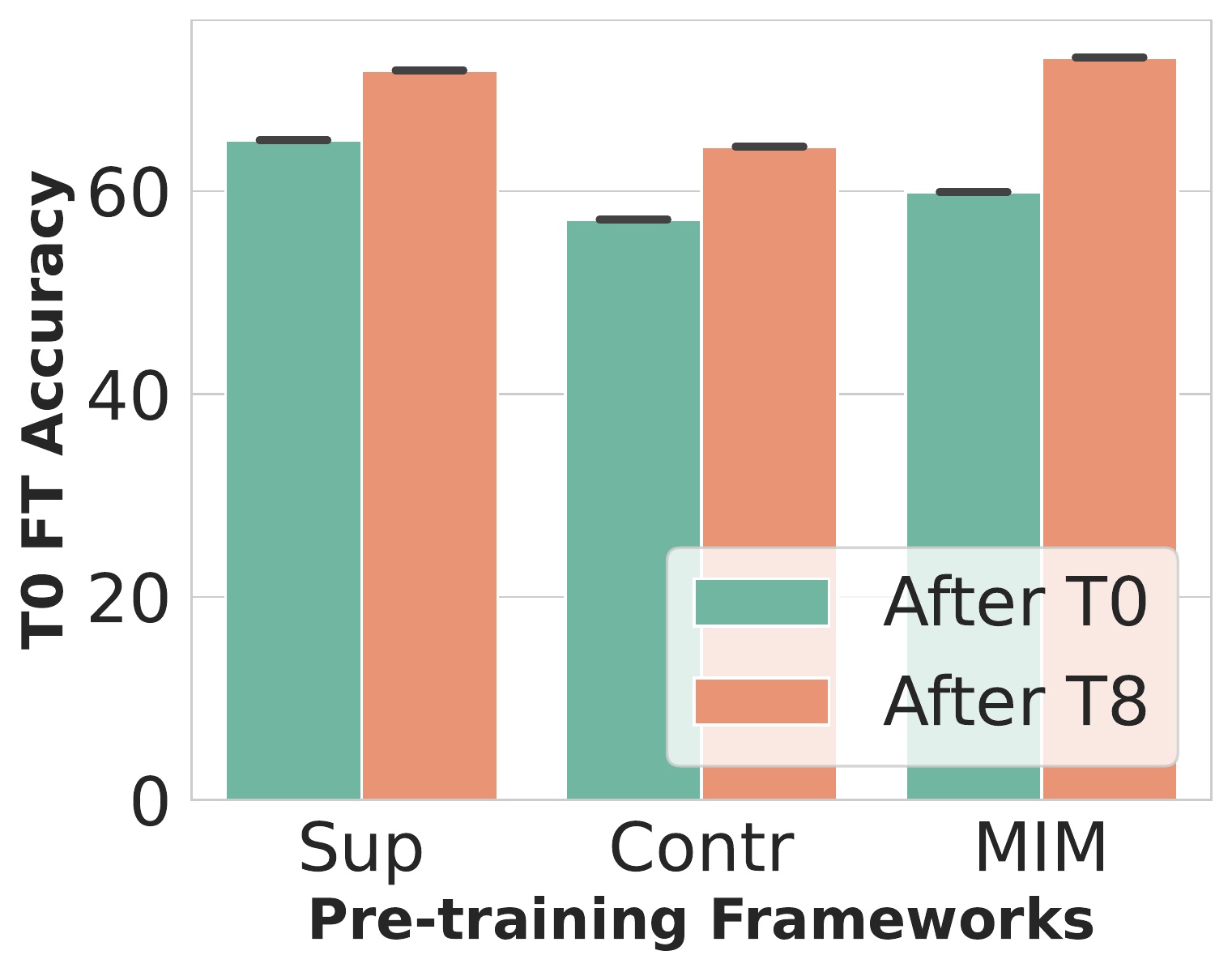}&\hspace{-0.1in}
    \includegraphics[height=3.cm]{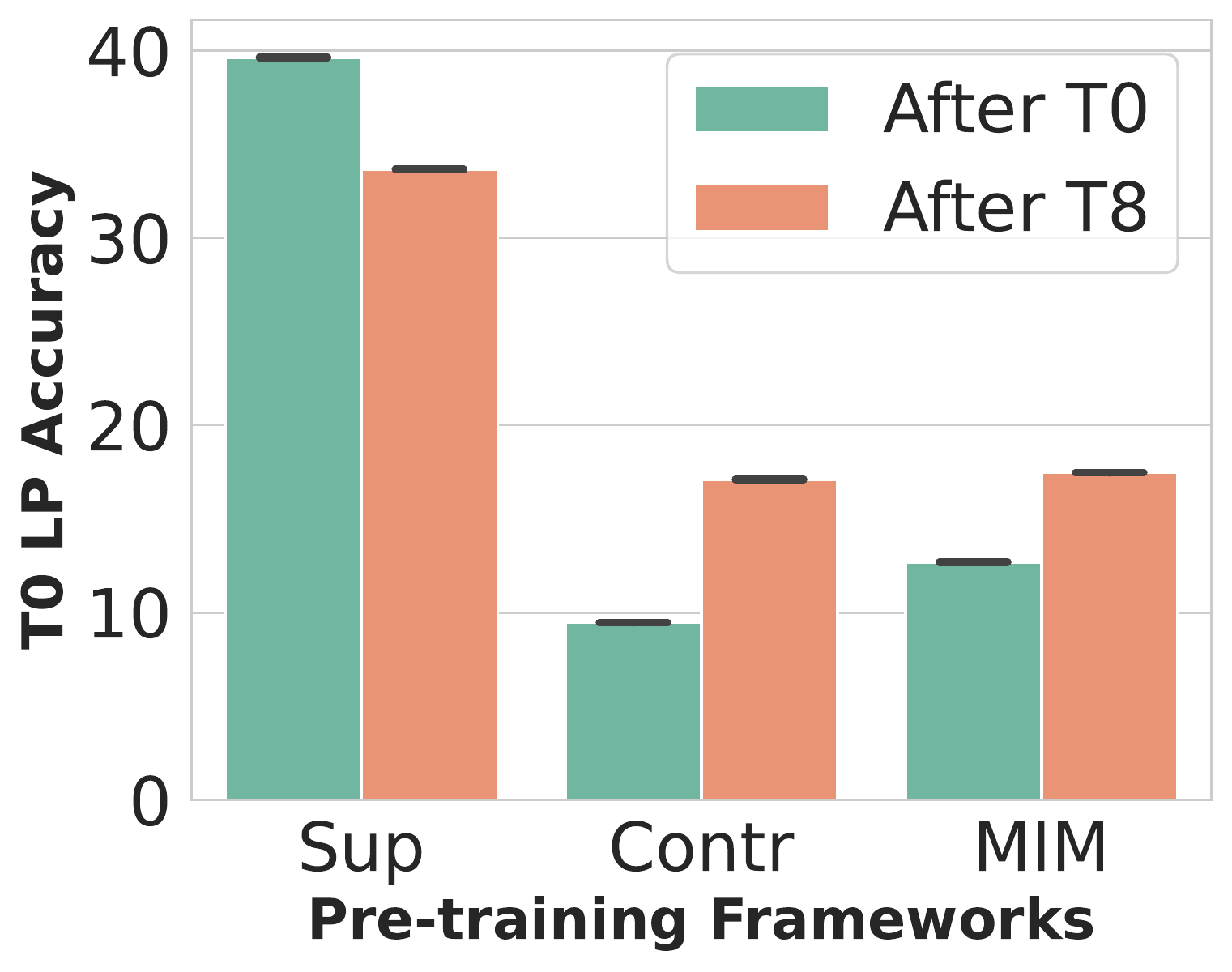}&\hspace{0.1in}
    \includegraphics[height=3.cm]{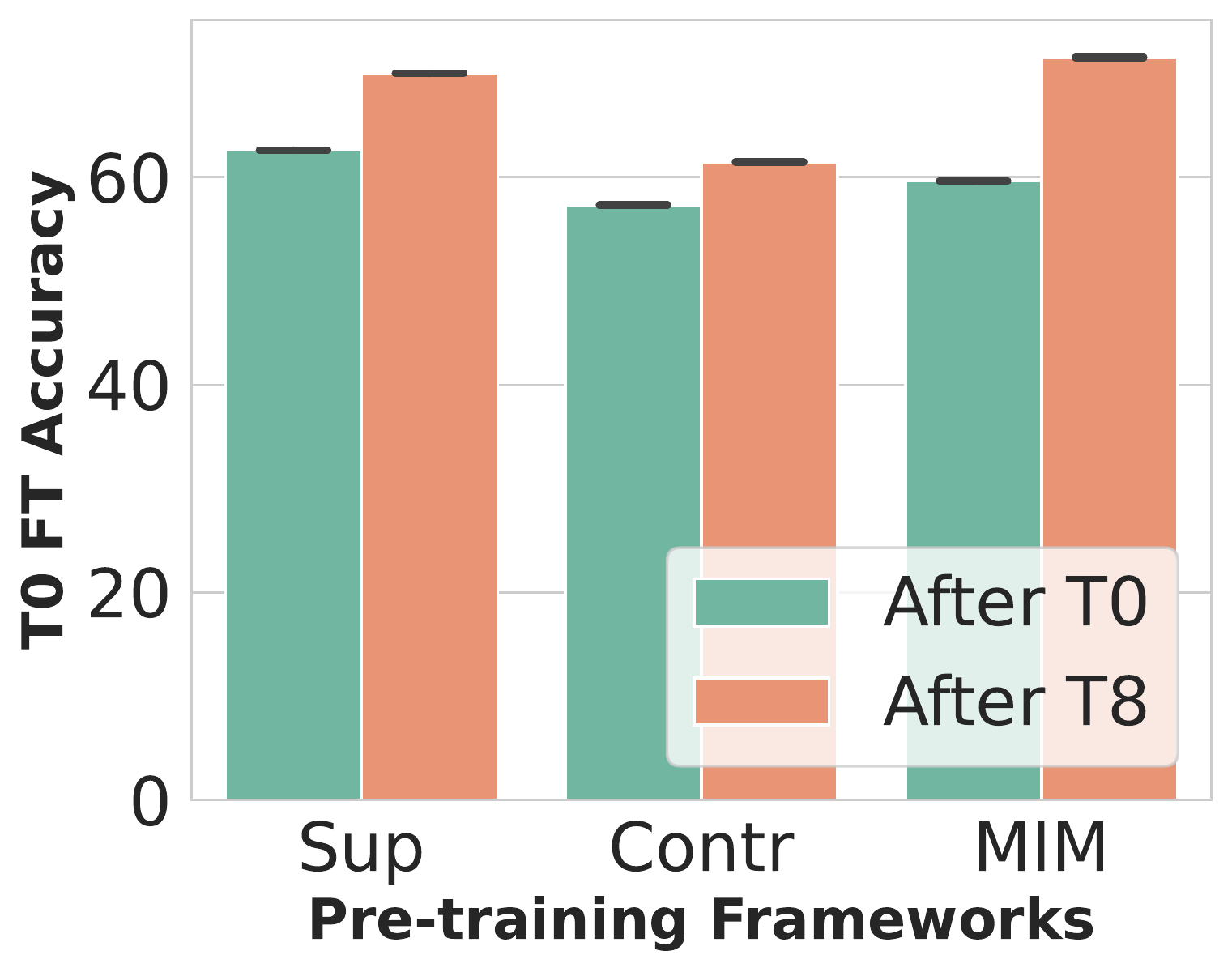}&\hspace{-0.1in}
    \includegraphics[height=3.cm]{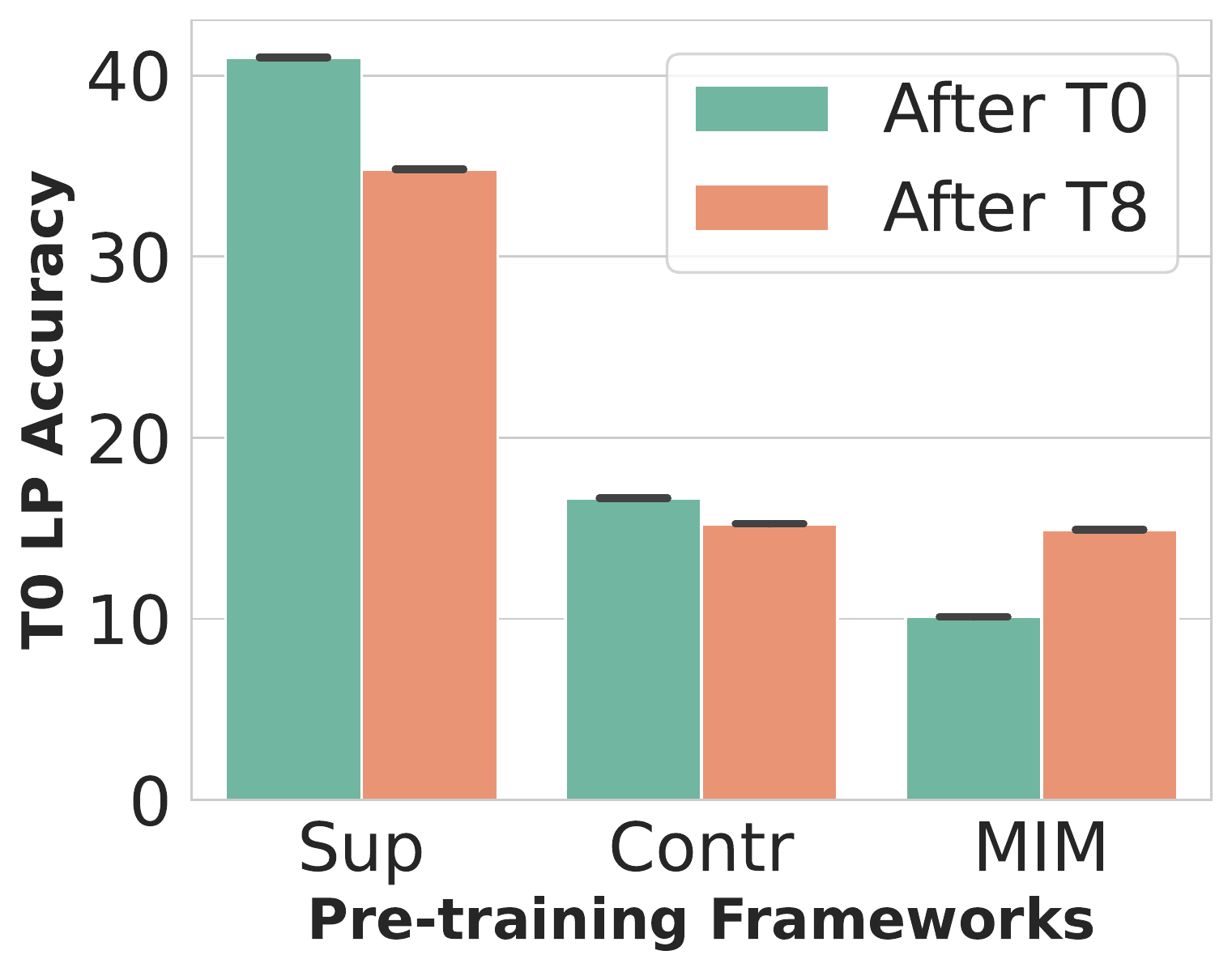}\\
    % \multicolumn{2}{c}{\footnotesize  (a) Finetuning} & \multicolumn{2}{c}{\footnotesize  (b) Linear Probe} \\
    \bf ~~(a) FInetuning - Base &\hspace{-0.1in}
    \bf ~~(b) Linear Probe - Base &\hspace{0.1in}
    \bf ~~(c) FInetuning - SI &\hspace{-0.1in}
    \bf ~~(d) Linear Probe - SI \\
    \end{tabular}}
    \vspace{-0.1in}
    \caption{\small \textbf{(a-b): Model transferability on ImageNet 1K Split.} We compare the fine-tuning accuracy on the OOD task (\emph{T9}) after pre-training the first task (\emph{After T0}) with the performance after pre-training on nine sequential tasks (\emph{After T8}). We continually pre-train with Supervised learning (Sup), Contrastive Self-supervised learning (Contr), and Masked Image Modeling (MIM). \textbf{(c-d): Same visualization with a CL method}, \emph{SI}, during pre-training.}
    \label{fig:mainfig}
    \vspace{-0.1in}
\end{figure*}

%% file: materials/figures/5_cf_attn_anal.tex
\begin{figure*}%[h!]
    % \vspace{-0.2in}
    \footnotesize
    \centering
    \resizebox{0.9\textwidth}{!}{%
    % \hspace{-0.2in}
    \begin{tabular}{ccc}    
    \includegraphics[align=c, height=3cm]{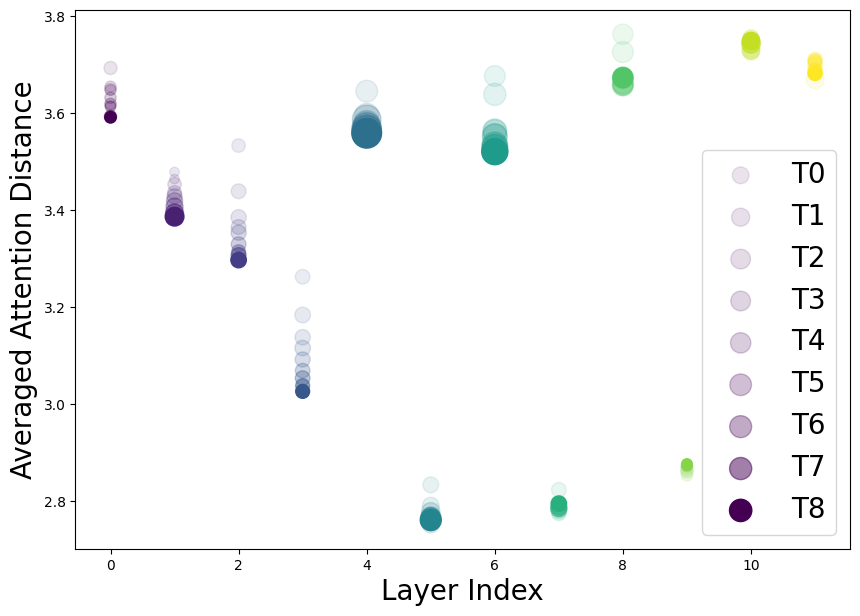}&%\hspace{-0.2in}
    \includegraphics[align=c, height=3cm]{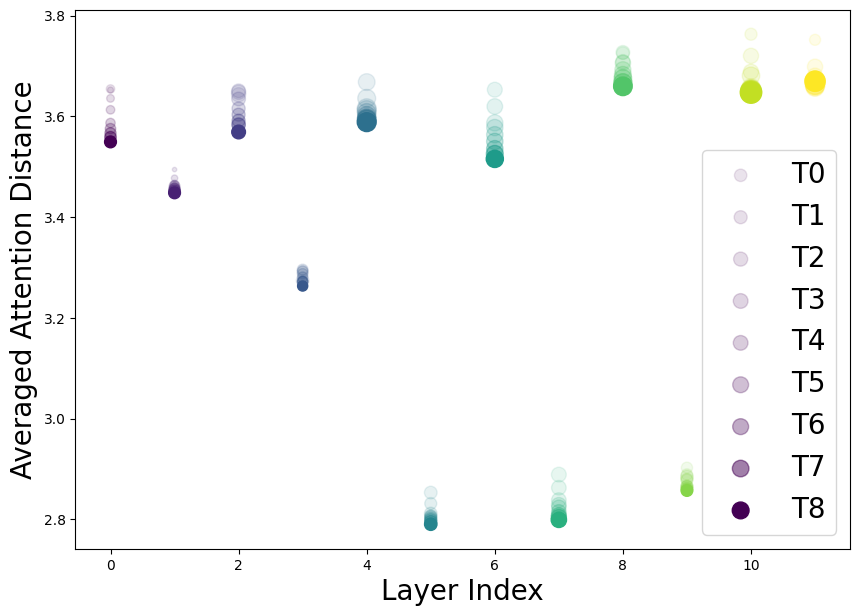}&%\hspace{-0.2in}
    \includegraphics[align=c, height=3cm]{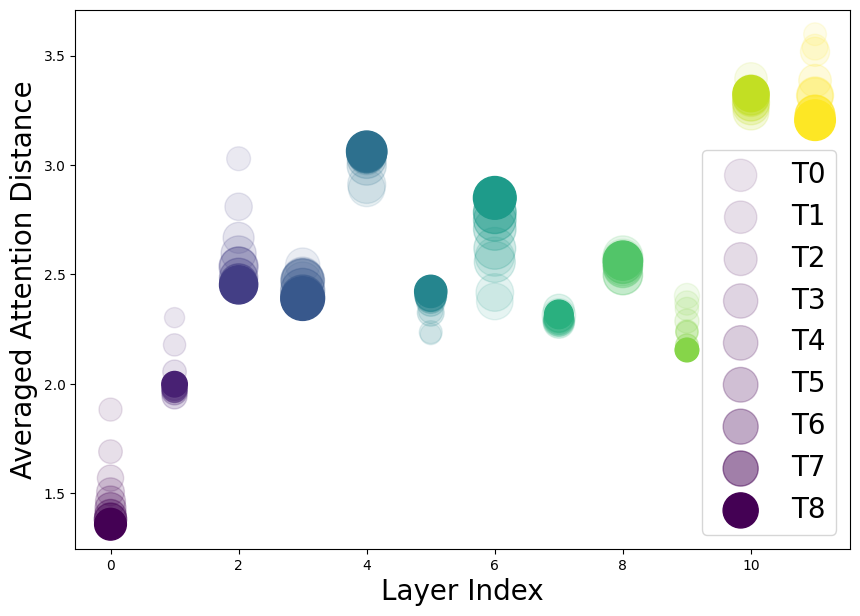}%    \vspace{-0.1in}
    \\    
    % \includegraphics[align=c, height=3cm]{figures/attn_div/sup_entr_diversity_attt9_v2.png}&
    % \includegraphics[align=c, height=3cm]{figures/attn_div/siam_entr_diversity_attt9_v2.png}&
    % \includegraphics[align=c, height=3cm]{figures/attn_div/mim_entr_diversity_attt9_v2.png}
    % \\
    \bf{Supervised} &
    \bf{SimSiam} &
    \bf{SimMIM} \\
    \end{tabular}}
    \vspace{-0.1in}
    \caption{\small \textbf{Visualization of aggregated attention distance} on an OOD task (T9) at each layer at the end of each continual learning task phase (T0$\rightarrow$T8). The radius of marker indicates the standard deviation over attention heads in the corresponding layer.}% \textbf{Bottom row: Averaged attention entropy} on T9.}
    \label{fig:swin-attn-div}
    \vspace{-0.1in}
\end{figure*}

% \begin{figure*}%[h!]
%     \vspace{-0.2in}
%     \footnotesize
%     \centering
%     \resizebox{.95\textwidth}{!}{%
%     % \hspace{-0.2in}
%     \begin{tabular}{ccc}    
%     \includegraphics[align=c, height=3cm]{figures/attn_div/sup_attn_diversity_attt9_v2.png}&%\hspace{-0.2in}
%     \includegraphics[align=c, height=3cm]{figures/attn_div/siam_attn_diversity_attt9_v2.png}&%\hspace{-0.2in}
%     \includegraphics[align=c, height=3cm]{figures/attn_div/mim_attn_diversity_attt9_v2.png}%    \vspace{-0.1in}
%     \\    
%     \includegraphics[align=c, height=3cm]{figures/attn_div/sup_entr_diversity_attt9_v2.png}&
%     \includegraphics[align=c, height=3cm]{figures/attn_div/siam_entr_diversity_attt9_v2.png}&
%     \includegraphics[align=c, height=3cm]{figures/attn_div/mim_entr_diversity_attt9_v2.png}
%     \\
%     {Supervised} &
%     {SimSiam} &
%     {SimMIM} \\
%     \end{tabular}}
%     \vspace{-0.1in}
%     \caption{\footnotesize \textbf{Top row: Visualization of averaged attention distance} on an OOD task (T9) at each layer at the end of each continual pre-training task phase (T0$\rightarrow$T8). The radius of marker indicates the standard deviation over attention heads in the corresponding layer. \textbf{Bottom row: Averaged attention entropy} on T9.}
%     \label{fig:swin-attn-div}
%     \vspace{-0.1in}
% \end{figure*}

%% file: materials/figures/5_experiments_anal.tex
\begin{figure}[h!]
    % \vspace{-0.1in}
    \footnotesize
    \centering
    \resizebox{0.5\textwidth}{!}{%
    % \hspace{-0.2in}
    \begin{tabular}{cc}    
    \includegraphics[height=4cm]{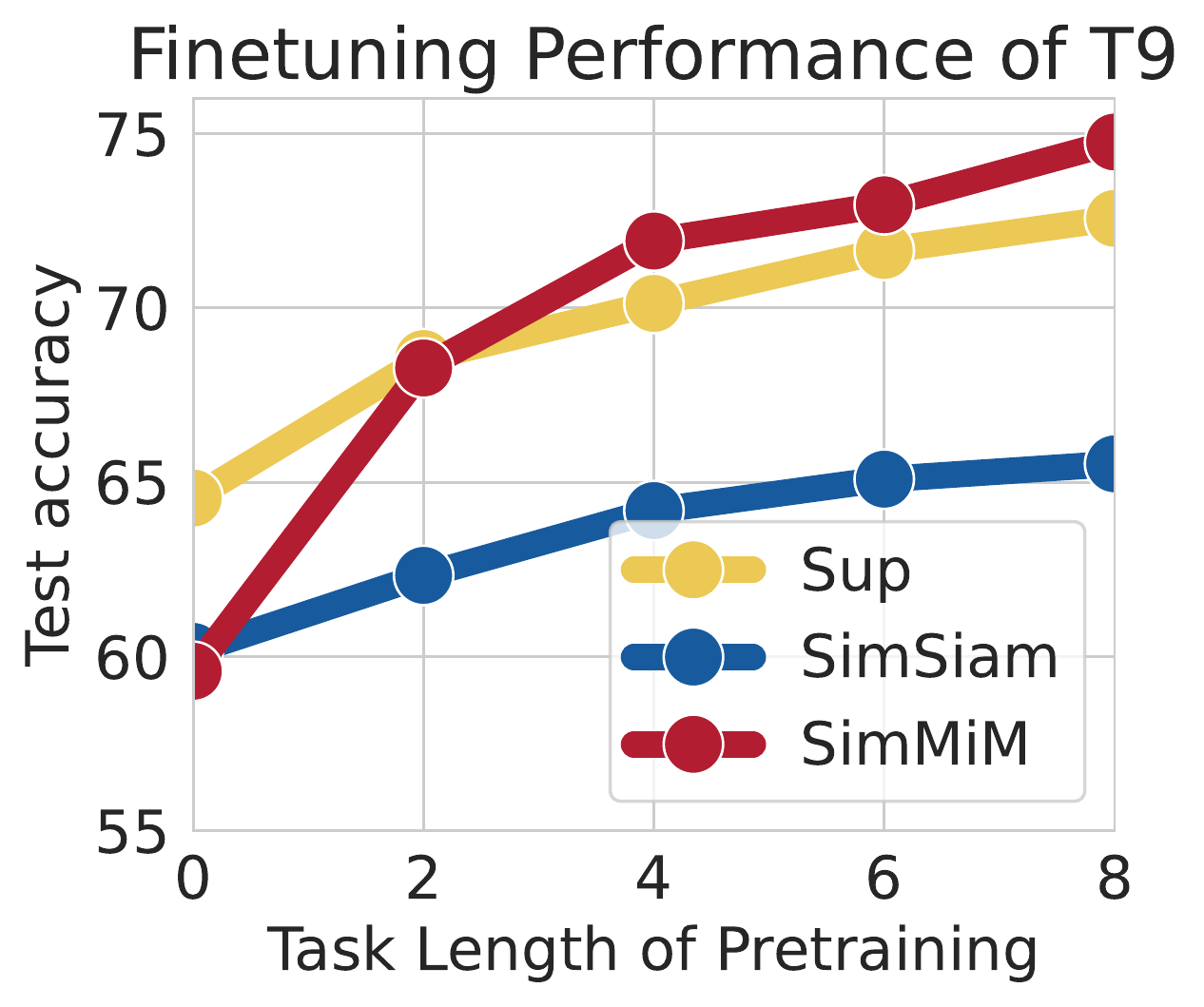} &
    \includegraphics[height=4cm]{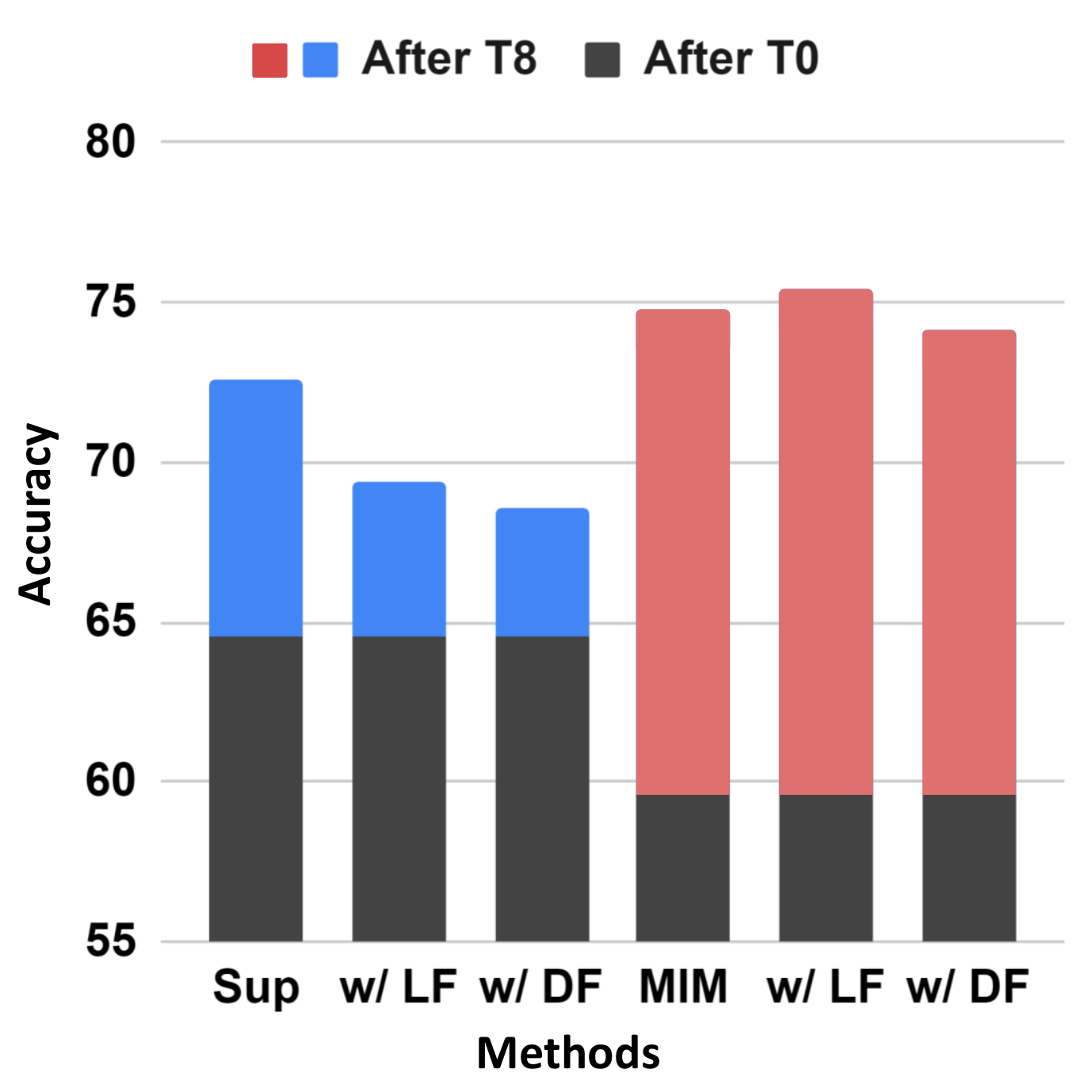} 
    \end{tabular}}
    \vspace{-0.15in}
    \caption{\small \textbf{Left: CL improves model generalization.} Sequential training on more tasks further increases fine-tuning performance on the unseen task (T9). \textbf{Right: The impact of freezing weights on a few lower/deeper layers during CL} after training the first task. We measure fine-tuning accuracy at the end of training the first task (T0) and last task (T8), and use supervised (\emph{Sup}) and masked image modeling (\emph{MIM}). \emph{LF} and \emph{DF} denote \emph{freezing lower layers} and \emph{freezing deeper layers}, respectively.}
    \label{fig:exp-anal}
    \vspace{-0.1in}
\end{figure}

%% file: materials/figures/5_glad_cifar100_figure.tex
\begin{figure}
\small
\centering
\resizebox{0.45\textwidth}{!}{%
    % \hspace{-0.2in}
\begin{tabular}{cc}    
\includegraphics[height=3.2cm]{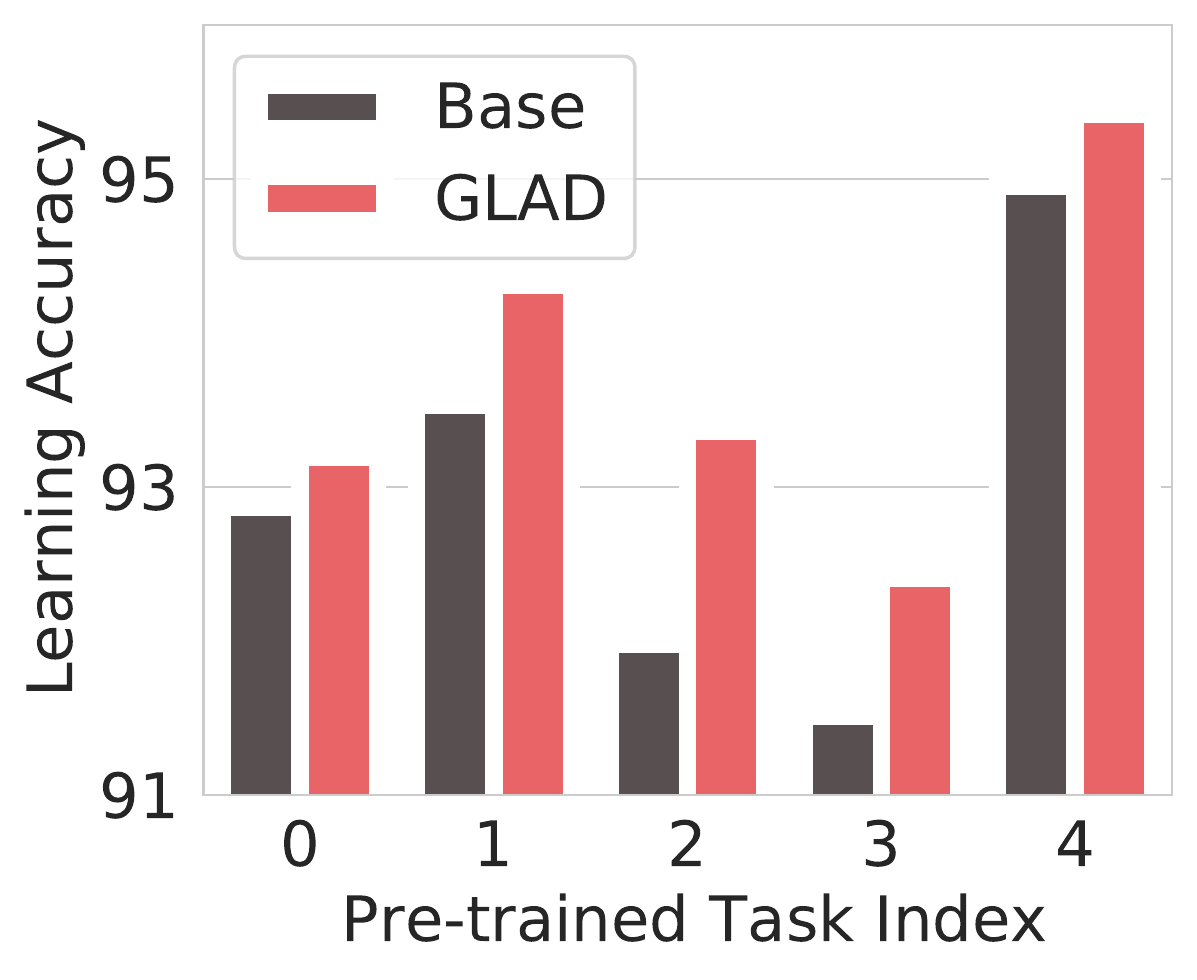}& 
\includegraphics[height=3.2cm]{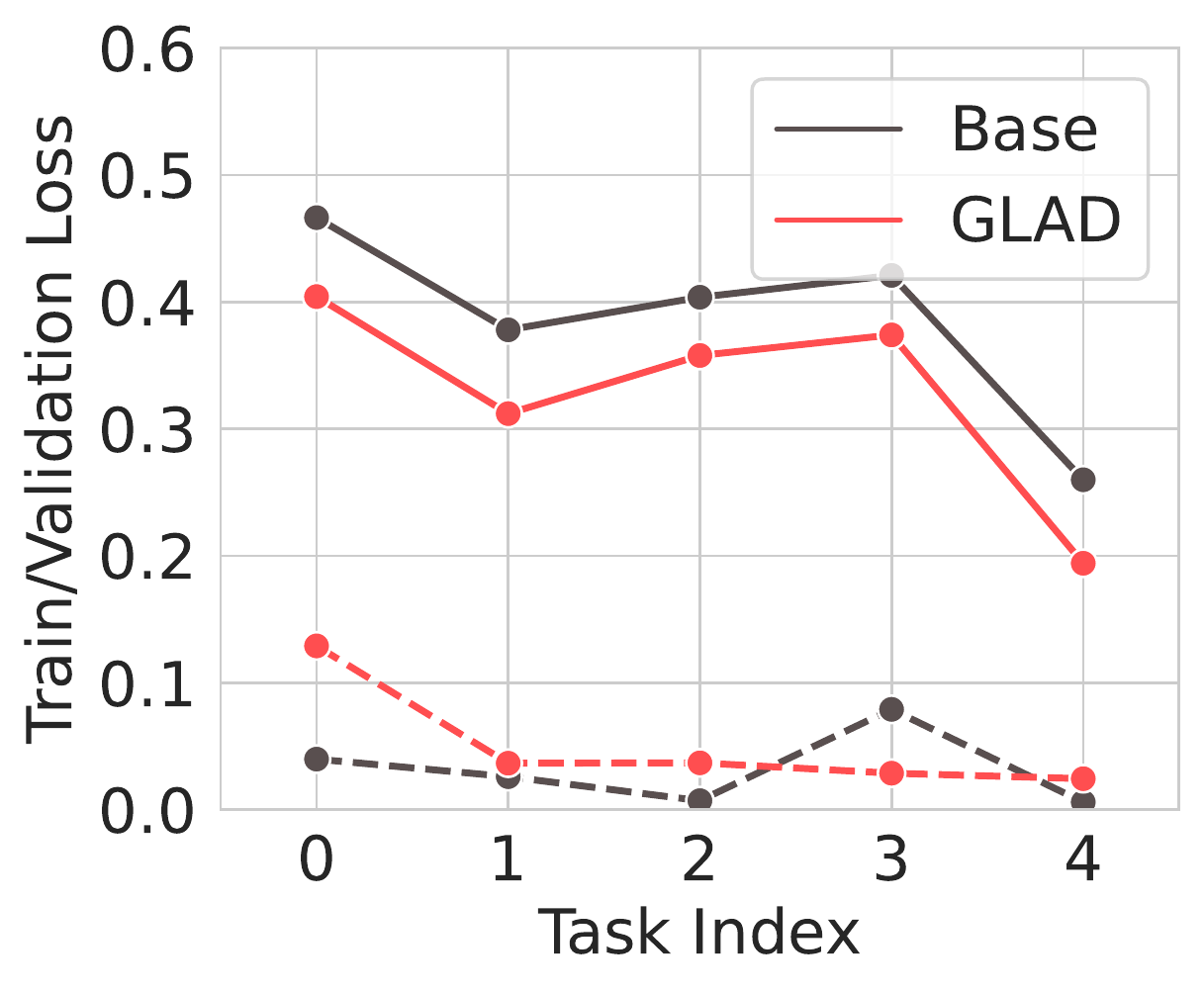}
\end{tabular}}
\vspace{-0.15in}
\caption{\small \textbf{Learning accuracy and loss of CIFAR-100 (T=5) at each continual pre-training step}. Right: Solid and dashed lines indicate validation and training loss, respectively.}\label{fig:glad-results-cifar}
\vspace{-0.05in}
\end{figure}

%% file: materials/tables/5_ours_table_one_column.tex
\begin{table}
\begin{center}
\small
\resizebox{0.44\textwidth}{!}{
\begin{tabular}{ccccc}
\toprule
% \multicolumn{1}{c}{\textsc{Methods}}&\multicolumn{4}{c}{\textsc{Fine-tuning Acc of T9}}\\
\multicolumn{1}{c}{\textsc{Pre-trained T0$\rightarrow$T4}}&\multicolumn{4}{c}{Continual learning over T5$\rightarrow$T8}\\
\midrule
& \textsc{T5}& \textsc{T6}& \textsc{T7}& \textsc{T8}\\
\midrule

\textsc{Supervised} & 
{64.56}  & 
{69.06}  &
{75.06}  & 
{77.80}  \\
\cdashlinelr{1-5}
\textsc{+ GLAD (Ours)} & 
\textbf{65.78}  & 
\textbf{69.84}  & 
\textbf{76.12}  & 
\textbf{79.04}  \\
\midrule
\midrule

\textsc{SimMIM}~{\cite{xie2021simmim}} & 
\textbf{68.34}  & 
{72.22}  & 
{77.38}  & 
{80.12}  \\
\cdashlinelr{1-5}
\textsc{+ GLAD (Ours)} & 
{68.24}  & 
\textbf{73.24}  & 
\textbf{78.94}  & 
\textbf{81.62}  \\
\midrule
\midrule

\textsc{MAE}~{\small \cite{he2022masked}} & 
\textbf{42.19} & 
{52.07}  & 
{63.07}  & 
{71.19}  \\
\cdashlinelr{1-5}
\textsc{+ GLAD (Ours)} & 
{41.75}  & 
\textbf{54.74}  & 
\textbf{68.01}  & 
\textbf{74.30}  \\
\bottomrule
\end{tabular}
}
\vspace{-0.1in}
\caption{\small \textbf{Fine-tuning accuracy of T9 at each continual pre-training step (T5 $\rightarrow$ T8)} over ImageNet1k-split. We initialized the model weights using continual (representation) learning over the earlier five sequential tasks (T0 $\rightarrow$ T4) to focus on continual pre-training performance on a few later tasks.}
\label{tab:glad-results-one}
\end{center}
\vspace{-0.2in}
\end{table}

%% file: sections/6_conclusion.tex
\section{Conclusion}
As powerful representation models have a great versability to solve various downstream tasks, exploring incremental pre-training strategy on a number of sequential tasks can be a practical and important approach.
This paper delves into how supervised and unsupervised continual learning affects model generalization from various perspectives. To our surprise, continual learning models preserve or even increment their transferability on in- and out-of-distribution tasks, increasing fine-tuning performance as pre-training more tasks. We scrutinize the behavior of representations in CL frameworks in the pre-training, including masked image modeling-based unsupervised continual learning, and find that the continual learner tends to forget class-discriminative features while progressively accumulating transferable features. 
Motivated by our observations, we propose a new method for continual pre-training to help backbone weights gain transferability during fine-tuning by introducing a new MSA module with parametric adaptors.
We believe the exploration of continuous learnability of the representation model would contribute to developing eco-friendly and resource-efficient training regimes for broad research/industry fields.

\section*{Acknowledgements}
We thank the anonymous reviewers for their insightful comments and suggestions. This work was supported by Microsoft Research Asia and Institute of Information \& communications Technology Planning \& Evaluation (IITP) grant funded by the Korea government(MSIT) (No.2019-0-00075, Artificial Intelligence Graduate School Program(KAIST)). %Any opinions, findings, and conclusions or recommendations expressed in this material are those of the authors and do not necessarily reflect the views of the funding agencies.

%% file: sections/7_appendix.tex
\mbox{~}
\clearpage
\appendix
\section{Details for Problem Setups}\label{sup:sec:details}
\paragraph{Datasets} We use ImageNet~\cite{deng2009imagenet} dataset, containing $1000$ classes of high-resolution object images with their corresponding labels. We split them into $10$ tasks, where each task consists of $100$ different classes. We use only $10\%$ of training instances in each task for pre-training, and use the full set for the fine-tuning and linear probe. Accuracy is measured by the validation dataset for each task. Note that \Cref{tab:glad-results-one} also use $10\%$ of the training set for pre-training earlier five sequential tasks from T0 to T4, and use the full training set for the sequential fine-tuning procedure from T5 to T8.

\paragraph{Architectures and baselines} 
We follow~\cite{madaan2022rethinking} for an unsupervised continual learning framework with contrastive self-supervised learning using SimSiam~\cite{chen2020simple}. 
For masked image modeling, we follow the setting of SimMIM~\cite{xie2021simmim} and MAE~\cite{he2022masked} using their official code repositories\footnote{https://github.com/microsoft/SimMIM}\footnote{https://github.com/facebookresearch/mae} where the masking ratio is $0.6$ and $0.75$, respectively. We use Vision Transformer~\cite{dosovitskiy2020image} (ViT-B) and Swin Transformer~\cite{liu2021swin} (Swin-T) for backbone architectures. In ViT-B, the embedding dimension is $768$, the layer depth is $12$, the number of heads is $12$, and the patch size is $16$. In Swin-T, the embedding dimension is $96$, the layer depth at each block is $\left[2, 2, 6, 2\right]$ (in total $12$), the number of heads at each block is $\left[3, 6, 12, 24\right]$, the patch size is $4$, and the sliding window size is $7$. We set the input image size to $224$ for all experiments but $192$ for SimMIM pre-training. For continual learning methods, we use SI~\cite{zenke2017continual}, DER~\cite{buzzega2020dark}, and LUMP~\cite{madaan2022rethinking}. The implementation is built upon an official code of LUMP\footnote{https://github.com/divyam3897/UCL}.

\paragraph{Training setups and hyperparameters}
We use AdamW optimizer~\cite{loshchilov2017decoupled} with cosine learning rate decay and the warmup for all experiments. For the pre-training phase at each task, we train the model $60$ epochs on supervised learning and $100$ epochs on unsupervised learning models as self-supervised learning methods without label supervision may require more iterations to converge. For fine-tuning, we basically perform $30$ epochs training. For fine-tuning from the model pre-trained Imagenet 1K \& 22K in ~\Cref{tab:main-table}, we set the number of training epochs to $10$ as they rapidly converge within a few iterations. We set the hyperparameter for balancing the degree of regularization term $\lambda=100$ for SI, $\lambda=0.1$ for DER, and $\alpha=0.1$ for LUMP. And the buffer size is $200$ for rehearsal-based continual learning methods like DER and LUMP. We set the batch size to $64$ for SimSiam pre-training, otherwise $128$. \Cref{tab:hyperparameters} summarizes the learning rate and training epochs for experiments, and we linearly scale the learning rate with $batch\_size / 512$ in practice to reflect the input variance, followed by 
\cite{goyal2017accurate}.

\input{materials/tables/7_lr.tex}

\paragraph{Evaluation metrics}
We additionally introduce a linear classifier per task and measure fine-tuning and linear probe (or linear evaluation) accuracy by using the pre-trained backbone. For linear evaluation, we independently train a classifier over the training set of the corresponding task while freezing the backbone weights, and measure the accuracy on the validation set. For fine-tuning, we train a classifier as well as backbone weights over the training set of the target task. In this paper, we denote the term \emph{backward transfer (BWT)}, $acc^{bwt}_{t}$, of task $t$ as the fine-tuning accuracy (or linear probe accuracy) disparity between the model after training task $t$ and the end of sequential training. Given $T$ sequential tasks $\{\mathcal{T}_t\}^T_{t=1}$:
\begin{equation}\label{eq:fwdtransfer}
\begin{split}
acc^{bwt}_{t}=acc_{t,T}-acc_{t,t},\\
\end{split}
\end{equation}
% where $\bm{a}_{i,j}$ is the measured accuracy for task $i$ from the backbone model pre-trained the first task $t=1$ to $j^{th}$ task sequentially.
where $acc_{i,j}$ is the measured accuracy for task $i$ using sequentially pre-trained backbone model from the first task to $j^{th}$ task. Followed by~\citet{xie2022revealing}, we compute the attention distance at each attention head to analyze the change of attention during continual learning. Let $\bm{a}^{l,i}$ be an $i^{th}$ attention head output at layer $l$, the averaged attention distance $d_{\bm{a}^{l,i}}$ is measured by
\begin{equation}\label{eq:adist}
\begin{split}
d_{\bm{a}^{l,i}}=\sum_{j}c^*\bar{\bm{a}}^{l,i}_j,\\
\end{split}
\end{equation}
where $c^*$ is the corresponding distance map and $\bar{\bm{a}}^{l,i}$ is a normalized attention matrix. %Unlike ViTs, which have the same number of heads across layers, Swin transformer-based frameworks increase the number of heads in deeper layers.
% Images are observed to exhibit strong locality: pixels near each other tend to be highly correlated [37], motivating the use of local priors in a wide range of visual perception architectures [21, 46, 44, 33, 55]. In the era of Vision Transformers, the usefulness of local priors has still undergone rich discussions and trials [17, 55, 49]. Thus it is valuable to investigate whether MIM models bring the locality inductive bias to the models. We do this by computing averaged attention distance in each attention head of each layer.

\section{Further Discussions on Meta Learning}\label{sup:sec:meta-learning}
\citet{chen2023forgetting} observes that the representation learned by \textit{meta-learning} can be generic and useful for CL in terms of forgetting and forward transfer. Provided results and insights are interesting, therefore, here we discuss a few clear differences compared to Meta-learning~\cite{finn2017model}. First, we do not need an external buffer for the rehearsal, while FOMAML~\cite{finn2017model} requires keeping a (sub)set of past task data for inner loop update. Their reliance on rehearsal buffers has been criticized~\cite{hadsell2020embracing,lomonaco2020rehearsal,wang2022dualprompt} in the CL field. The performance of the rehearsal-based methods is sensitive to the size of the buffer~\cite{prabhu2020gdumb}, and they cannot be used in applications with concerns about data privacy. The MAML-based method requires exhaustive computational cost due to iterative and sequential inner loop updates for sampled past tasks at each iteration of current task training. On the other hand, our GLAD requires a marginal additional computation over the original training. Our GLAD gains benefit from adaptor-based modulation. In GLAD, the backbone focuses on capturing task-generic information, and a lightweight adaptor transforms them to task-adaptive attention for downstream task adaptation. Therefore, we can anytime recover the past model from the current one without storing full weights, by simply re-attaching the GLAD adaptor learned on the past task. Moreover, we can remove the unnecessary past task-specific knowledge from the model, or avoid the threat of performance degradation from training on noisy tasks/datasets - since the module for task-adaptive transformations is physically separate from generic representation (which is similar to \cite{Yoon2020Scalable}). These practical utilizations are unavailable for Meta-learning as they update the entire weights without careful consideration of the knowledge modulation.

\section{Additional Analyses}\label{sup:sec:analyses}
\subsection{Evaluation on Task Order Shuffling}\label{sup:subsec:task_order}
\input{materials/tables/5_task_order_table}
We additionally performed the evaluation on three different task orders for further reliable analyses in our experiments. At first, we measure fine-tuning and linear evaluation performance with the backward transfer using CIFAR-100. Similar to our experimental setting using ImageNet, we split CIFAR100 into ten tasks, containing ten disjoint classes per task. Then we train on the earlier nine tasks sequentially and measure the accuracy of the OOD task (10th), which is not seen during continual learning, as well as the forgetting of the first task with the backward transfer.
% In short, we observed similar trends with the ImageNet results of~\Cref{tab:main-table}. The MIM-based UCL method achieves higher fine-tuning performance on the OOD (T9) task and backward transfer on the base model and SI compared to the supervised CL frameworks, which are consistently observed over all three task orders.
Regardless of the three orders of the task sequence, we observed that our proposed masked modeling-based UCL framework consistently surpasses supervised CL in terms of the fine-tuning performance on OOD task (T9) over all continual learning methods (Base/SI/LUMP), as similar to ImageNet results in Table 1.
This is because masked modeling-based UCL continuously trains on more improved generic representations, evident in~\Cref{fig:swin-attn-div} that continual learners gradually capture richer task-general (or low-level) features behaving with more local attention (i.e., lower attention distance), which retains localized information with strong local inductive bias, such as edges, patterns, and textures.

\subsection{Attention Distance and Entropy from Different Transformer Backbones}\label{sup:subsec:analyses-vitswin}
We plot the attention distance and distribution of attention heads per layer for ViT-B in \Cref{sup:fig:vit-attn} and \Cref{sup:fig:vit-entr}, respectively. And also, plot the attention distance and entropy of the distribution of attention heads per layer for Swin-T in \Cref{sup:fig:swin-attn} and \Cref{sup:fig:swin-entr}. Both self-attention-based architectures 
similarly behave according to the learning frameworks, i.e., \emph{Supervised}, \emph{SimSiam}, and \emph{SimMIM}. Note that two consecutive layers in Swin-T repeat relatively high and low values for both metrics since two successive swin transformer blocks (\emph{S-MSA} and \emph{SW-MSA} in their original paper) aggregate locality in different ranges.

% Also, we find that the averaged aggregated distances in two consecutive layers are one high and one low. This is due to the shifted windowing scheme in Swin Transformer, that is, the ranges that each pixel can aggregate in two consecutive layers are different.
% \input{materials/figures/5_attn_dist.tex}
% \input{materials/figures/5_swin_attn.tex}

\input{materials/figures/7_cf_entropy_anal.tex}
\subsection{Change of Aggregated Attention Distance and Entropy during Continual Learning}{sup:subsec:analyses-entr}
We additionally visualize the movement of aggregated entropy of the distribution from each attention head in \Cref{sup:fig:swin-attn-entr-diw}. We visualize the plot for attention distance in~\Cref{fig:swin-attn-div} again for a better comparison between them. Similar to observations in attention distance visualization, aggregated attention entropy gradually decreases as proceeding to pre-train more tasks, encouraging incremental model generalization. And aggregated attention entropy for supervised and contrastive learning-based continual learning frameworks suffer from a small diversity with a high average amount of information for all attention heads, compared to SimMIM. 

\subsection{Aggregated Attention Distance and Entropy while Freezing Partial Layers}\label{sup:subsec:analyses-freeze}
We further visualize the movement of aggregated attention distance and entropy when freezing the two lowest and deepest layers in ~\Cref{sup:fig:partial-freeze-swin-attn-div} and \Cref{sup:fig:partial-freeze-swin-entr-div}, respectively. These experiments are exactly from \Cref{tab:glad-results-one}. Interestingly, if SimMIM freezes a few layers during continual learning, the remaining trainable layers tend to decrease their attention distance and entropy more actively. We believe that this is a reason that the SimMIM does not find noticeable performance degeneration in fine-tuning, even freezing a few layers during continual learning. However, we didn't observe a significant change in supervised continual learning. This is because supervised learning is prone to rigidly focus on different features according to the layer depths (i.e., global to local), and therefore cannot flexibly cope with capturing locality inductive bias.

\section{Limitations and Societal Impact}
\paragraph{Limitations} Although we have shown promising results and findings in multiple CL frameworks, our proposed GLAD module requires extra memory and computation for the regularization term. Moreover, We design the experiment of continual pre-training with less than ten tasks, which is insufficient to evaluate lifelong learning. Reducing the memory cost for the additional adaptor and extending our framework to a larger number of continual pre-training tasks will be important for future work. 
\paragraph{Negative societal impact} Our work doesn't store past task data in the buffer so that we can avoid the negative societal impact raised by data privacy issues in the community.

\input{materials/figures/5_attn_dist.tex}
\input{materials/figures/5_swin_attn.tex}

\input{materials/figures/7_cf_partial_div.tex}
\input{materials/figures/7_cf_partial_div_entr.tex}

%% file: materials/tables/7_lr.tex
\begin{table}[H]
\caption{{\bf Basic configurations} for three continual learning frameworks during pre-training and fine-tuning, where $\eta=2e^{-4}$. We report the best combinations of the learning rate for pre-training and fine-tuning in the range of $\left[0.1, 0.2, 0.5, 1.0, 2.0, 5.0, 10.0\right] \times\eta$ and $\left[0.1, 0.5, 1.0, 5.0\right]\times\eta$, respectively.\label{tab:hyperparameters}}
\vspace{-0.1in}
\small
\centering\begin{tabular}{rcc}
\toprule
\textsc{} & \textsc{pre-training} & \textsc{fine-tuning} \\
\midrule
\textsc{supervised} & 
\textit{lr}: $1.0\eta$,~~\textit{epoch}: $60$&
\textit{lr}: $0.5\eta$,~~\textit{epoch}: $30$ \\
\textsc{contrastive}  &
\textit{lr}: $0.2\eta$,~~\textit{epoch}: $100$&
\textit{lr}: $1.0\eta$,~~\textit{epoch}: $30$ \\
\textsc{mim}  &
\textit{lr}: $5.0\eta$,~~\textit{epoch}: $100$&
\textit{lr}: $5.0\eta$,~~\textit{epoch}: $30$ \\
\bottomrule
\end{tabular}
% \centering\begin{tabular}{lccc}
% \toprule
% \textsc{} & \textsc{Supervised} & \textsc{Contrastive} & \textsc{Masked Modeling} \\
% \midrule
% \textsc{Pre-training} & 
% \textit{lr}: $1.0~\eta$,~~\textit{epochs}: $60$&
% \textit{lr}: $0.2~\eta$,~~\textit{epochs}: $100$&
% \textit{lr}: $5.0~\eta$,~~\textit{epochs}: $100$\\
% \textsc{Fine-tuning}  &
% \textit{lr}: $0.5~\eta$,~~\textit{epochs}: $30$ &
% \textit{lr}: $1.0~\eta$,~~~~\textit{epochs}: $30$ &
% \textit{lr}: $5.0~\eta$,~~~~\textit{epochs}: $30$ \\
% \bottomrule
% \end{tabular}
\vspace{-0.1in}
\end{table}

%% file: materials/tables/5_task_order_table.tex
\begin{table}
\begin{center}
\small
\setlength{\tabcolsep}{3pt} % 
\resizebox{0.48\textwidth}{!}{
\begin{tabular}{l@{\hspace{6pt}}crrrr}
\toprule
\multicolumn{2}{c}{}&\multicolumn{2}{c}{\textsc{Supervised}}&\multicolumn{2}{c}{Masked Model}\\
\midrule
&
& \textsc{Final Acc} & \textsc{Neg. BWT}
& \textsc{Final Acc} & \textsc{Neg. BWT}\\
\midrule
\multirow{3}{*}{\rotatebox[origin=c]{90}{\small FT}}
% \cmidrule{2-6}
& \textsc{Base} & 
{70.47} \tiny{($\pm$ 7.91)}  & {1.77} \tiny{($\pm$ 1.56)} &
\textbf{86.17 \tiny{($\pm$ 1.99)}}  & \textbf{20.90 \tiny{($\pm$ 1.88)}}\\

& \textsc{Si}& 
{70.93} \tiny{($\pm$ 6.11)}  & {4.80} \tiny{($\pm$ 2.24)} &
{83.50} \tiny{($\pm$ 3.04)}  & {16.80} \tiny{($\pm$ 3.33)}\\

& \textsc{LUMP}& 
N/A & N/A &
\textbf{85.80 \tiny{($\pm$ 2.57)}} & {18.30} \tiny{($\pm$ 1.18)}\\
\midrule
\multirow{3}{*}{\rotatebox[origin=c]{90}{\small LP}}
% \cmidrule{2-6}

& \textsc{Base} & 
{62.53} \tiny{($\pm$ 6.58)}  & {0.03} \tiny{($\pm$ 0.05)} &
{67.53} \tiny{($\pm$ 4.05)}  & {10.03} \tiny{($\pm$ 1.17)}\\
& \textsc{Si}& 
{62.47} \tiny{($\pm$ 5.12)}  & {2.07} \tiny{($\pm$ 2.16)} &
{64.60} \tiny{($\pm$ 2.72)}  & {9.77} \tiny{($\pm$ 0.87)} \\

& \textsc{LUMP} & 
N/A & N/A &
\textbf{68.97 \tiny{($\pm$ 5.98)}}  & \textbf{12.60 \tiny{($\pm$ 0.99)}}\\
\bottomrule
\end{tabular}}
\vspace{-0.1in}
\caption{\small \textbf{Averaged fine-tuning and linear probe performance} on CIFAR-100 Split. We report the mean and standard deviation for three independent runs with randomly shuffled task orders.}
\label{tab:order-cifar}
\end{center}
\vspace{-0.15in}
\end{table}

%% file: materials/figures/7_cf_entropy_anal.tex
\begin{figure*}[h!]
    % \vspace{-0.2in}
    \footnotesize
    \centering
    \resizebox{1\textwidth}{!}{%
    % \hspace{-0.2in}
    \begin{tabular}{ccc}    
    \includegraphics[align=c, height=3cm]{figures/attn_div/sup_attn_diversity_attt9_v2.png}&%\hspace{-0.2in}
    \includegraphics[align=c, height=3cm]{figures/attn_div/siam_attn_diversity_attt9_v2.png}&%\hspace{-0.2in}
    \includegraphics[align=c, height=3cm]{figures/attn_div/mim_attn_diversity_attt9_v2.png}%    \vspace{-0.1in}
    \\    
    \includegraphics[align=c, height=3cm]{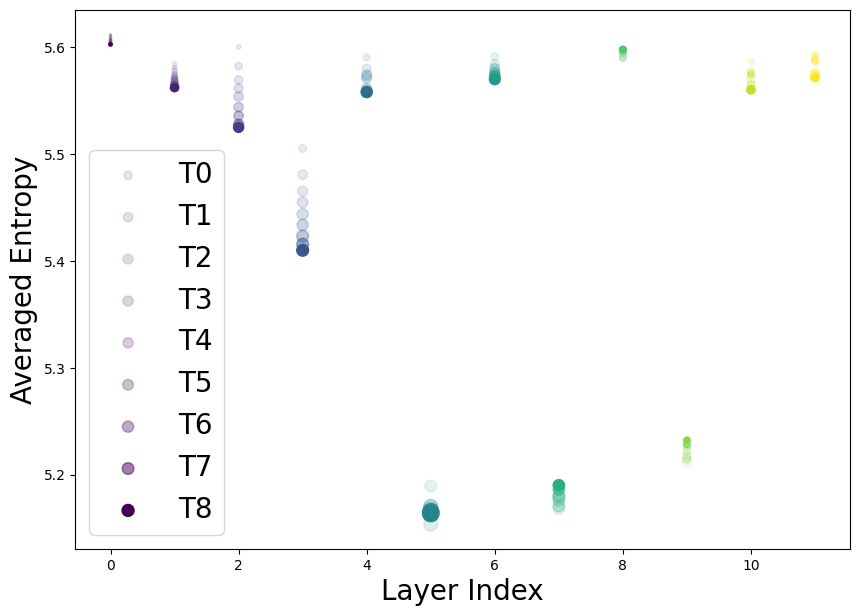}&
    \includegraphics[align=c, height=3cm]{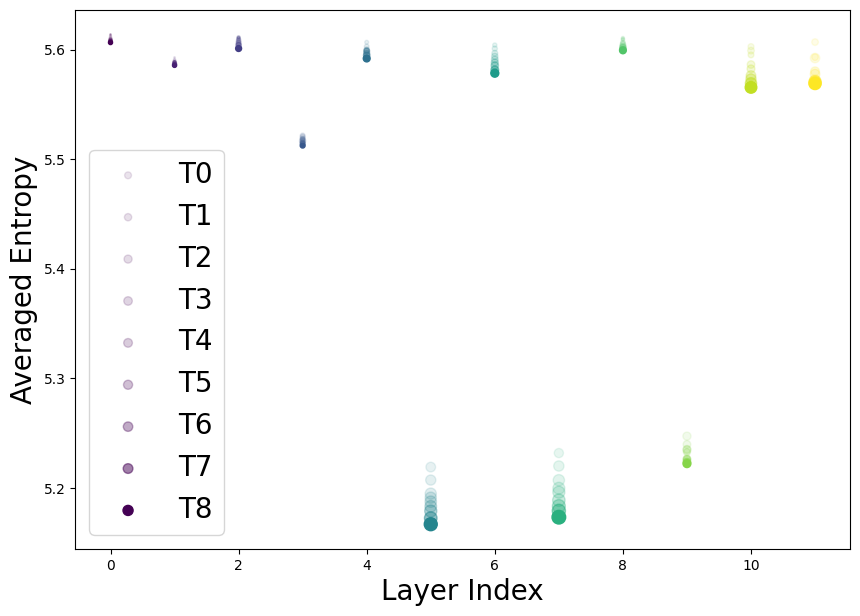}&
    \includegraphics[align=c, height=3cm]{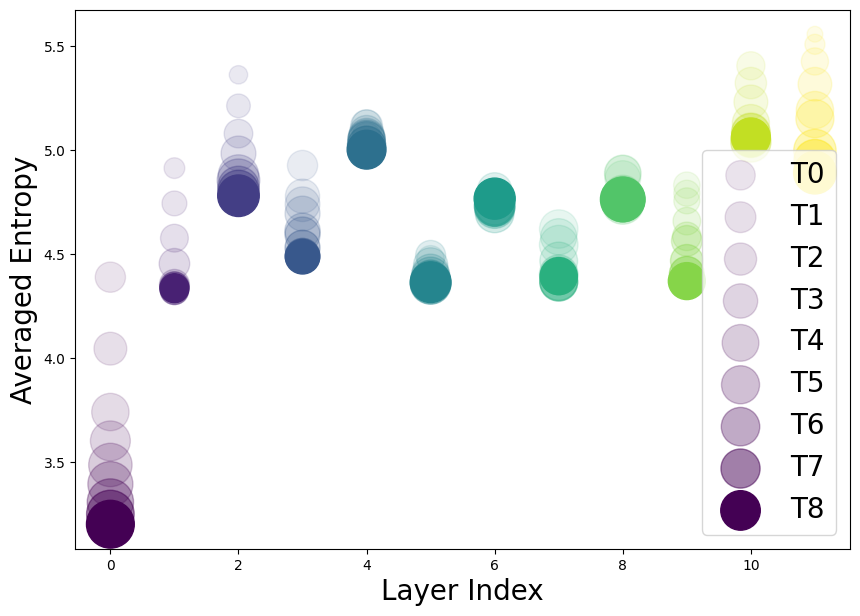}
    \\
    {Supervised} &
    {SimSiam} &
    {SimMIM} \\
    \end{tabular}}
    \vspace{-0.1in}
    \caption{\footnotesize \textbf{Top row: Visualization of aggregated attention distance} on an OOD task (T9) at each layer at the end of each continual pre-training task phase (T0$\rightarrow$T8). The radius of marker indicates the standard deviation over attention heads in the corresponding layer. \textbf{Bottom row: Aggregated attention entropy} on T9.}
    \label{sup:fig:swin-attn-entr-diw}
    % \vspace{-0.1in}
\end{figure*}

%% file: materials/figures/5_attn_dist.tex
\begin{figure*}%[h!]
    \centering
    \resizebox{0.9\textwidth}{!}{%
    \vspace{-0.2in}
    \begin{tabular}{l@{\hspace{6pt}}ccccc}
    \rotatebox[origin=c]{90}{\footnotesize Supervised}\hspace{-0.1in}    
    &
    \includegraphics[align=c, height=3.cm]{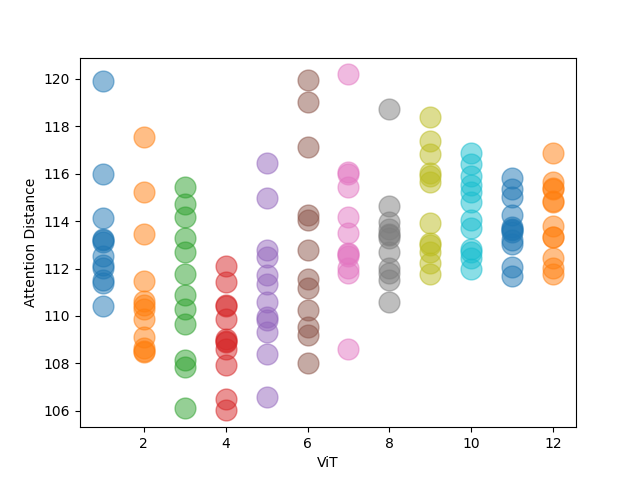}&\hspace{-0.3in}
    \includegraphics[align=c, height=3.cm]{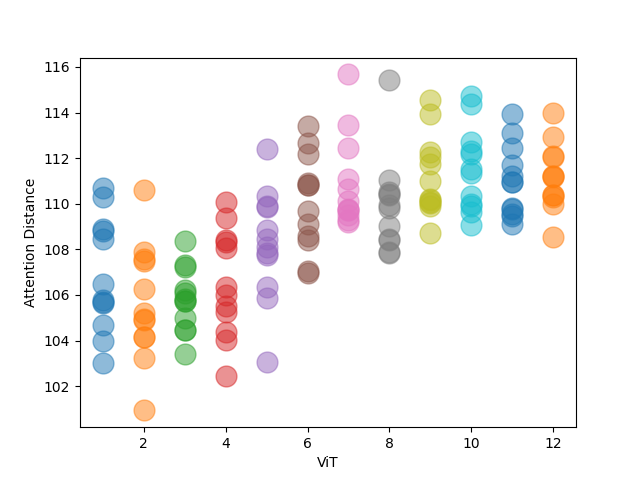}&\hspace{-0.2in}
    \includegraphics[align=c, height=3.cm]{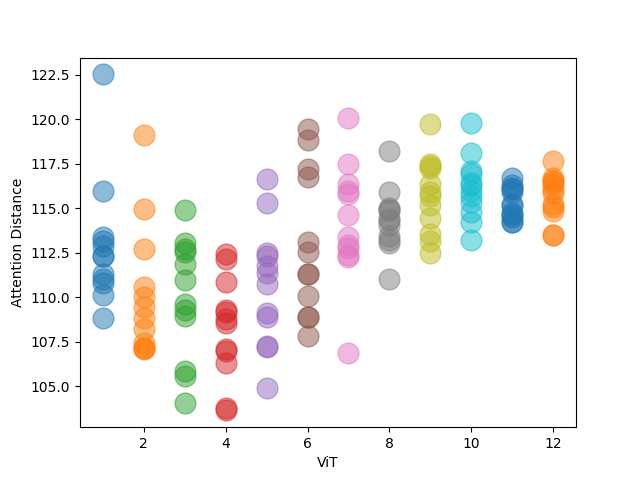}&\hspace{-0.3in}
    \includegraphics[align=c, height=3.cm]{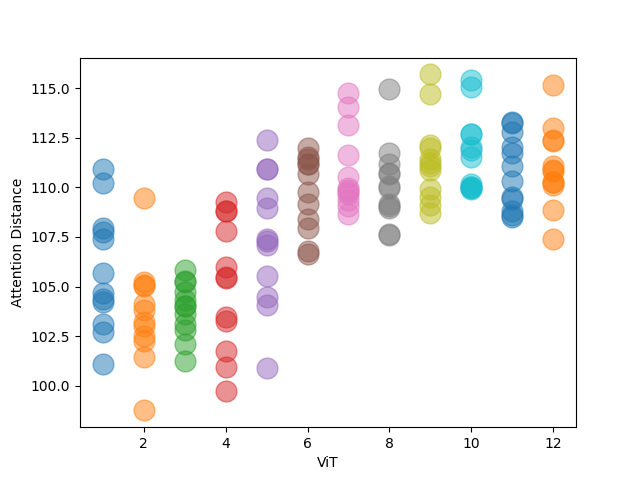}\\
    \rotatebox[origin=c]{90}{\footnotesize SimSiam}\hspace{-0.1in}    
    &
    \includegraphics[align=c, height=3.cm]{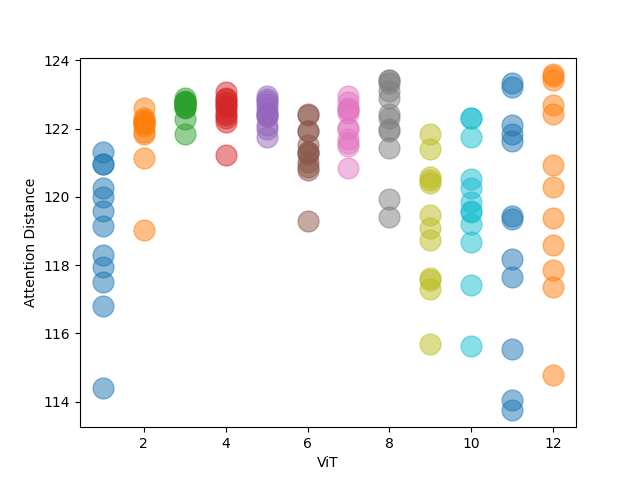}&\hspace{-0.3in}
    \includegraphics[align=c, height=3.cm]{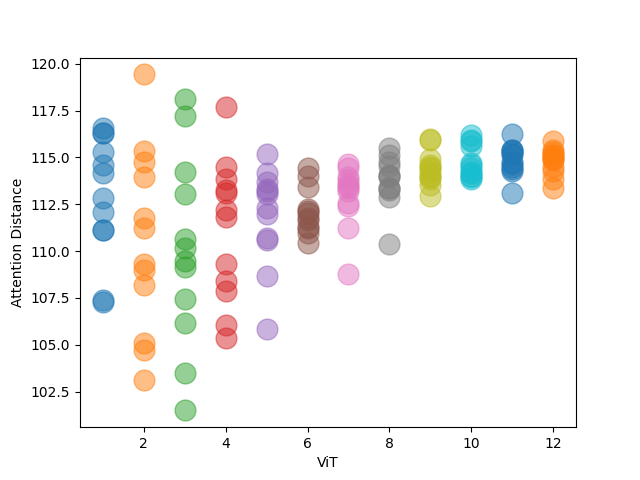}&\hspace{-0.2in}
    \includegraphics[align=c, height=3.cm]{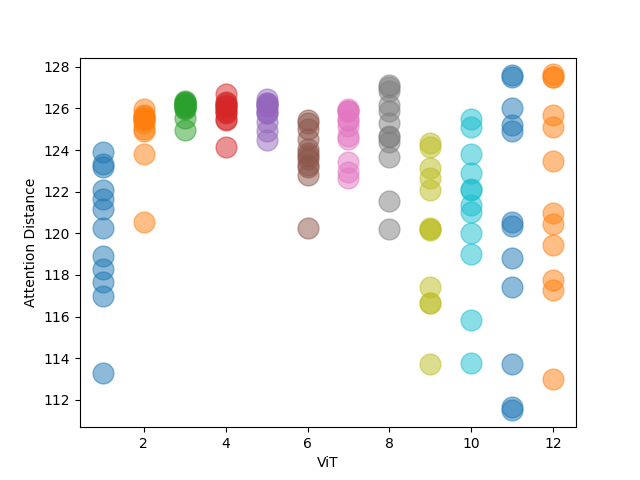}&\hspace{-0.3in}
    \includegraphics[align=c, height=3.cm]{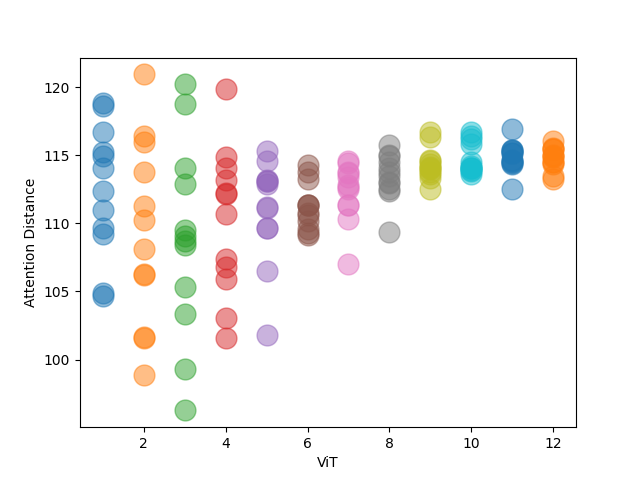}\\
    \rotatebox[origin=c]{90}{\footnotesize SimMIM}\hspace{-0.1in}
    &
    \includegraphics[align=c, height=3.cm]{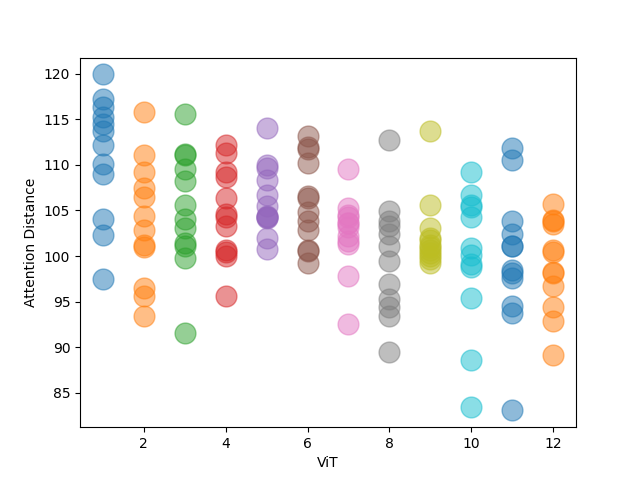}&\hspace{-0.3in}
    \includegraphics[align=c, height=3.cm]{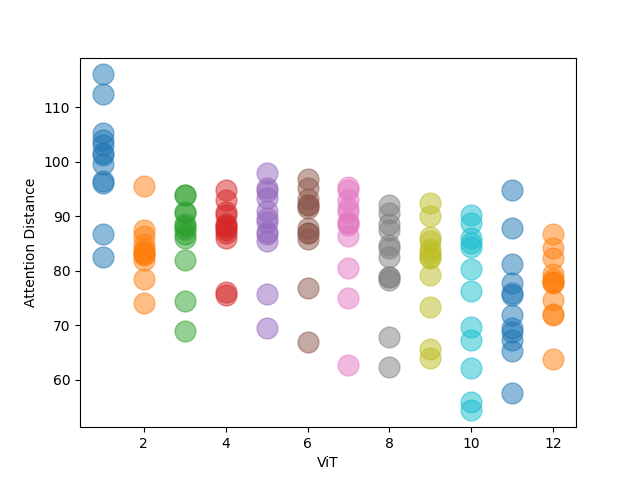}&\hspace{-0.2in}
    \includegraphics[align=c, height=3.cm]{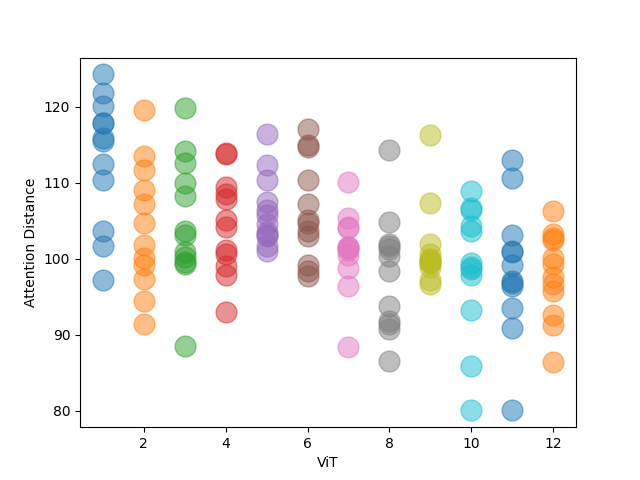}&\hspace{-0.3in}
    \includegraphics[align=c, height=3.cm]{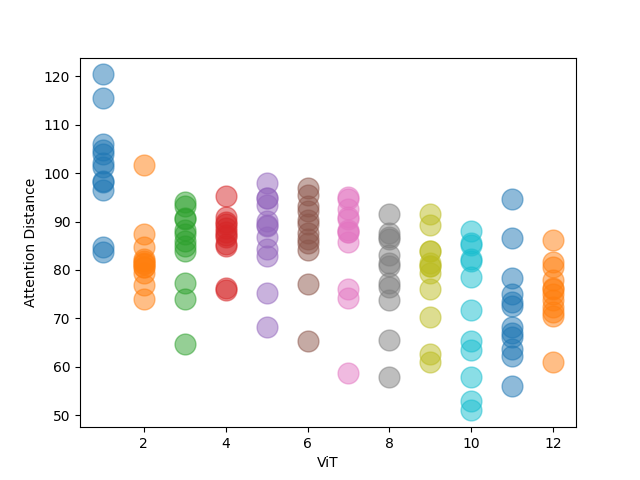}\\
    % &\multicolumn{2}{c}{\footnotesize  (a) Task 0} & \multicolumn{2}{c}{\footnotesize  (b) Task 8} \\
    &
    \footnotesize $\bm{w}_{\bf{T0}}$ & \hspace{-0.3in}
    \footnotesize $\bm{w}_{\bf{T0 \rightarrow T8}}$ &\hspace{-0.2in}
    \footnotesize $\bm{w}_{\bf{T0}}$ &\hspace{-0.3in}
    \footnotesize $\bm{w}_{\bf{T0 \rightarrow T8}}$ \\
    \end{tabular}}
    \vspace{-0.1in}
    \caption{\footnotesize \textbf{ViT-B Attention distance} of the first (T0) and the last task (T8) in a task sequence with respect to three continual pre-trained frameworks right after the completion of the first ($\bm{w}_{\bf{T0}}$) and last task ($\bm{w}_{\bf{T0\rightarrow T8}}$).}
    \label{sup:fig:vit-attn}
    \vspace{-0.1in}
\end{figure*}

\begin{figure*}%[h!]
    \centering
    \resizebox{.9\textwidth}{!}{%
    % \hspace{-0.2in}
    \begin{tabular}{l@{\hspace{6pt}}ccccc}
    \rotatebox[origin=c]{90}{\footnotesize Supervised}\hspace{-0.1in}    
    &
    \includegraphics[align=c, height=3.cm]{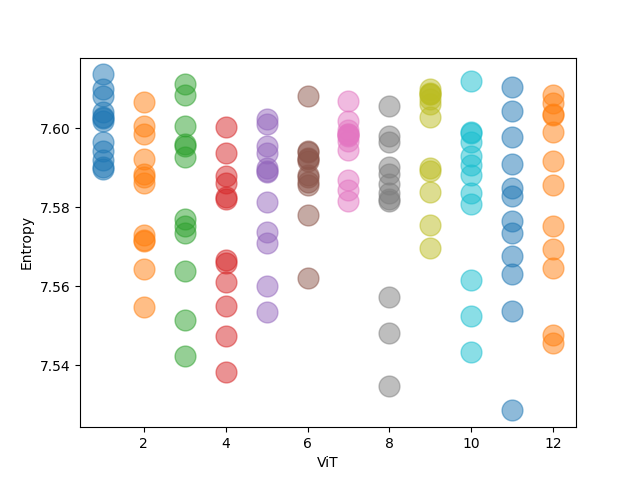}&\hspace{-0.3in}
    \includegraphics[align=c, height=3.cm]{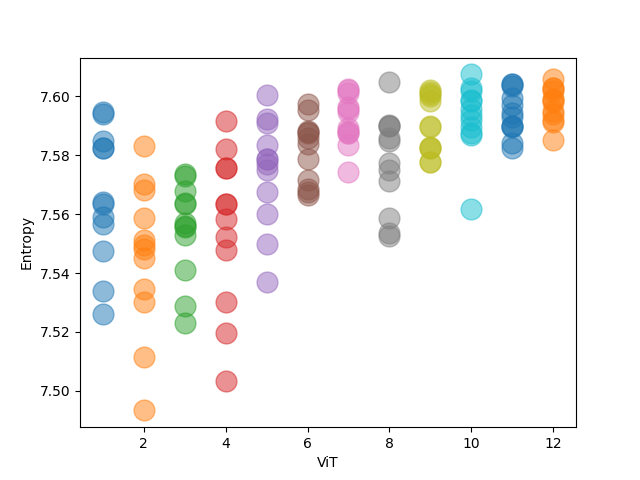}&\hspace{-0.2in}
    \includegraphics[align=c, height=3.cm]{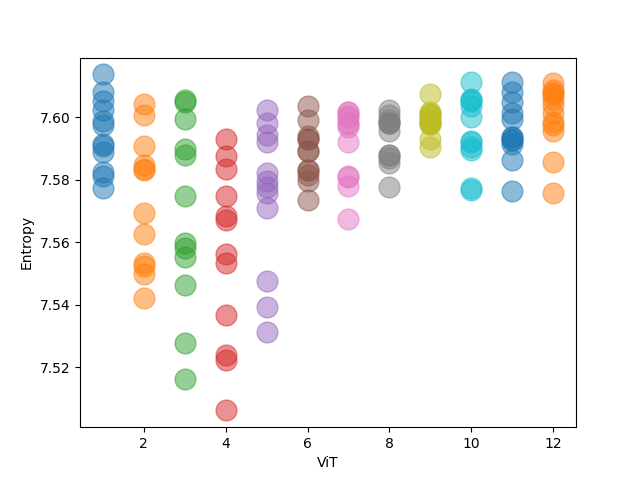}&\hspace{-0.3in}
    \includegraphics[align=c, height=3.cm]{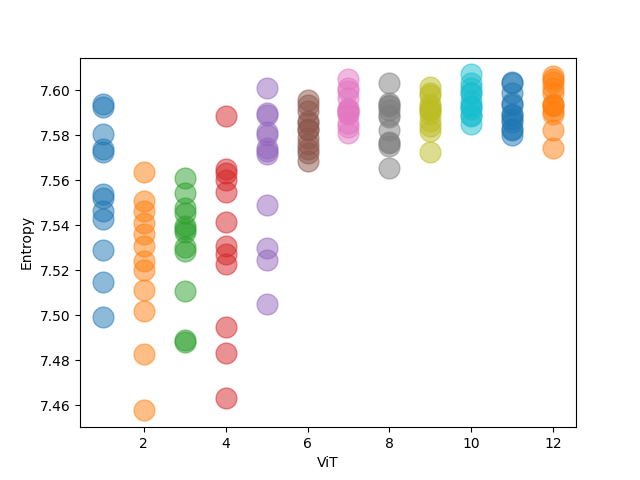}\\
    \rotatebox[origin=c]{90}{\footnotesize SimSiam}\hspace{-0.1in}    
    &
    \includegraphics[align=c, height=3.cm]{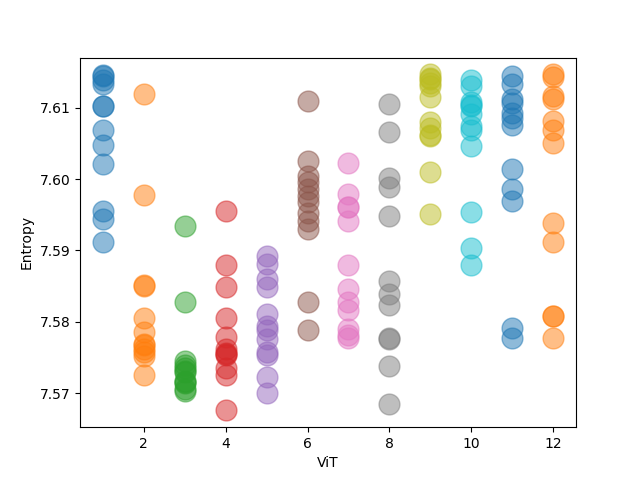}&\hspace{-0.3in}
    \includegraphics[align=c, height=3.cm]{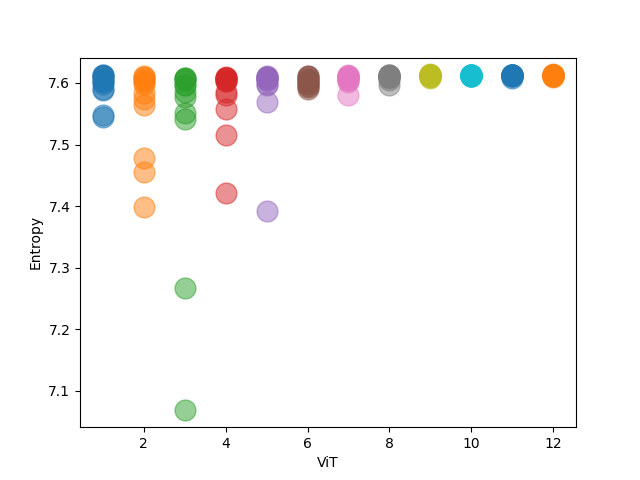}&\hspace{-0.2in}
    \includegraphics[align=c, height=3.cm]{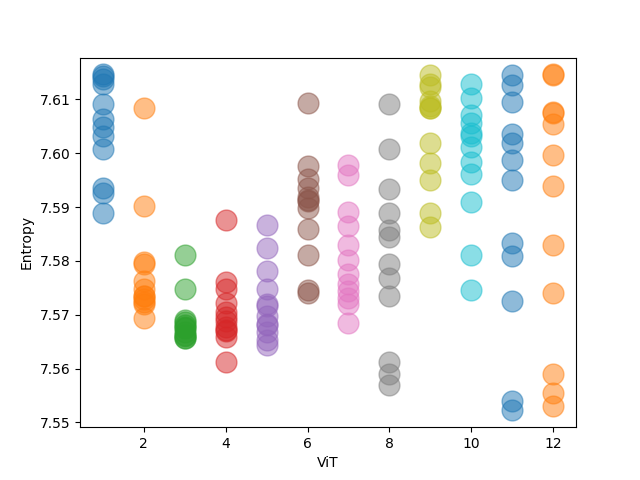}&\hspace{-0.3in}
    \includegraphics[align=c, height=3.cm]{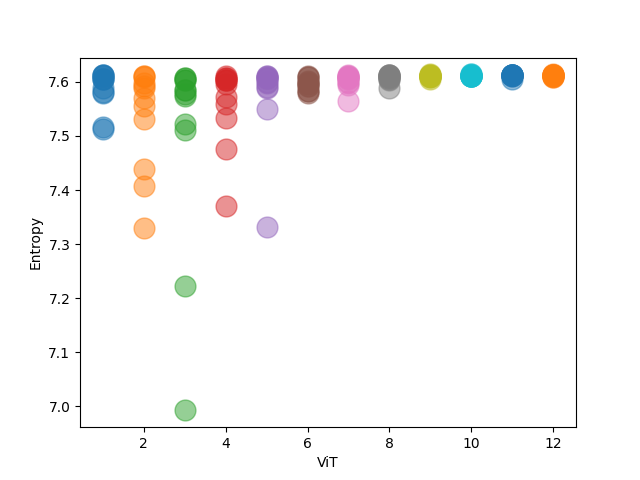}\\
    \rotatebox[origin=c]{90}{\footnotesize SimMIM}\hspace{-0.1in}
    &
    \includegraphics[align=c, height=3.cm]{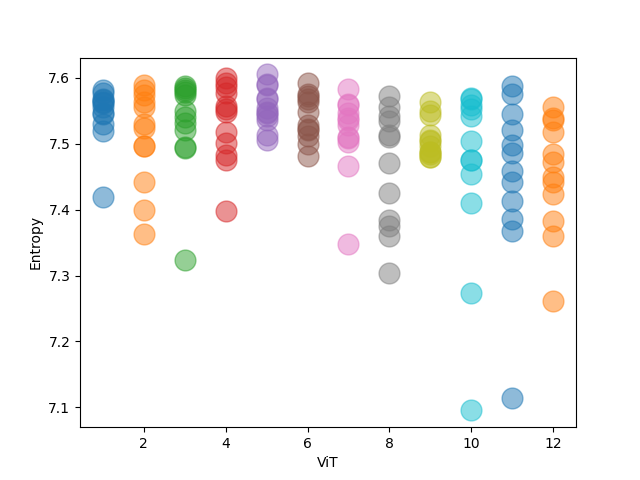}&\hspace{-0.3in}
    \includegraphics[align=c, height=3.cm]{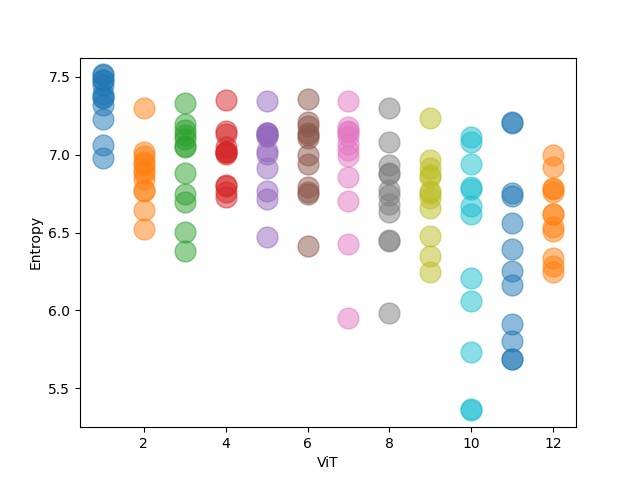}&\hspace{-0.2in}
    \includegraphics[align=c, height=3.cm]{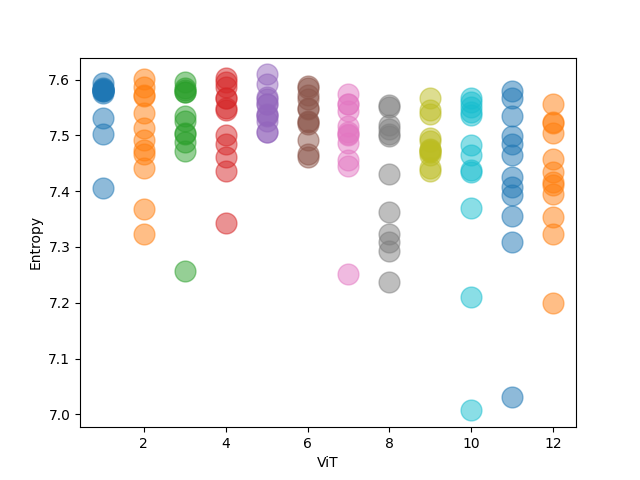}&\hspace{-0.3in}
    \includegraphics[align=c, height=3.cm]{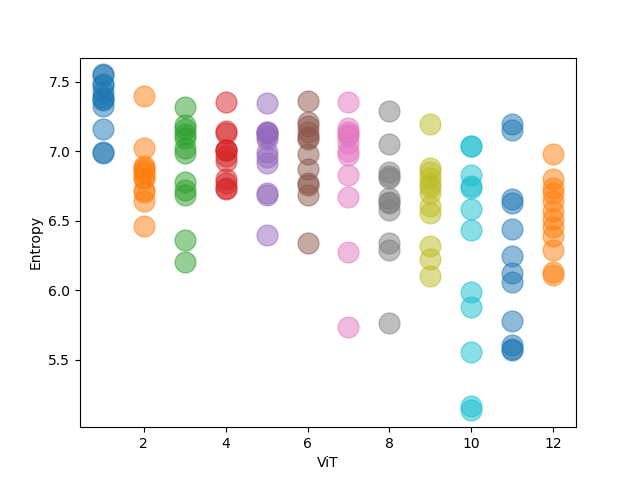}\\
    % \multicolumn{2}{c}{\footnotesize  (a) Supervised Learning} & \multicolumn{2}{c}{\footnotesize  (b) SimSiam (Contrastive)} & \multicolumn{2}{c}{\footnotesize  (b) SimMIM (Reconstruction)}\\
    % &\multicolumn{2}{c}{\footnotesize  (a) Task 0} & \multicolumn{2}{c}{\footnotesize  (b) Task 8} \\
    &
    \footnotesize $\bm{w}_{\bf{T0}}$ & \hspace{-0.3in}
    \footnotesize $\bm{w}_{\bf{T0 \rightarrow T8}}$ &\hspace{-0.2in}
    \footnotesize $\bm{w}_{\bf{T0}}$ &\hspace{-0.3in}
    \footnotesize $\bm{w}_{\bf{T0 \rightarrow T8}}$ \\
    \end{tabular}}
    \vspace{-0.1in}
    \caption{\footnotesize \textbf{ViT-B Attention entrtopy} of the first (T0) and the last task (T8) in a task sequence with respect to three continual pre-trained frameworks right after the completion of the first ($\bm{w}_{\bf{T0}}$) and last task ($\bm{w}_{\bf{T0\rightarrow T8}}$).}
    \label{sup:fig:vit-entr}
    \vspace{-0.1in}
\end{figure*}

%% file: materials/figures/5_swin_attn.tex
\begin{figure*}%[h!]
    \centering
    \resizebox{1\textwidth}{!}{%
    \hspace{-0.2in}
    \begin{tabular}{l@{\hspace{6pt}}cccccc}
    \rotatebox[origin=c]{90}{\footnotesize Supervised}\hspace{-0.1in}    
    &
    \includegraphics[align=c, height=2.8cm]{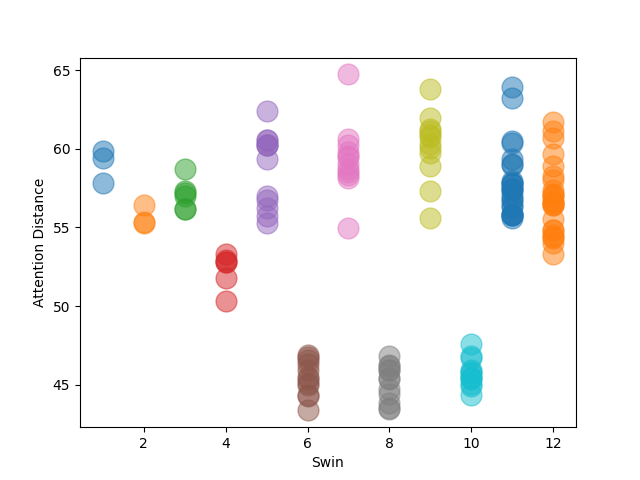}&\hspace{-0.3in}
    \includegraphics[align=c, height=2.8cm]{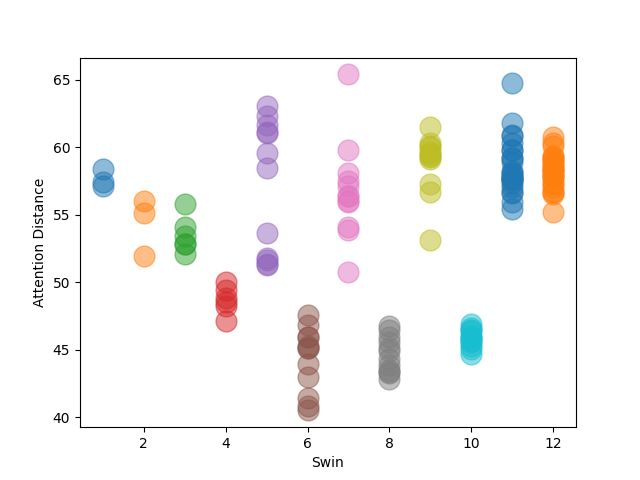}&\hspace{-0.2in}
    \includegraphics[align=c, height=2.8cm]{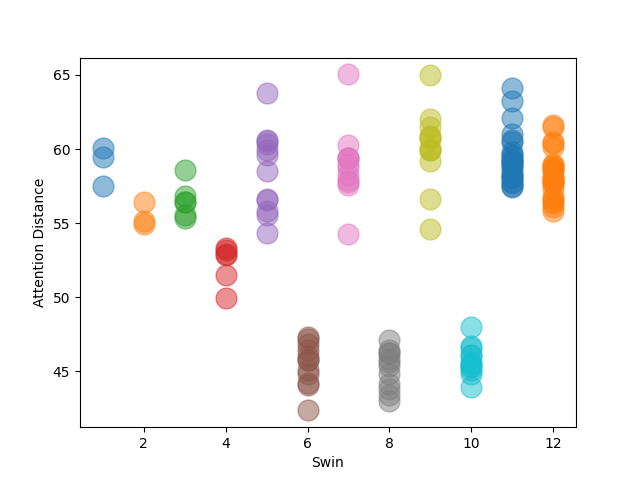}&\hspace{-0.3in}
    \includegraphics[align=c, height=2.8cm]{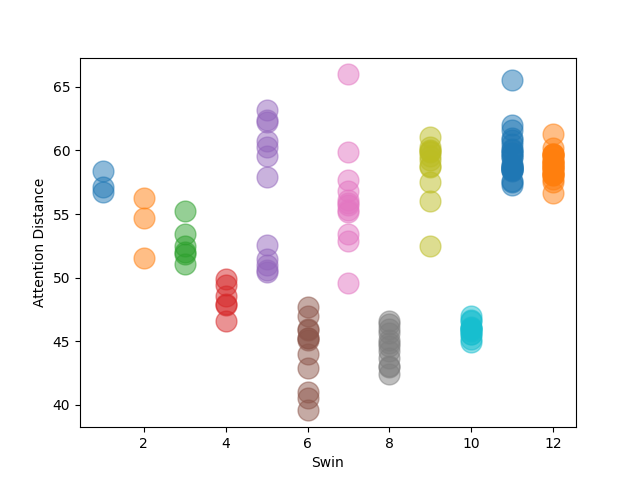}&\hspace{-0.2in}
    \includegraphics[align=c, height=2.8cm]{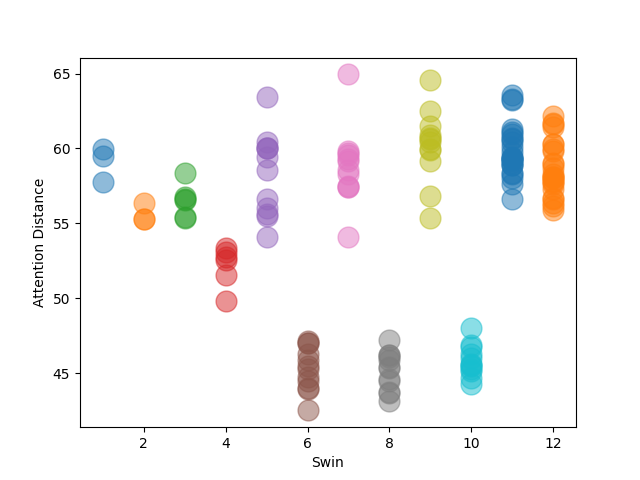}&\hspace{-0.3in}
    \includegraphics[align=c, height=2.8cm]{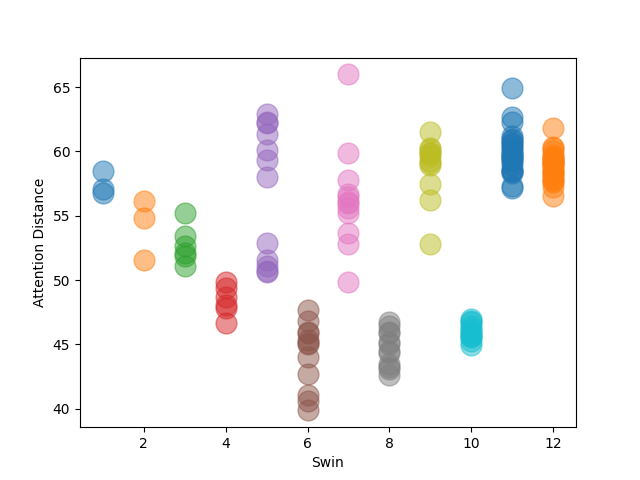}\\
    % \rotatebox[origin=c]{90}{\footnotesize Supervised (ep100)}\hspace{-0.1in}    
    % &
    % \includegraphics[align=c, height=2.8cm]{figures/sup/swin_attn_ep100/prtt0attt0.png}&\hspace{-0.3in}
    % \includegraphics[align=c, height=2.8cm]{figures/sup/swin_attn_ep100/prtt8attt0.png}&\hspace{-0.2in}
    % \includegraphics[align=c, height=2.8cm]{figures/sup/swin_attn_ep100/prtt0attt8.png}&\hspace{-0.3in}
    % \includegraphics[align=c, height=2.8cm]{figures/sup/swin_attn_ep100/prtt8attt8.png}&\hspace{-0.2in}
    % \includegraphics[align=c, height=2.8cm]{figures/sup/swin_attn_ep60/aa.png}&\hspace{-0.3in}
    % \includegraphics[align=c, height=2.8cm]{figures/sup/swin_attn_ep60/aa.png}\\
    \rotatebox[origin=c]{90}{\footnotesize SimSiam}\hspace{-0.1in}    
    &
    \includegraphics[align=c, height=2.8cm]{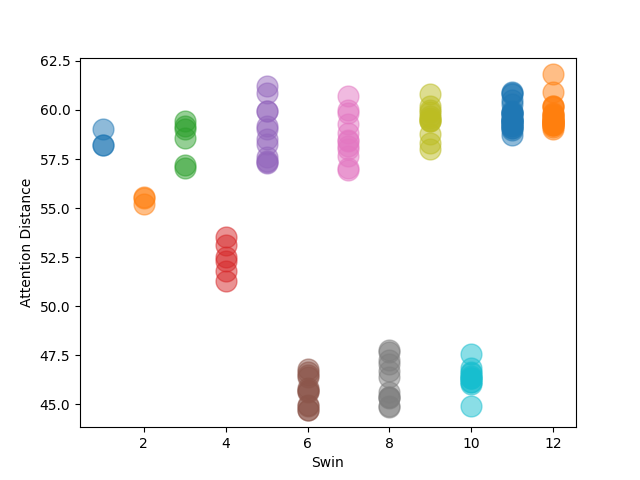}&\hspace{-0.3in}
    \includegraphics[align=c, height=2.8cm]{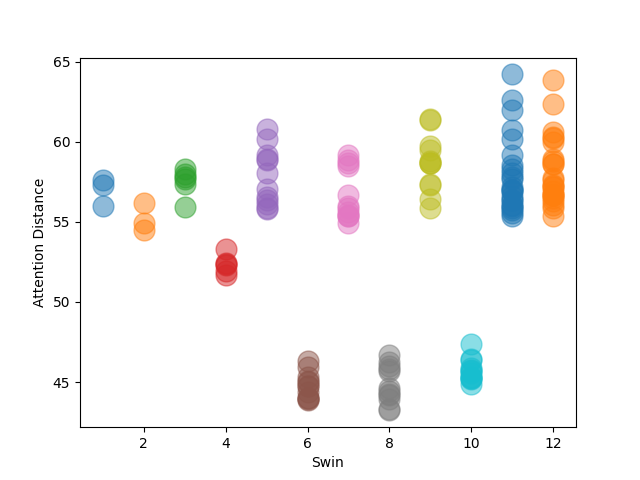}&\hspace{-0.2in}
    \includegraphics[align=c, height=2.8cm]{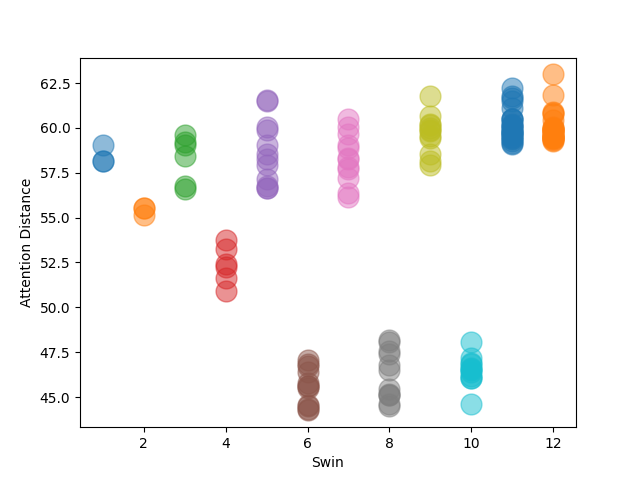}&\hspace{-0.3in}
    \includegraphics[align=c, height=2.8cm]{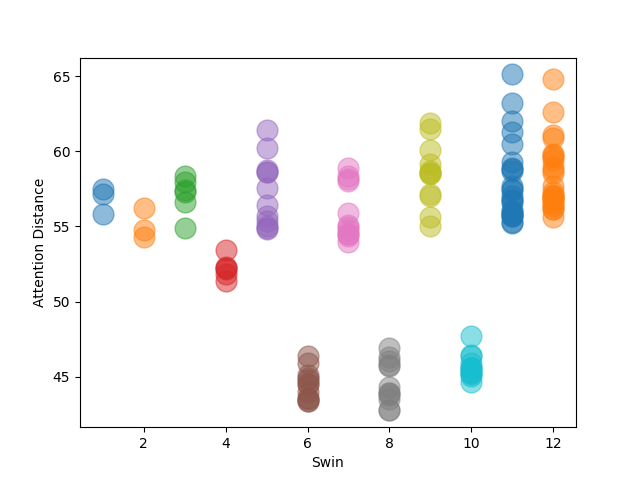}&\hspace{-0.2in}
    \includegraphics[align=c, height=2.8cm]{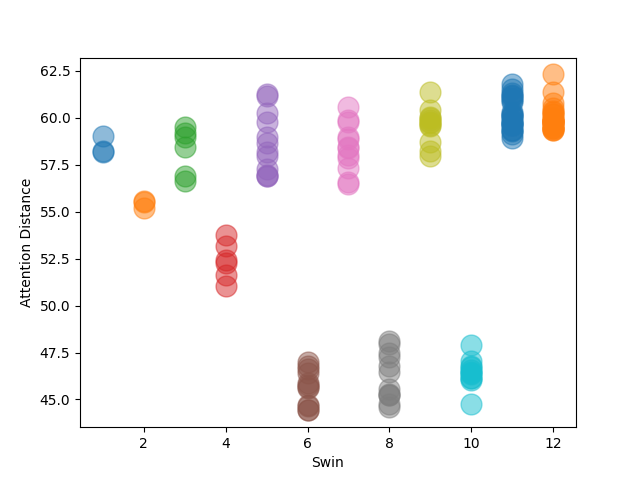}&\hspace{-0.3in}
    \includegraphics[align=c, height=2.8cm]{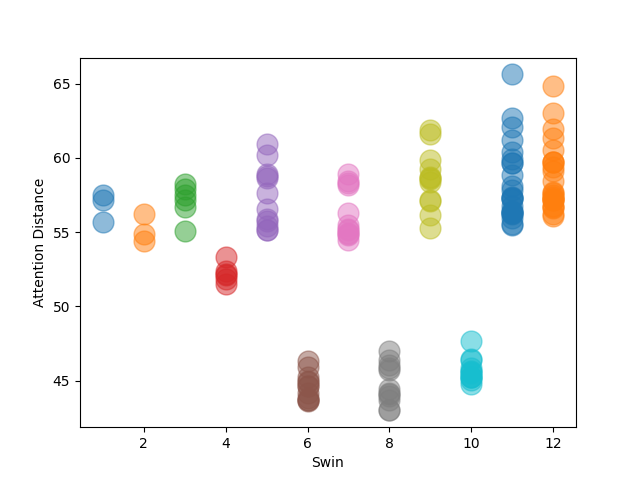}\\
    \rotatebox[origin=c]{90}{\footnotesize SimMIM}\hspace{-0.1in}
    &
    \includegraphics[align=c, height=2.8cm]{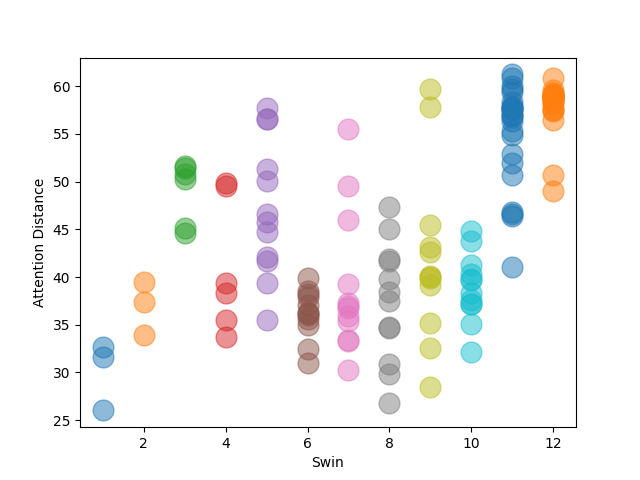}&\hspace{-0.3in}
    \includegraphics[align=c, height=2.8cm]{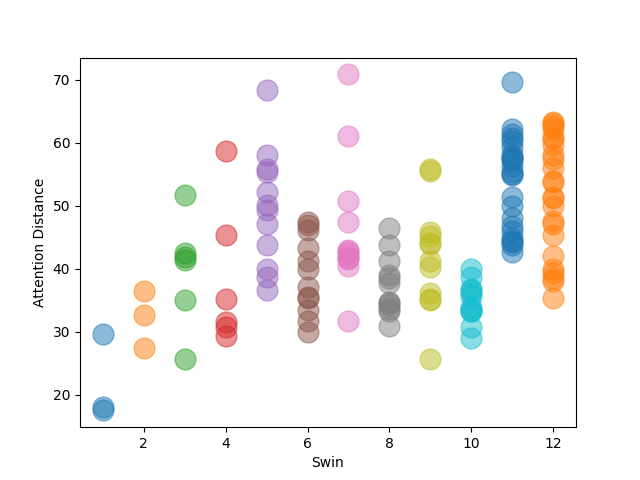}&\hspace{-0.2in}
    \includegraphics[align=c, height=2.8cm]{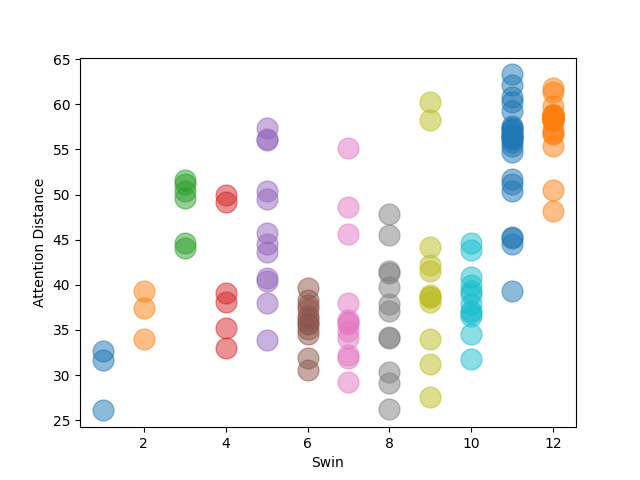}&\hspace{-0.3in}
    \includegraphics[align=c, height=2.8cm]{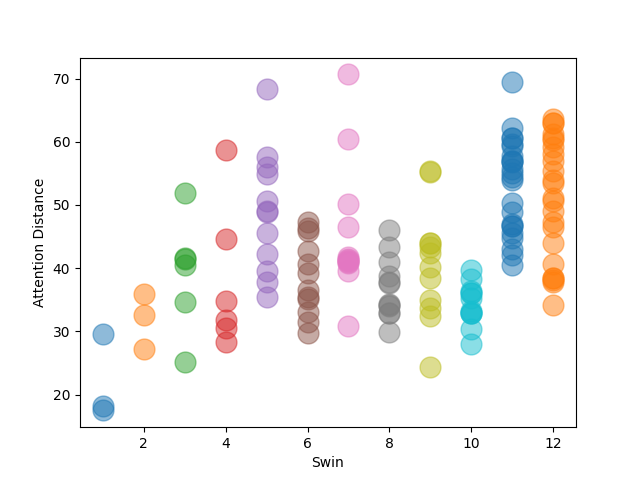}&\hspace{-0.2in}
    \includegraphics[align=c, height=2.8cm]{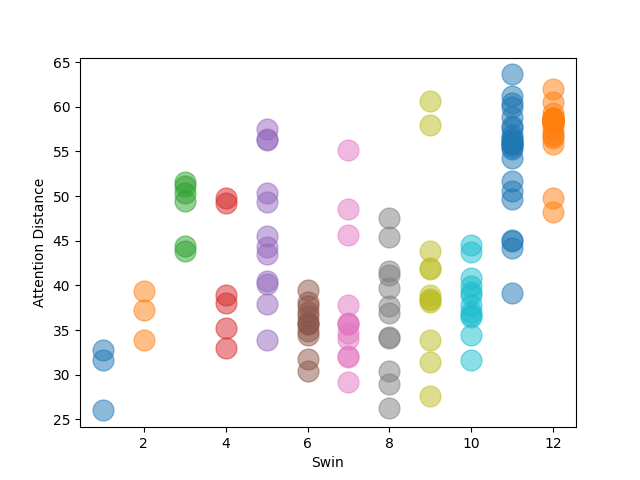}&\hspace{-0.3in}
    \includegraphics[align=c, height=2.8cm]{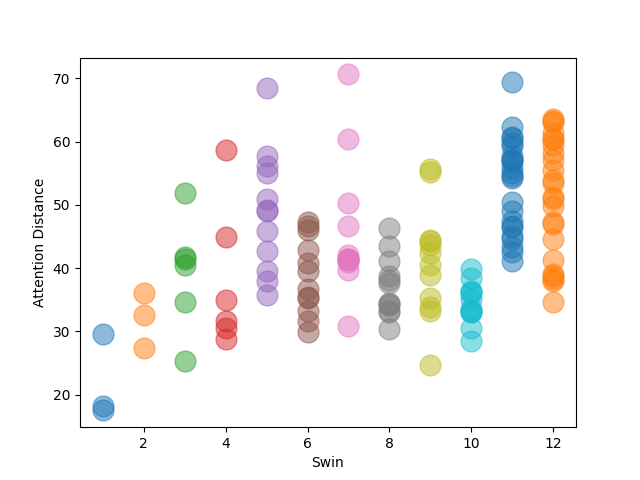}\\
    % \multicolumn{2}{c}{\footnotesize  (a) Supervised Learning} & \multicolumn{2}{c}{\footnotesize  (b) SimSiam (Contrastive)} & \multicolumn{2}{c}{\footnotesize  (b) SimMIM (Reconstruction)}\\
    % &\multicolumn{2}{c}{\footnotesize  (a) Task 0} & \multicolumn{2}{c}{\footnotesize  (b) Task 8} \\
    & 
    $\bf T0$ w.r.t. $\bf \mathcal{W}_{T0}$ &\hspace{-0.3in}
    $\bf T0$ w.r.t. $\bf \mathcal{W}_{T0\rightarrow T8}$ &\hspace{-0.2in}
    $\bf T8$ w.r.t. $\bf \mathcal{W}_{T0}$ &\hspace{-0.3in}
    $\bf T8$ w.r.t. $\bf \mathcal{W}_{T0\rightarrow T8}$ &\hspace{-0.2in}
    $\bf T9$ w.r.t. $\bf \mathcal{W}_{T0}$ &\hspace{-0.3in}
    $\bf T9$ w.r.t. $\bf \mathcal{W}_{T0\rightarrow T8}$ \\
    \end{tabular}}
    \vspace{-0.1in}
    \caption{\footnotesize \textbf{Swin-T attention distance} of an in-distribution (T0) and ood task (T9) with respect to three continual pre-trained frameworks right after the completion of the first ($\bm{w}_{\bf{T0}}$) and last task ($\bm{w}_{\bf{T0\rightarrow T8}}$).}
    \label{sup:fig:swin-attn}
    \vspace{-0.1in}
\end{figure*}
\begin{figure*}%[h!]
    \centering
    \resizebox{1\textwidth}{!}{%
    \hspace{-0.2in}
    \begin{tabular}{l@{\hspace{6pt}}ccccccc}
    \rotatebox[origin=c]{90}{\footnotesize Supervised}\hspace{-0.1in}    
    &
    \includegraphics[align=c, height=2.8cm]{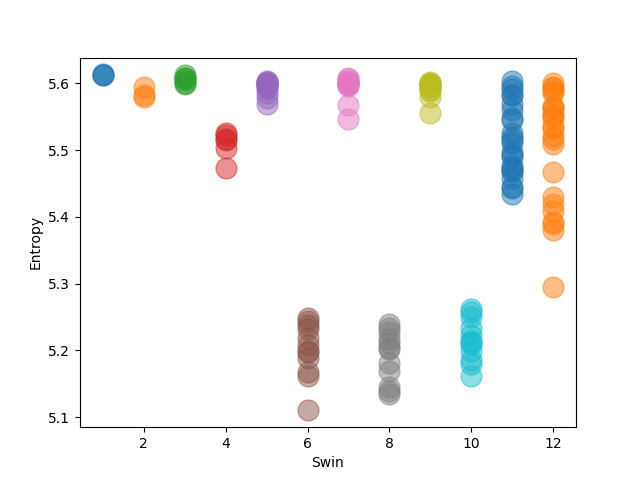}&\hspace{-0.3in}
    \includegraphics[align=c, height=2.8cm]{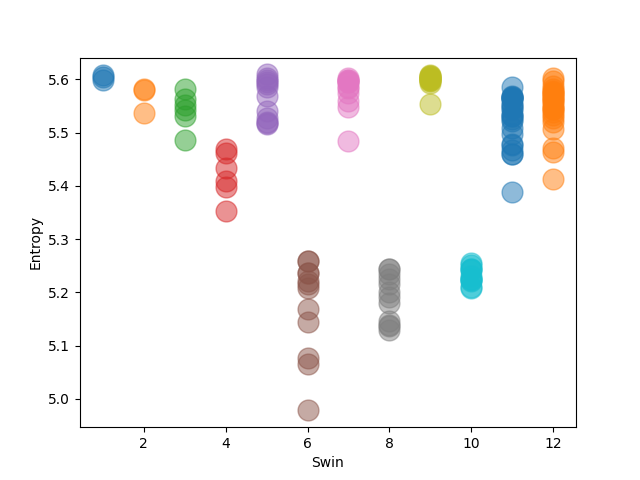}&\hspace{-0.2in}
    \includegraphics[align=c, height=2.8cm]{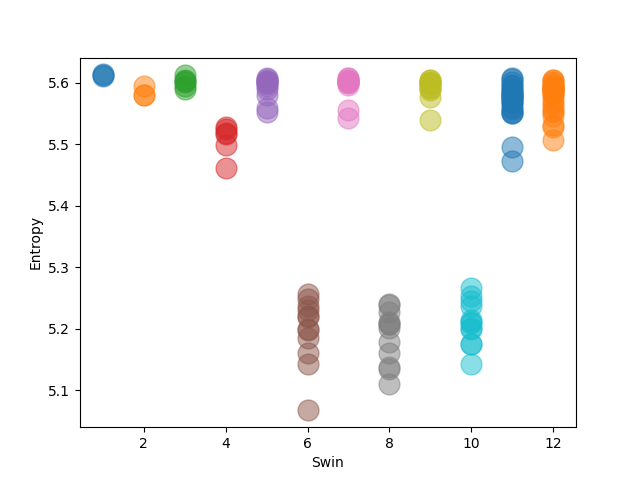}&\hspace{-0.3in}
    \includegraphics[align=c, height=2.8cm]{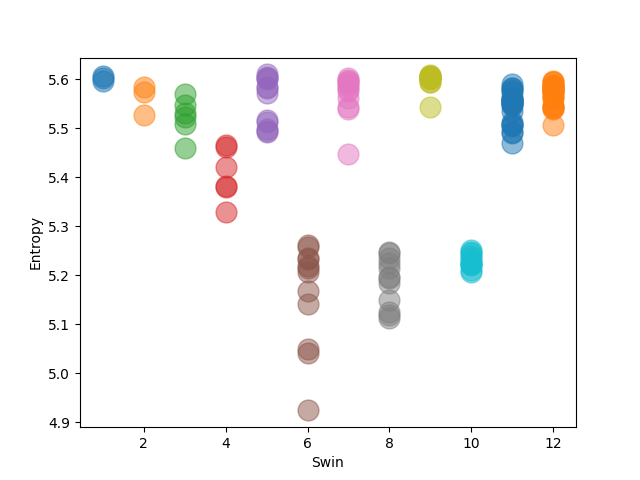}&\hspace{-0.2in}
    \includegraphics[align=c, height=2.8cm]{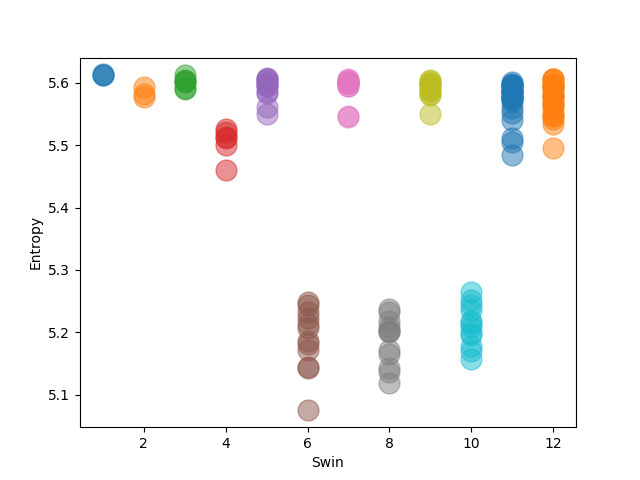}&\hspace{-0.3in}
    \includegraphics[align=c, height=2.8cm]{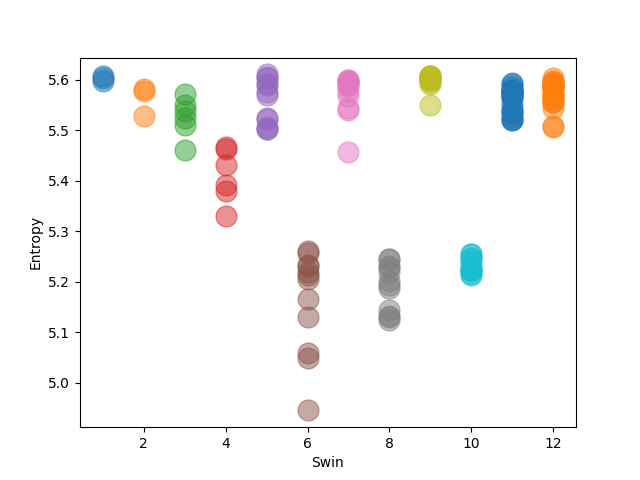}\\
    % \rotatebox[origin=c]{90}{\footnotesize Supervised (ep100)}\hspace{-0.1in}    
    % &
    % \includegraphics[align=c, height=2.8cm]{figures/sup/swin_entropy_ep100/prtt0attt0.png}&\hspace{-0.3in}
    % \includegraphics[align=c, height=2.8cm]{figures/sup/swin_entropy_ep100/prtt8attt0.png}&\hspace{-0.2in}
    % \includegraphics[align=c, height=2.8cm]{figures/sup/swin_entropy_ep100/prtt0attt8.png}&\hspace{-0.3in}
    % \includegraphics[align=c, height=2.8cm]{figures/sup/swin_entropy_ep100/prtt8attt8.png}\\
    \rotatebox[origin=c]{90}{\footnotesize SimSiam}\hspace{-0.1in}    
    &
    \includegraphics[align=c, height=2.8cm]{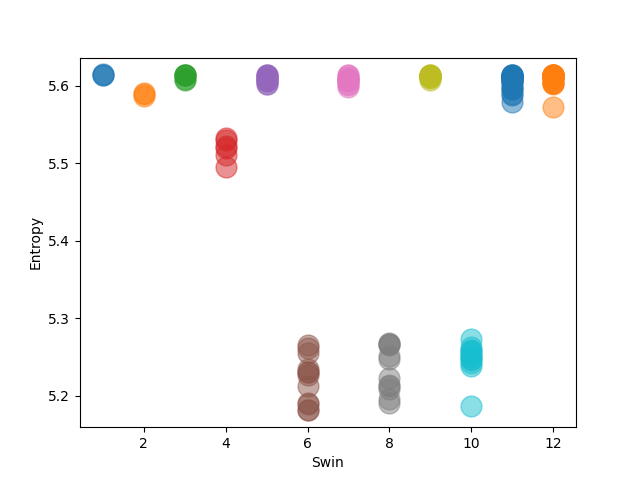}&\hspace{-0.3in}
    \includegraphics[align=c, height=2.8cm]{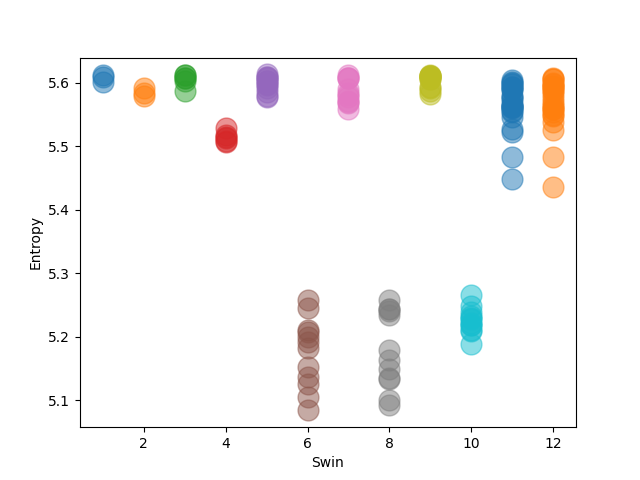}&\hspace{-0.2in}
    \includegraphics[align=c, height=2.8cm]{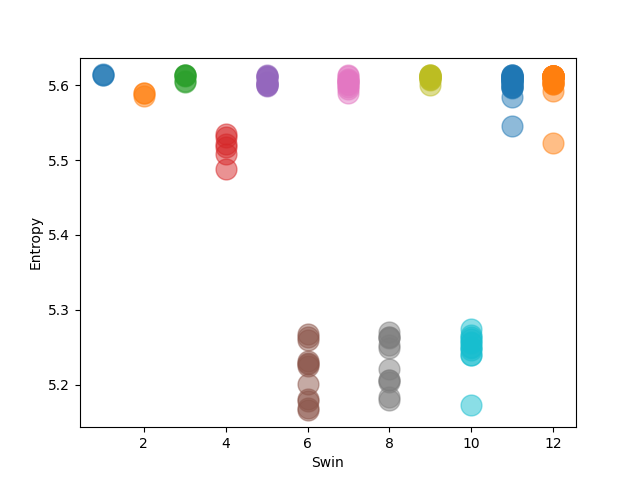}&\hspace{-0.3in}
    \includegraphics[align=c, height=2.8cm]{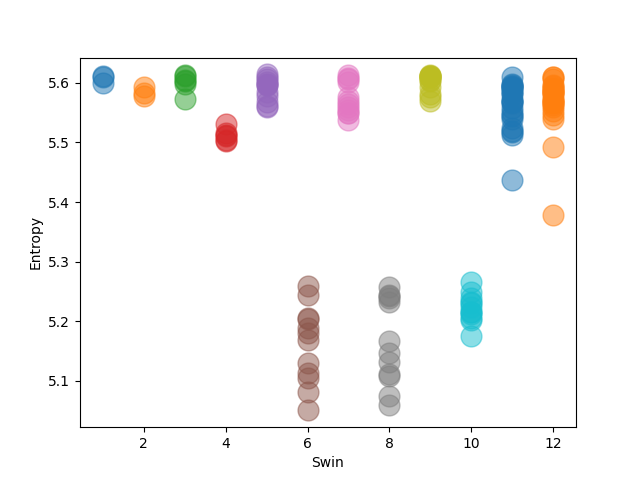}&\hspace{-0.2in}
    \includegraphics[align=c, height=2.8cm]{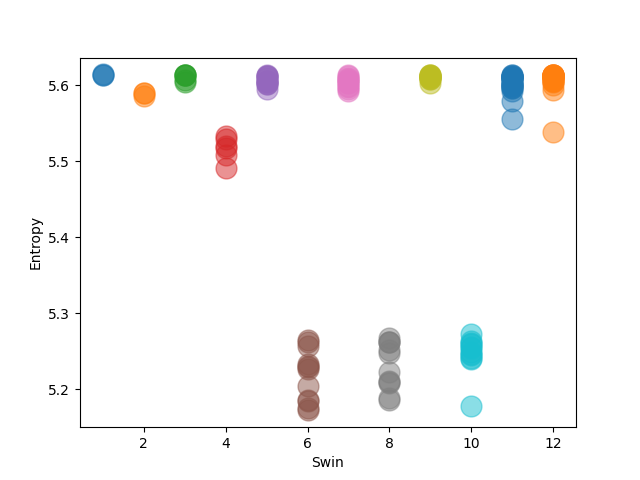}&\hspace{-0.3in}
    \includegraphics[align=c, height=2.8cm]{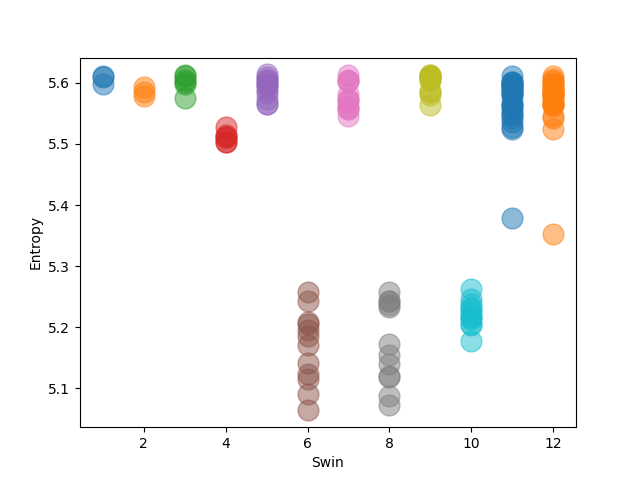}\\
    \rotatebox[origin=c]{90}{\footnotesize SimMIM}\hspace{-0.1in}
    &
    \includegraphics[align=c, height=2.8cm]{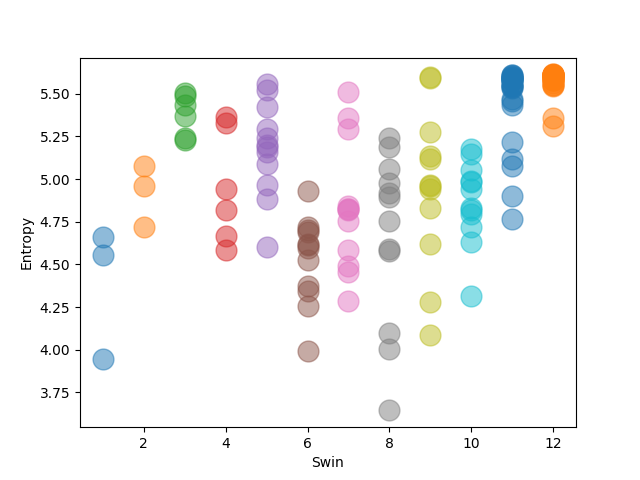}&\hspace{-0.3in}
    \includegraphics[align=c, height=2.8cm]{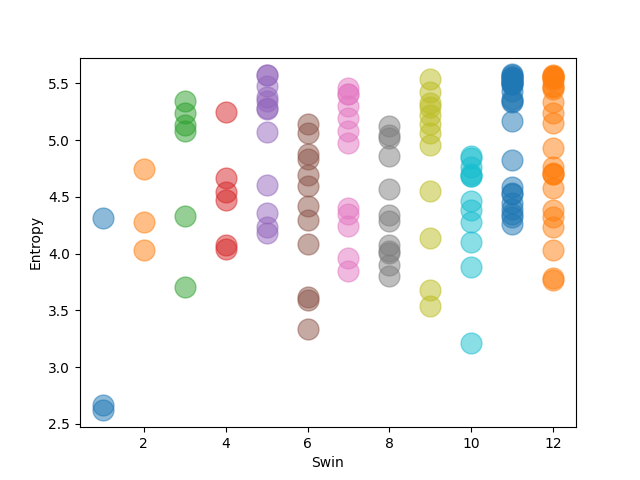}&\hspace{-0.2in}
    \includegraphics[align=c, height=2.8cm]{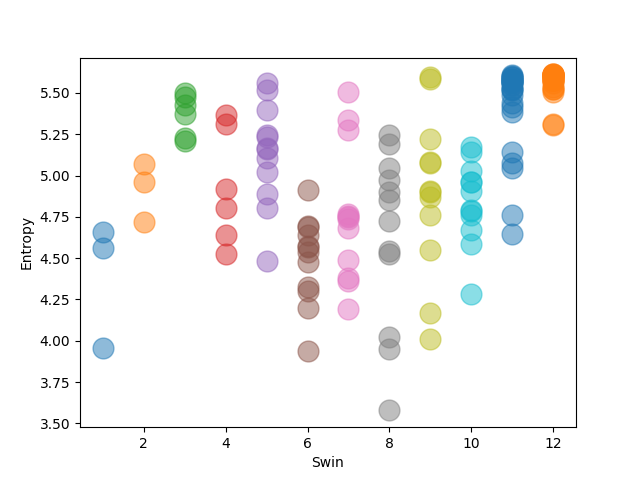}&\hspace{-0.3in}
    \includegraphics[align=c, height=2.8cm]{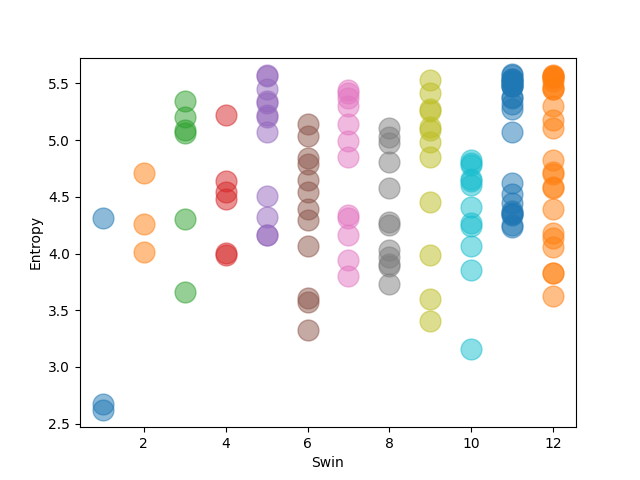}&\hspace{-0.2in}
    \includegraphics[align=c, height=2.8cm]{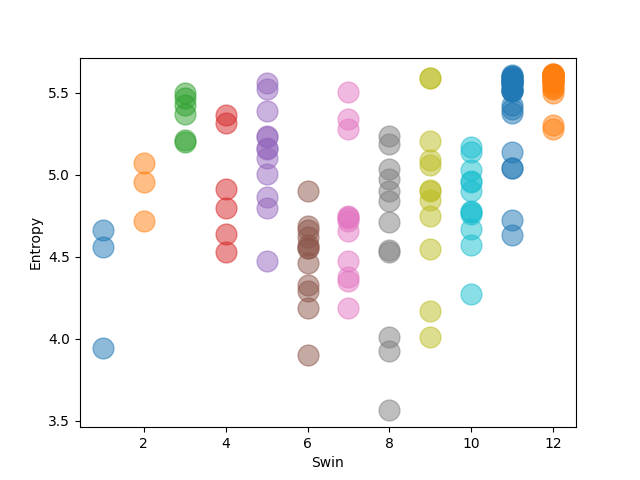}&\hspace{-0.3in}
    \includegraphics[align=c, height=2.8cm]{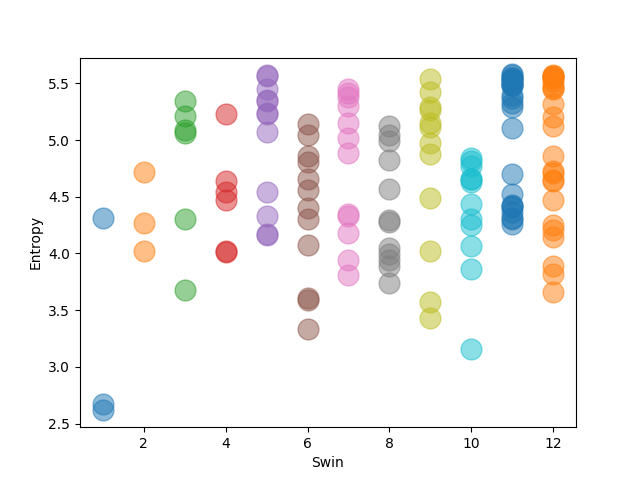}\\
    & 
    $\bf T0$ w.r.t. $\bf \mathcal{W}_{T0}$ &\hspace{-0.3in}
    $\bf T0$ w.r.t. $\bf \mathcal{W}_{T0\rightarrow T8}$ &\hspace{-0.2in}
    $\bf T8$ w.r.t. $\bf \mathcal{W}_{T0}$ &\hspace{-0.3in}
    $\bf T8$ w.r.t. $\bf \mathcal{W}_{T0\rightarrow T8}$ &\hspace{-0.2in}
    $\bf T9$ w.r.t. $\bf \mathcal{W}_{T0}$ &\hspace{-0.3in}
    $\bf T9$ w.r.t. $\bf \mathcal{W}_{T0\rightarrow T8}$ \\
    \end{tabular}}
    \vspace{-0.1in}
    \caption{\footnotesize \textbf{Swin-T attention entropy} of an in-distribution (T0) and ood task (T9) with respect to three continual pre-trained frameworks right after the completion of the first ($\bm{w}_{\bf{T0}}$) and last task ($\bm{w}_{\bf{T0\rightarrow T8}}$).}
    \label{sup:fig:swin-entr}
    % \vspace{-0.25in}
\end{figure*}

%% file: materials/figures/7_cf_partial_div.tex
\begin{figure*}[h!]
    % \vspace{-0.2in}
    \footnotesize
    \centering
    \resizebox{1\textwidth}{!}{%
    % \hspace{-0.2in}
    \begin{tabular}{lccc}
    \rotatebox[origin=c]{90}{\footnotesize Supervised}\hspace{-0.1in}&
    \includegraphics[align=c, height=3cm]{figures/attn_div/sup_attn_diversity_attt9_v2.png}&
    \includegraphics[align=c, height=3cm]{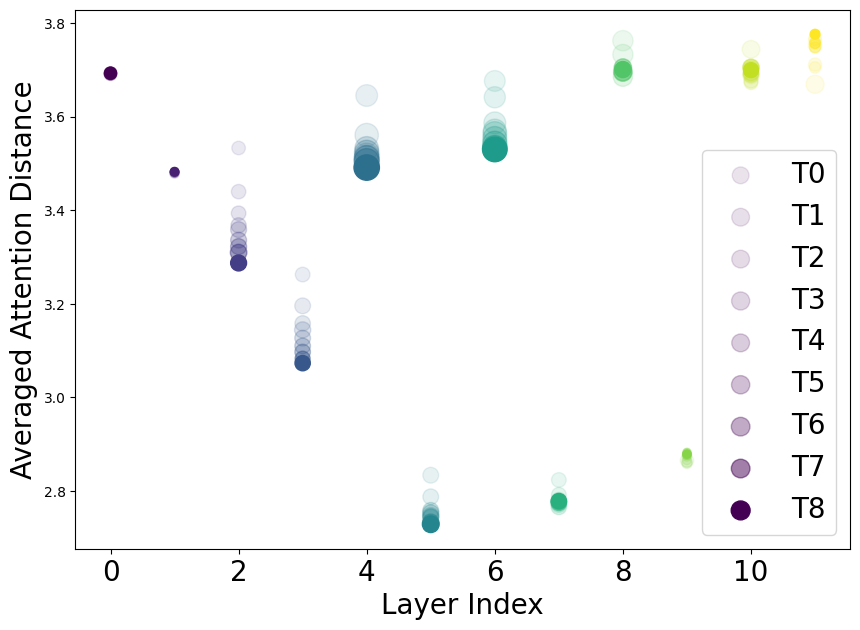}&
    \includegraphics[align=c, height=3cm]{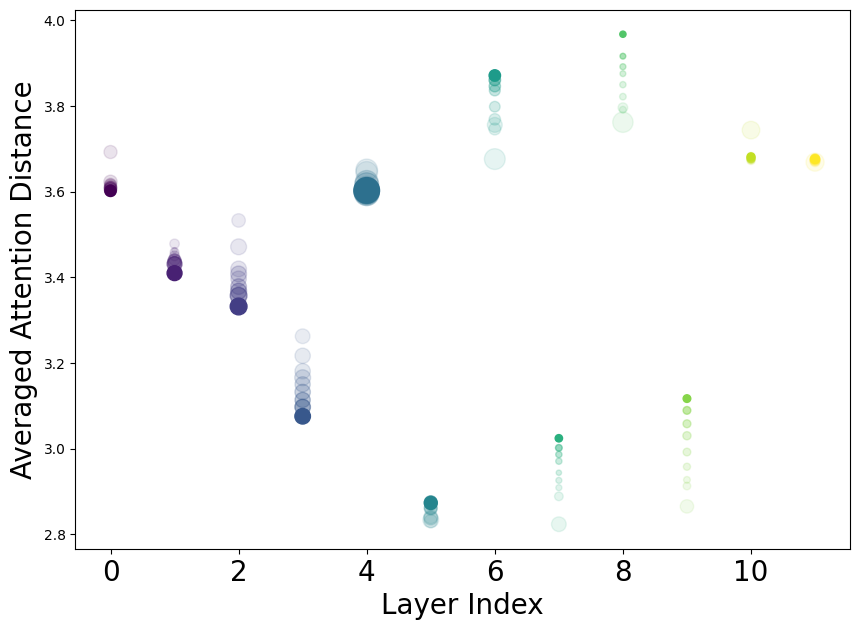}\\
    \rotatebox[origin=c]{90}{\footnotesize SimMIM}\hspace{-0.1in}&
    \includegraphics[align=c, height=3cm]{figures/attn_div/mim_attn_diversity_attt9_v2.png}&
    \includegraphics[align=c, height=3cm]{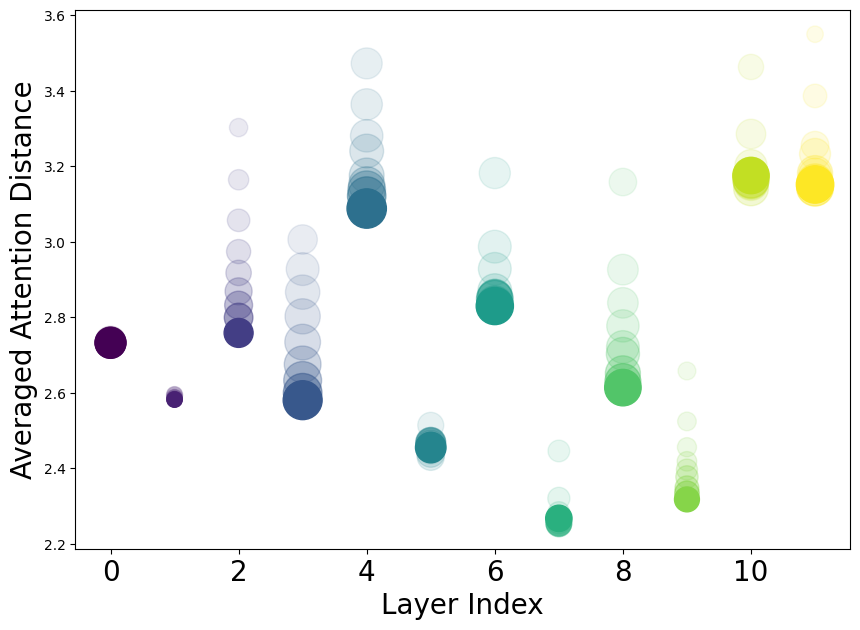}&
    \includegraphics[align=c, height=3cm]{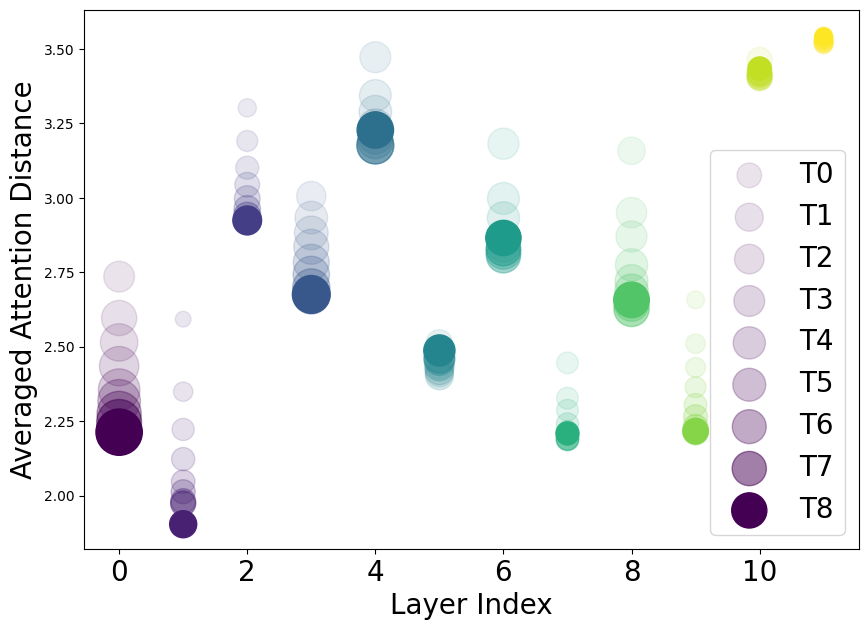}
    \\    
    &
    {Full-finetuning} &
    {Freezing lower layers} &
    {Freezing deeper layers} \\    
    % \rotatebox[origin=c]{90}{\footnotesize Attention Entropy}\hspace{-0.1in}&
    % \includegraphics[align=c, height=3cm]{figures/attn_div_partial/sup_entr_lower.png}&
    % \includegraphics[align=c, height=3cm]{figures/attn_div_partial/sup_entr_upper.png}&
    % \includegraphics[align=c, height=3cm]{figures/attn_div_partial/mim_entr_lower.png}&
    % \includegraphics[align=c, height=3cm]{figures/attn_div_partial/mim_entr_upper.png}
    % \\
    % &
    % {Sup freezing lower layers} &
    % {Sup freezing deeper layers} &
    % {MIM freezing lower layers} &
    % {MIM freezing deeper layers} \\
    \end{tabular}}
    \vspace{-0.1in}
    \caption{\footnotesize \textbf{Visualization of aggregated attention distance} on an OOD task (T9) at each layer at the end of each continual pre-training task phase (T0$\rightarrow$T8). We freeze the two lowest or deepest layers after pre-training the first task (T0). The radius of the marker indicates the standard deviation over attention heads in the corresponding layer.}
    \label{sup:fig:partial-freeze-swin-attn-div}
    % \vspace{-0.1in}
\end{figure*}

%% file: materials/figures/7_cf_partial_div_entr.tex
\begin{figure*}[h!]
    % \vspace{-0.2in}
    \footnotesize
    \centering
    \resizebox{1\textwidth}{!}{%
    % \hspace{-0.2in}
    \begin{tabular}{lccc}
    \rotatebox[origin=c]{90}{\footnotesize Supervised}\hspace{-0.1in}&
    \includegraphics[align=c, height=3cm]{figures/attn_div/sup_entr_diversity_attt9_v2.png}&
    \includegraphics[align=c, height=3cm]{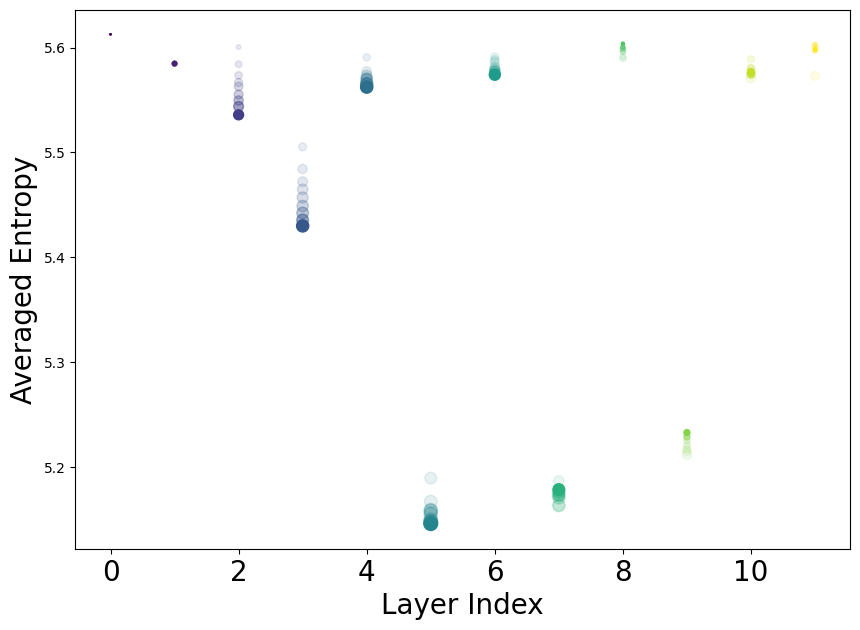}&
    \includegraphics[align=c, height=3cm]{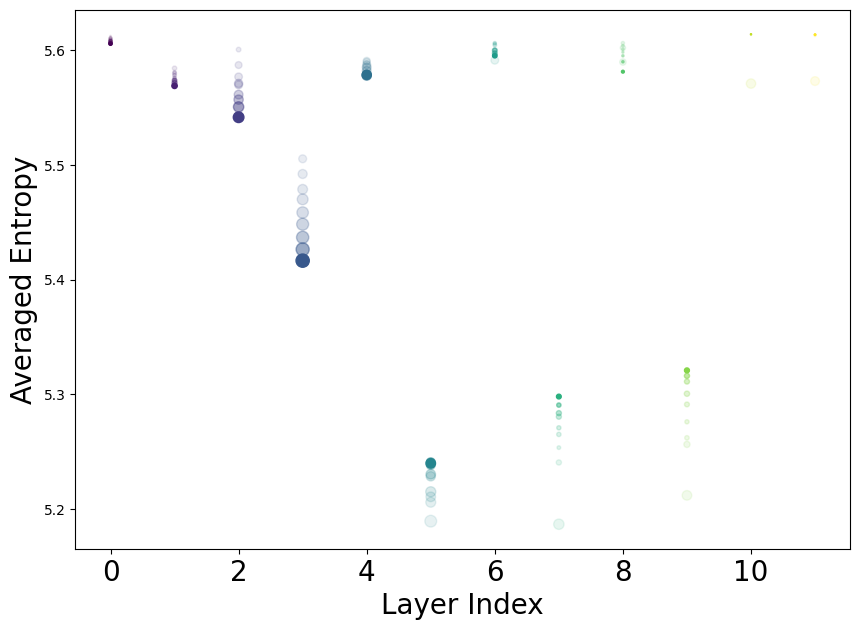}\\
    \rotatebox[origin=c]{90}{\footnotesize SimMIM}\hspace{-0.1in}&
    \includegraphics[align=c, height=3cm]{figures/attn_div/mim_entr_diversity_attt9_v2.png}&
    \includegraphics[align=c, height=3cm]{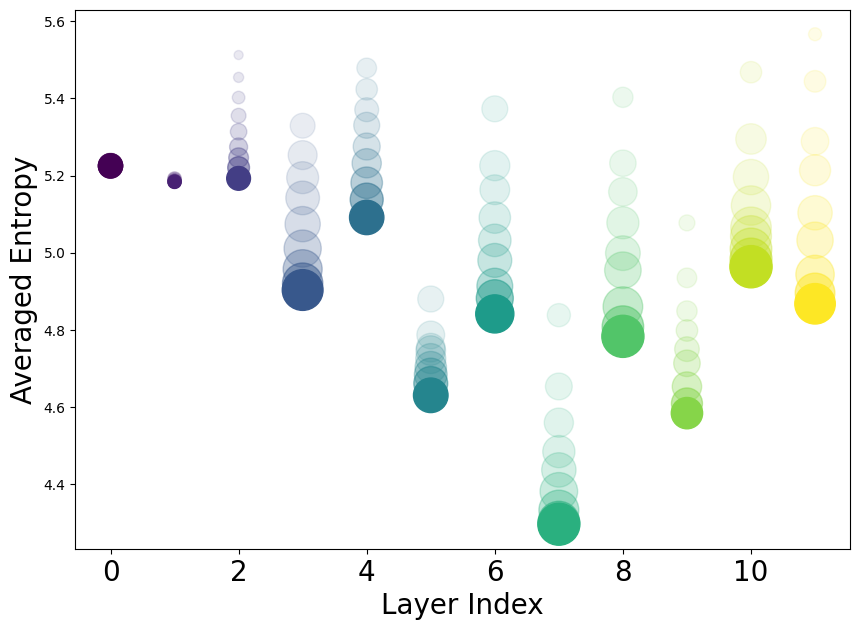}&
    \includegraphics[align=c, height=3cm]{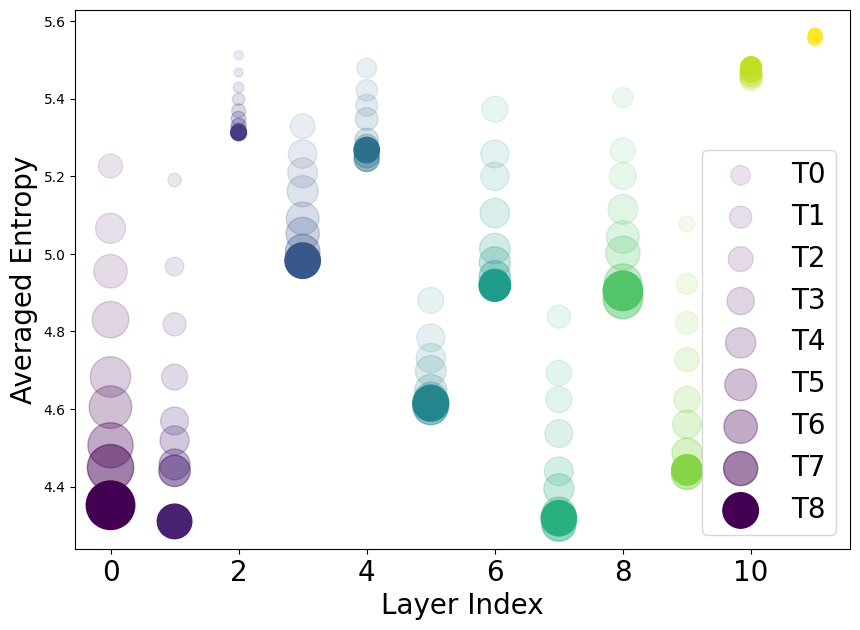}
    \\    
    &
    {Full-finetuning} &
    {Freezing lower layers} &
    {Freezing deeper layers} \\    
    \end{tabular}}
    \vspace{-0.1in}
    \caption{\footnotesize \textbf{Visualization of aggregated attention entropy} on an OOD task (T9) at each layer at the end of each continual pre-training task phase (T0$\rightarrow$T8). We freeze the two lowest or deepest layers after pre-training the first task (T0). The radius of the marker indicates the standard deviation over attention heads in the corresponding layer.}
    \label{sup:fig:partial-freeze-swin-entr-div}
    % \vspace{-0.1in}
\end{figure*}